\documentclass{article}

\usepackage[final]{corl_2022} 

\title{TrackletMapper: Ground Surface Segmentation \\ and Mapping from Traffic Participant Trajectories}

\author{
  Jannik Zürn$^{1}$\thanks{Equal contribution} \quad Sebastian Weber$^{1*}$ \quad Wolfram Burgard$^{2}$ \\
  {$^1$University of Freiburg} \\ $^2$University of Technology Nuremberg \\
  {\tt\small zuern@cs.uni-freiburg.de, wolfram.burgard@utn.de} \\
}

\usepackage{amsmath} 
\usepackage{amssymb}  
\usepackage{graphicx}
\usepackage{hyperref}
\usepackage[table]{xcolor}
\usepackage{siunitx}
\usepackage{array}
\usepackage{adjustbox}
\usepackage{todonotes}
\usepackage{pifont}
\usepackage{url}
\usepackage{booktabs}
\usepackage{makecell}
\usepackage{wrapfig,lipsum,booktabs}
\usepackage{caption} 
\usepackage{listings}

\newcommand\crule[3][black]{\textcolor{#1}{\rule{#2}{#3}}}
\newcolumntype{P}[1]{>{\centering\arraybackslash}p{#1}}

\definecolor{road}{RGB}{128, 148, 255}
\definecolor{crossing}{RGB}{42, 255, 61}
\definecolor{pedestrian}{RGB}{255, 226, 108}
\definecolor{obstacle}{RGB}{178, 73, 73}

\newcommand{\etal}{\emph{et~al.}}

\DeclareMathOperator*{\argmax}{arg\,max}

\newcolumntype{R}[2]{%
    >{\adjustbox{angle=#1,lap=\width-(#2)}\bgroup}%
    l%
    <{\egroup}%
}

\definecolor{codegreen}{rgb}{0,0.6,0}
\definecolor{codegray}{rgb}{0.5,0.5,0.5}
\definecolor{codepurple}{rgb}{0.58,0,0.82}
\definecolor{backcolour}{rgb}{0.95,0.95,0.92}

\lstdefinestyle{mystyle}{
    backgroundcolor=\color{backcolour},   
    commentstyle=\color{codegreen},
    keywordstyle=\color{magenta},
    numberstyle=\tiny\color{codegray},
    stringstyle=\color{codepurple},
    basicstyle=\ttfamily\footnotesize,
    breakatwhitespace=false,         
    breaklines=true,                 
    captionpos=b,                    
    keepspaces=true,                 
    numbers=left,                    
    numbersep=5pt,                  
    showspaces=false,                
    showstringspaces=false,
    showtabs=false,                  
    tabsize=2
}

\newcommand\rebuttal[1]{\textcolor{black}{#1}}

\begin{document}
\maketitle


\begin{abstract}
    Robustly classifying ground infrastructure such as roads and street crossings is an essential task for mobile robots operating alongside pedestrians. While many semantic segmentation datasets are available for autonomous vehicles, models trained on such datasets exhibit a large domain gap when deployed on robots operating in pedestrian spaces. Manually annotating images recorded from pedestrian viewpoints is both expensive and time-consuming. To overcome this challenge, we propose \textit{TrackletMapper}, a framework for annotating ground surface types such as sidewalks, roads, and street crossings from object tracklets without requiring human-annotated data. To this end, we project the robot ego-trajectory and the paths of other traffic participants into the ego-view camera images, creating sparse semantic annotations for multiple types of ground surfaces from which a ground segmentation model can be trained. We further show that the model can be self-distilled for additional performance benefits by aggregating a ground surface map and projecting it into the camera images, creating a denser set of training annotations compared to the sparse tracklet annotations. We qualitatively and quantitatively attest our findings on a novel large-scale dataset for mobile robots operating in pedestrian areas. Code and dataset are available at \url{http://trackletmapper.cs.uni-freiburg.de}.
\end{abstract}


\begin{figure}[h]
\centering
\includegraphics[width=14cm]{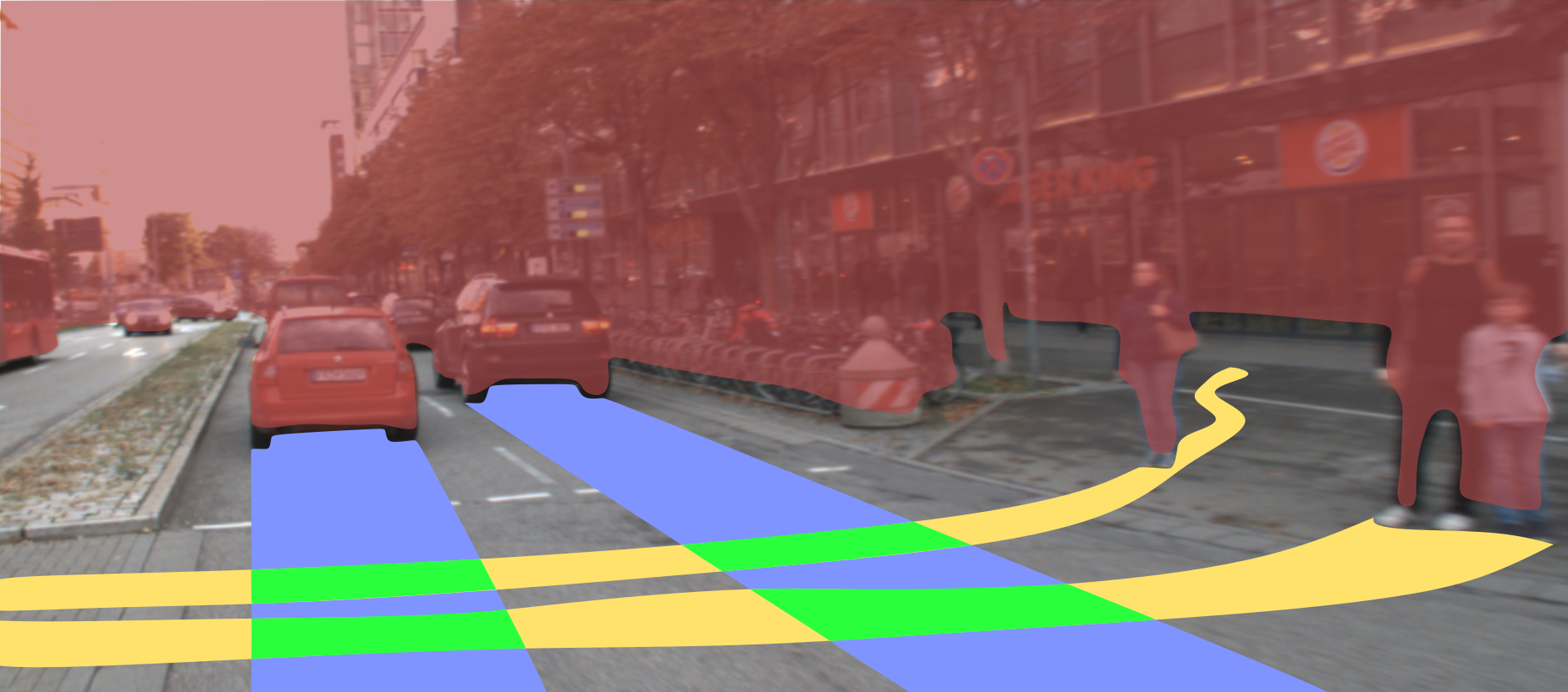}
\caption{We present \textit{TrackletMapper}, a novel approach for ground surface segmentation that leverages observed traffic participant tracklets to supervise a surface segmentation model. Our pipeline automatically annotates obstacles (red), pedestrian areas/sidewalks (yellow), street crossings (green) and roads (blue) based on the observed trajectories of pedestrians and vehicles.}
\label{fig:covergirl}
\end{figure}


\section{Introduction}
\label{sec:introduction}

Mobile ground robots operating in urban areas encounter a wide range of environments. It is essential for autonomous robots to robustly navigate through such environments even without access to human-annotated map data. Urban environments feature different types of ground surfaces restricted for use only by particular traffic participants. While vehicles are mostly permitted to operate on roads, pedestrians are generally only allowed on sidewalks and in pedestrian areas. Street crossings are permitted to be crossed both by vehicles and by pedestrians. To allow for robust and safe navigation, autonomously operating robots in urban environments are required to localize nearby traffic participants accurately \cite{radwan2017did, radwan2020multimodal, zurn2022self} and classify ground surfaces robustly. While autonomous vehicles typically require a binary distinction between road and non-road surfaces, mobile robots operating in pedestrian spaces must crucially be able to distinguish between sidewalks, roads, and road crossings in order to navigate urban environments safely \cite{kummerle2015autonomous, radwan2018effective, petek2021robust}.

In recent years, a multitude of semantic segmentation datasets for urban autonomous driving has been proposed \cite{neuhold2017mapillary, cordts2016cityscapes, maddern20171}. These datasets, however, are recorded from the vantage point of street vehicles. Therefore, models trained on these datasets exhibit a strong bias toward the camera viewpoint, introducing a significant domain gap when deployed in areas intended for non-vehicle usages such as pedestrian areas or sidewalks. While it is possible to manually annotate images obtained from the pedestrian viewpoint, this is an expensive and time-consuming task. Automatic annotation of images offers a promising alternative to manual annotations made by human annotators. Previously proposed automatic annotation approaches typically leverage the ego-motion of a data collection platform to obtain spatially sparse image-level labels of traversable ground surfaces  \cite{barnes2017find, onozuka2021weakly} or are based on proprioceptive sensors such as sound and vibration \cite{brooks2012self, zurn2020self, wellhausen2019should}. In contrast to existing work, we additionally leverage the trajectories of other traffic participants such as vehicles and pedestrians, and project them into the camera images. This enables us to label multiple types of ground surfaces, including roads, sidewalks or pedestrian areas, and street crossings based on the type of tracked objects. Hereby, we leverage the fact that under most circumstances, pedestrians walk in areas reserved for them and vehicles drive on roads or through street crossings but not on the sidewalk. The object detector used to generate the pedestrian- and vehicle trajectories does not suffer from the viewpoint-induced domain gap present in segmentation models. To further boost model performance, we build a ground surface map from these predictions by spatially aggregating the predictions. Aggregation of semantic segmentation predictions has been previously proposed~\cite{paz2020probabilistic}, however, the generated maps have previously not been used as an annotation source for semantic segmentation models. We show that it is possible to self-distill the segmentation model by re-projecting the aggregated surface map back into the camera images and using them as annotations, boosting the model performance.

In summary, this work offers the following key contributions: (i) A novel automatic annotation approach that leverages trajectories of traffic participants such as vehicles and pedestrians for generating sparse multi-class semantic pixel annotations. (ii) A segmentation model self-distillation pipeline to generate training annotations from projections of an aggregated surface map. (iii) The \textit{Freiburg Pedestrian Scenes} dataset recorded with a robot platform navigating through a wide range of urban pedestrian environments.

\section{Related Works}
\label{sec:relatedworks}

Self-supervised methods for visual terrain segmentation in off-road driving applications were investigated in \cite{brooks2012self, otsu2016autonomous, valada2017deep, wellhausen2019should, zurn2020self}. In these works, labels obtained from a proprioceptive sensor modality (i.e. vibration, sound) are used to partially annotate exteroceptive sensor modalities (i.e. RGB vision). Other non-learning approaches leverage geometric features in LiDAR point clouds to classify vertical and horizontal surfaces \cite{moosmann2009segmentation, douillard2010hybrid, aijazi2013segmentation}.

One of the first works to consider auto-generated annotations for semantic image segmentation in the context of autonomous driving was Barnes \etal~\cite{barnes2017find}. The authors propose a self-supervised approach for generating drivable paths in monocular RGB images from projected ego-trajectories of the recording vehicle on popular urban driving datasets. Mayr \etal~\cite{mayr2018self} propose a self-labeling pipeline for drivable road area segmentation. Based on stereo disparity maps and ground plane fitting, they extract drivable road areas from images and use the annotated RGB images to train a binary segmentation model. Cho \etal~\cite{cho2018multi} estimate drivable space and surface normal vectors from stereo images, which are used as pseudo-ground-truth to train a segmentation model. Bruls \etal~\cite{bruls2018mark} leverage weakly-labeled annotations for urban road markings based on LiDAR reflectance values and potentials from a Conditional Random Field. Wang \etal~\cite{wang2019self} propose a self-supervised drivable area and road anomaly segmentation approach from RGB-D data. They leverage a stereo depth image to obtain weak labels for obstacles sticking out from the ground level. Wellhausen \etal~\cite{wellhausen2019should} propose a self-supervised weak image labeling scheme based on a proprioceptive vibration-based terrain classifier. Labels predicted by the proprioceptive classifier are projected into the robot camera ego-view.  Z\"urn \etal~\cite{zurn2020self} propose a self-supervised labeling scheme based on an unsupervised audio clustering approach, where the cluster indices serve as weak labels and are projected into the robot camera images. Most recently, Onozuka \etal~\cite{onozuka2021weakly} propose a traversable area segmentation approach for personal mobility systems such as intelligent wheelchairs.
 
To summarize, existing methods for automatic annotation or self-supervised approaches do not leverage the additional data provided by the trajectories of other traffic participants, thus, ignoring relevant information. In addition, our work makes use of the aggregated surface map as an additional annotation source, further boosting the segmentation model performance by increasing the number of annotated pixels.
\section{Technical Approach}
\label{sec:approach}

Our goal is to label the surface classes \textit{Pedestrian}, \textit{Road}, \textit{Crossing}, and \textit{Obstacle}. The classes \textit{Pedestrian} and \textit{Road} contain surface areas, where either of the two classes is exclusively permitted. Areas intended for pedestrian use include sidewalks, pedestrian zones, and footpaths while vehicle areas include all road sections without crossings. The class \textit{Crossing} is intended to annotate asphalt surfaces at street crossings (zebra crossings or signaled pedestrian crossings). Both pedestrians and vehicles are permitted to cross these areas. Pixels labeled from the ego-trajectory and those obtained from pedestrian trajectories are jointly used to provide annotations for the class \textit{Pedestrian} since we assume that the robot is teleoperated to only traverse pedestrian surfaces. The class \textit{Obstacle} annotates different kinds of non-traversable surfaces such as buildings, moving or static objects extending over ground or vegetation. The class \textit{Unknown} serves as a filler class for all pixels where no annotation is provided. In the following, we will first discuss the automatic generation of image annotations from the robot ego-trajectory and traffic participant tracklets (Subsec. \ref{sec:auto}) and subsequently the generation and projection of the semantic surface map for additional model performance gains (Subsec. \ref{sec:mapping}).

\subsection{Surface Annotations from Tracklets}
\label{sec:auto}

We first perform LiDAR-SLAM \cite{kummerle2011g}, generating a list of poses $\mathbf{p}_i \in \mathbb{SE}(3)$ for the robot base for each data collection run. In the following, we will discuss the projection of the ego-trajectory into image coordinates. In order to project the robot trajectory into the viewpoint of the onboard camera, we associate a time-synchronized robot pose with each of the camera images. Assuming a static transform \rebuttal{$\mathbf{T}^B_C$}  between the robot base and the camera mounting position relative to the base, the robot trajectory in homogeneous pixel coordinates $\mathbf{u} = [u,v,1]^T$ can be expressed as

\rebuttal{
\begin{equation}
\label{eq:projection}
\mathbf{u} = \mathbf{K} \mathbf{T}^B_C \mathbf{T}^W_B  \mathbf{\hat{p}},
\end{equation}
}

where \rebuttal{$\mathbf{T}^W_B$} is the time-dependent transformation between the world coordinates and the current robot base position, obtained from $\mathbf{p}_i$, $\mathbf{K} \in \mathbb{R}^{3 \times 3} $ denotes the intrinsic camera matrix, and $\mathbf{\hat{p}} \in \mathbb{R}^3$ denotes the ego-trajectory projected onto the ground surface. For brevity, we omit the superscript $t$ for time-dependent variables. Note that we dilate the robot trajectory laterally by half its base width in order to label all pixels within the robot footprint.

To obtain the trajectories of other traffic participants such as vehicles and pedestrians, we leverage the ByteTrack \cite{zhang2021bytetrack} object tracker with an EfficientDet \cite{tan2020efficientdet} object detector pre-trained on the MS-COCO dataset \cite{lin2014microsoft}. The object trajectory in 3D world coordinates is obtained by projecting the tracklet bounding box center point coordinates into 3D world coordinates. To perform this transformation, we interpolate sparse depth images obtained from the LiDAR points, which provides an accurate depth estimation for a given object bounding box. Formally, the projection of tracklets into 3D world coordinates $\mathbf{x} \in \mathbb{R}^3$ follows the inverse projection equation:

\rebuttal{
\begin{equation}
\label{eq:backwardprojection}
\mathbf{x} = \mathbf{T}^B_W \mathbf{T}^C_B d ~ \mathbf{K}^{-1} \mathbf{u},
\end{equation}
}

where we follow the same naming convention as in Eq. \ref{eq:projection} and the scalar $d \in \mathbb{R}$ denotes the depth scaling factor. Similar to the projection of the ego-trajectory, we assign labels to image pixels according to the 3-D world tracklet projection into image coordinates, according to Eq. \ref{eq:projection}. Similar to the ego-trajectory, we laterally dilate the tracklet line segments by a fixed object width, which is set to be $\SI{0.5}{m}$ for pedestrians and $\SI{2}{m}$ for vehicles. Street crossings are defined to be traversable by both pedestrian and motorized traffic participants. We, therefore, define the set of all pixels indicating a street crossing $\mathcal{S}_C$ as the intersection of pixels indicating pedestrian usage $\mathcal{S}_P$ and vehicle usage $\mathcal{S}_V$. More formally, we define $\mathcal{S}_C := \mathcal{S}_P \cap \mathcal{S}_V$. Obstacles are defined as objects extending substantially above the ground plane. To detect the ground plane, we segment the LiDAR point cloud using the pre-trained ground plane estimation network GroundNet~\cite{paigwar2020gndnet}. After projecting the segmented point cloud into each RGB image, we label each RGB image pixel located more than $\SI{20}{cm}$ above the ground plane as \textit{Obstacle}, following existing stixel-based approaches \cite{ramos2017detecting}. We denote the set of so-produced annotations for the surface classes \textit{Pedestrian}, \textit{Road}, \textit{Crossing}, and \textit{Obstacle} as dataset $\mathcal{D}_0$.

\begin{figure}
\centering
\includegraphics[width=\textwidth]{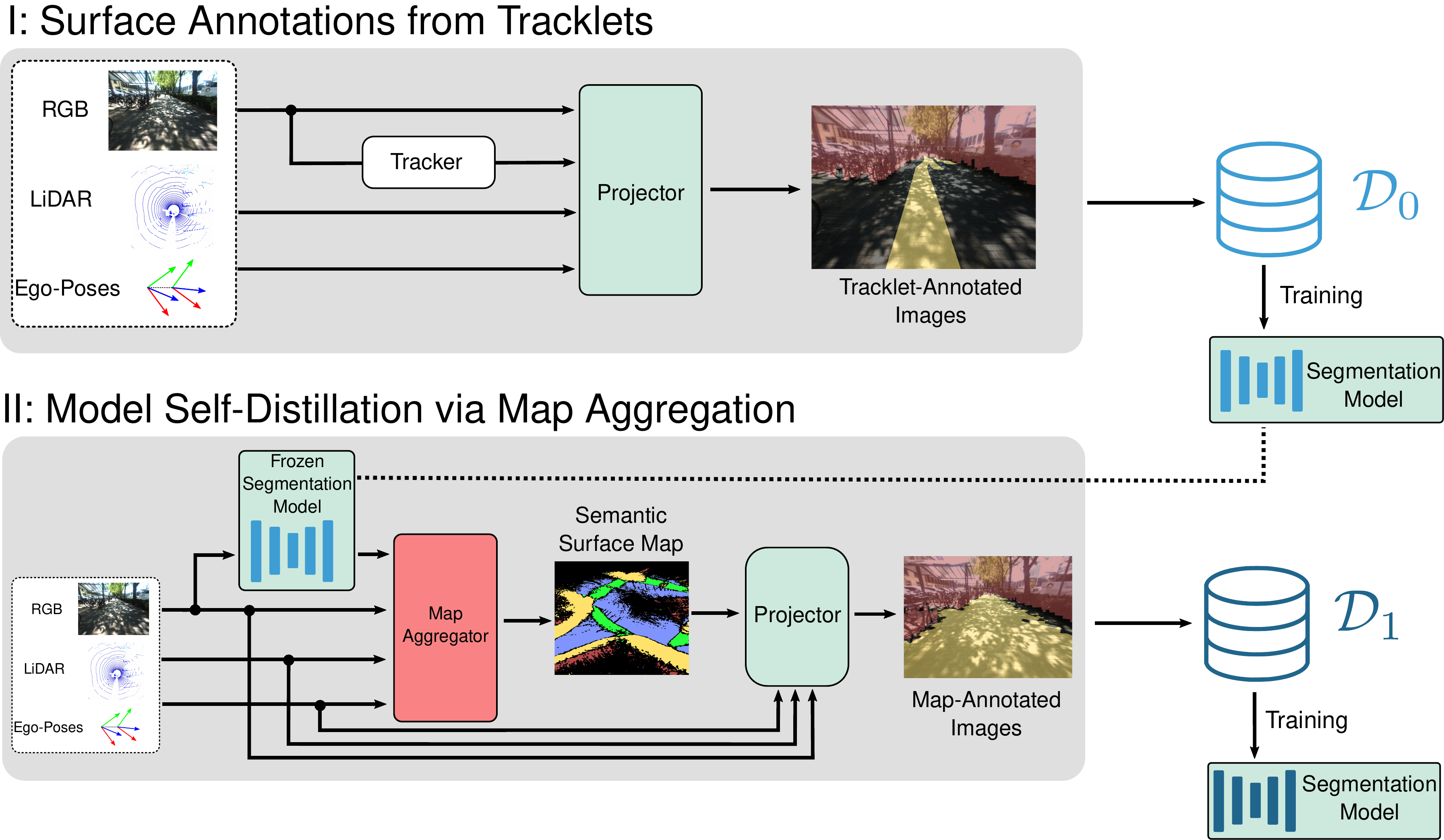}
\caption{Visualization of our automatic annotation pipeline. In step I, we leverage RGB images, LiDAR point clouds, ego-poses, and an object tracker to project the ego-poses and the observed tracklets into camera images, generating a sparsely-annotated semantic segmentation dataset $\mathcal{D}_0$. In step II, we use a frozen segmentation model trained on $\mathcal{D}_0$ to obtain semantic annotations and aggregate them into a global semantic surface map. Finally, we project this map into the camera images and obtain denser and more consistent annotations, denoted as $\mathcal{D}_1$.}
\label{fig:approach}
\end{figure}

\subsection{Surface Mapping and Self-Distillation via Aggregation and Reprojection}
\label{sec:mapping}

In addition to the aforementioned annotation procedure, we propose a novel self-distillation method for the semantic segmentation model. We argue that the inherent class prediction uncertainty in the segmentation model can be reduced by aggregating multiple predictions for a given patch of ground and re-training the model with these aggregated predictions. Prior works \cite{zhang2019your, gou2021knowledge} have shown how model self-distillation can help improve model performance. In this work we perform model self-distillation by spatially aggregating predictions in order to re-train the model on these aggregated predictions. Consider a surface patch $S^i$. Following similar formulations by \cite{asgharivaskasi2021active} and \cite{dewan2020deeptemporalseg}, we associate a belief $\mathbf{h}^i_t \in \mathbb{R}^K$ with $S^i$, containing the log odds vector of $S^i$ being of class $k$. We denote $K$ as the total number of considered classes. We collect all model predictions $\mathbf{p}_i \in \mathbb{R}^K$ that contain that patch of ground. In the beginning, the vector is initialized with a uniform distribution $\mathbf{h}_0$ over the classes and is updated according to the update rule $\mathbf{h}^i_{t + 1} = \mathbf{h}^i_t + (\mathbf{l}_t^i - \mathbf{h}_0)$, where $\mathbf{l}^i_t$ denotes the inverse observation model log odds:

\begin{equation}
\mathbf{l}_t^i = \Big[ \log \frac{p_t^i(k\!=\!1)}{1-p^i_t(k\!=\!1)} , \quad \log \frac{p_t^i(k\!=\!2)}{1-p_t^i(k\!=\!2)} , \quad \cdots, \quad \log \frac{p_t^i(k\!=\!K)}{1-p^i_t(k\!=\!K)} \Big]^T,
\end{equation}

and $p_t^i(k)$ denotes the model prediction for ground surface patch $S^i$ at time step $t$. After all belief updates have been executed, we transform the log-odds vector into class probabilities using the \texttt{softmax} function. We take the $\argmax$ over the probability vector to obtain the most likely surface class and annotate patch $S^i$ with that class. In order to obtain a dense surface representation suitable for training a segmentation model, we triangulate all surface patch center points and create a triangular mesh of ground surfaces. As a post-processing step, we smooth the surface mesh using the Taubin filter \cite{taubin1995curve}. To generate training data for the segmentation model, we again use the 3D poses of the camera and project the semantic surface mesh back into the camera RGB images as dense semantic annotations. Due to the larger spatial extent of the surface map compared to the tracklets, we can significantly increase the number of annotated pixels in each image. We denote the set of so-produced annotations as dataset $\mathcal{D}_1$.

\subsection{Model Training}

The aforementioned annotation scheme labels pixels that are associated with obstacles or have been traversed either by the robot or by other traffic participants. All other pixels in the images are assigned the label \textit{Unknown}. We pose the ground segmentation task as a segmentation task with sparse label supervision, where only a subset of the pixels in each image has annotations available. As the model architecture, we use the DeepLabv3+ model architecture~\citep{chen2018encoder}. We use a standard cross-entropy loss for all non-\textit{Unknown} image pixels. \textit{Unknown} pixels are ignored during training. 

\section{Dataset}
\label{sec:dataset}

\begin{figure}
\centering
\includegraphics[width=14cm]{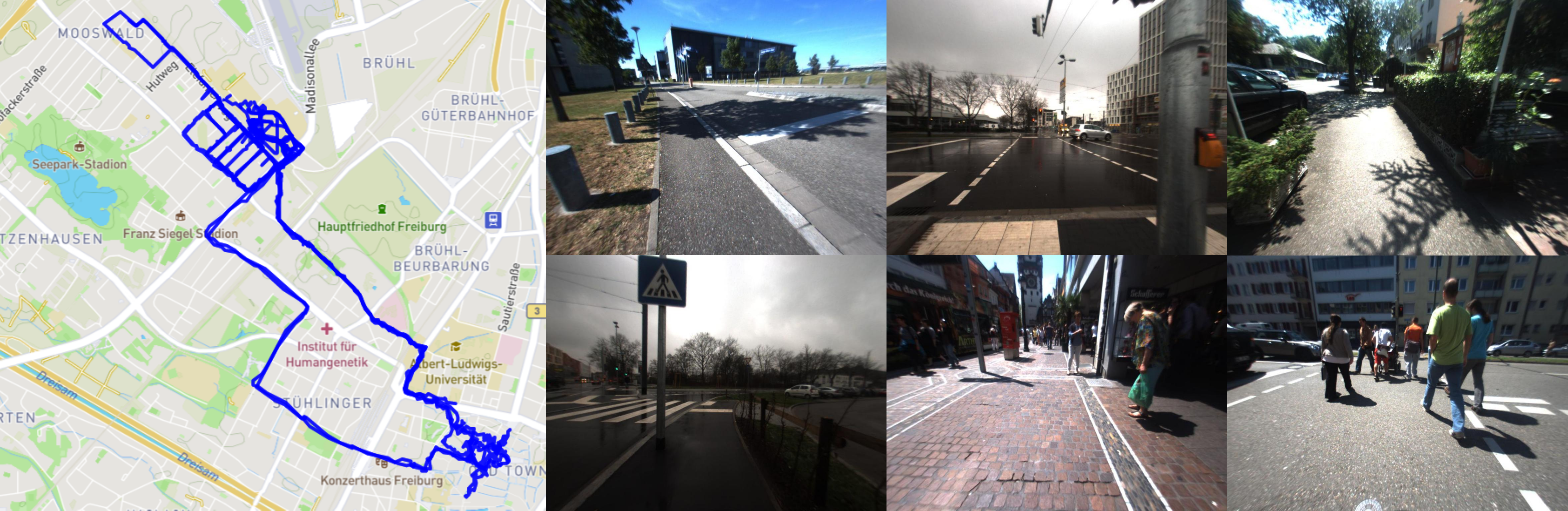}
\caption{Ego-trajectories (left, blue color) and camera images from our \textit{Freiburg Pedestrian Scenes} dataset. The dataset features a wide range of urban environments including busy streets, pedestrian areas, and road crossings with varying weather conditions.}
\label{fig:dataset}
\end{figure}

We present the \textit{Freiburg Pedestrian Scenes} dataset recorded with our robot platform. During each data collection run, the robot is teleoperated through semi-structured urban environments and moves alongside pedestrians on sidewalks, pedestrian areas, and street crossings. Each data collection run consists of time-synchronized sensor measurements from a Bumblebee Stereo RGB camera, a Velodyne HDL 32-beam rotating LiDAR scanner, an IMU, and a GPS/GNSS receiver. Furthermore, we provide Graph-SLAM poses~\cite{grisetti2010tutorial}. In total, the dataset comprises 15 highly diverse and challenging urban scenes. The data collection runs cover a wide range of illumination conditions, weather conditions, and structural diversity. Figure \ref{fig:dataset} illustrates exemplary RGB images and the recording locations. The dataset was recorded over the course of multiple years in the city of Freiburg, Germany. The dataset key statistics are listed in Tab. \ref{tab:dataset}. To evaluate our approach, we manually annotated 50 ego-view RGB images from five data collection runs not included in the training set. We also hand-annotated large sections of the traversed areas with a semantic BEV map in order to be able to compare aggregated and ground-truth maps. Exemplary visualizations of this map are visualized in Fig. \ref{fig:bev}.

\begin{wraptable}{r}{8cm}
\normalsize
\caption{\textit{Freiburg Pedestrian Scenes} dataset details}
\begin{tabular}{|p{2.4cm}|p{1.6cm}|p{2.5cm}|}
Modality & Quantity & Frequency [Hz] \\
 \midrule
Stereo RGB    & 260k   & 5   \\
LiDAR         & 490k   & 9   \\
IMU           & 4.2M   & 100 \\
GPS           & 490k   & 9  \\
SLAM poses    & 49k    & 1 \\
\hline
Map annotations    & 112523 \SI{}{m^2}  & - 
\label{tab:dataset}
\end{tabular}
\end{wraptable}

\section{Experimental Results}
\label{sec:experiments}

\begin{table*}
\centering
\caption{\rebuttal{Model performances when trained on the \textit{Freiburg Pedestrian Scenes} / Vistas datasets and evaluated on the \textit{Freiburg Pedestrian Scenes} dataset}. We denote the IoU values in \%.}
\begin{tabular}{|p{3cm}|>{\raggedleft\arraybackslash}p{1.3cm}|>{\raggedleft\arraybackslash}p{2.0cm}|>{\raggedleft\arraybackslash}p{1.8cm}|>{\raggedleft\arraybackslash}p{1.8cm}|>{\raggedleft\arraybackslash}p{0.8cm}}
Annotation Source & \crule[road]{0.2cm}{0.2cm} Road	& \crule[pedestrian]{0.2cm}{0.2cm} Pedestrian & \crule[crossing]{0.2cm}{0.2cm} Crossing & \crule[obstacle]{0.2cm}{0.2cm} Obstacle & Mean \\
\hline
Mapillary Vistas \cite{neuhold2017mapillary}   & 12.1  &  20.0  &  0.5  &  \textbf{89.2}  & 30.4 \\
Ego   &  0 & 37.3 & 0 & 85.8  &  30.8 \\
Ego + Tracklets   & 35.9 &  67.5 & 43.4 & 88.3 & 58.8 \\
Map Reprojection  & \textbf{38.4} & \textbf{69.2} & \textbf{48.2} & 85.9 & \textbf{60.4} 
\end{tabular}
\label{tab:results}
\vspace{-5mm}
\end{table*}

\begin{figure}
\centering
\footnotesize
\setlength{\tabcolsep}{0.0cm}
    \begin{tabular}{P{2.8cm}P{2.8cm}P{2.8cm}P{2.8cm}P{2.8cm}}
    RGB Input & Ground Truth & \thead{Trained on \\ Vistas\cite{neuhold2017mapillary}}   & \thead{Trained on \\ Ego-Trajectory} & \thead{Trained on \\  Map Reprojection} \\ [0.1cm] \\

          \includegraphics[width=0.95\linewidth]{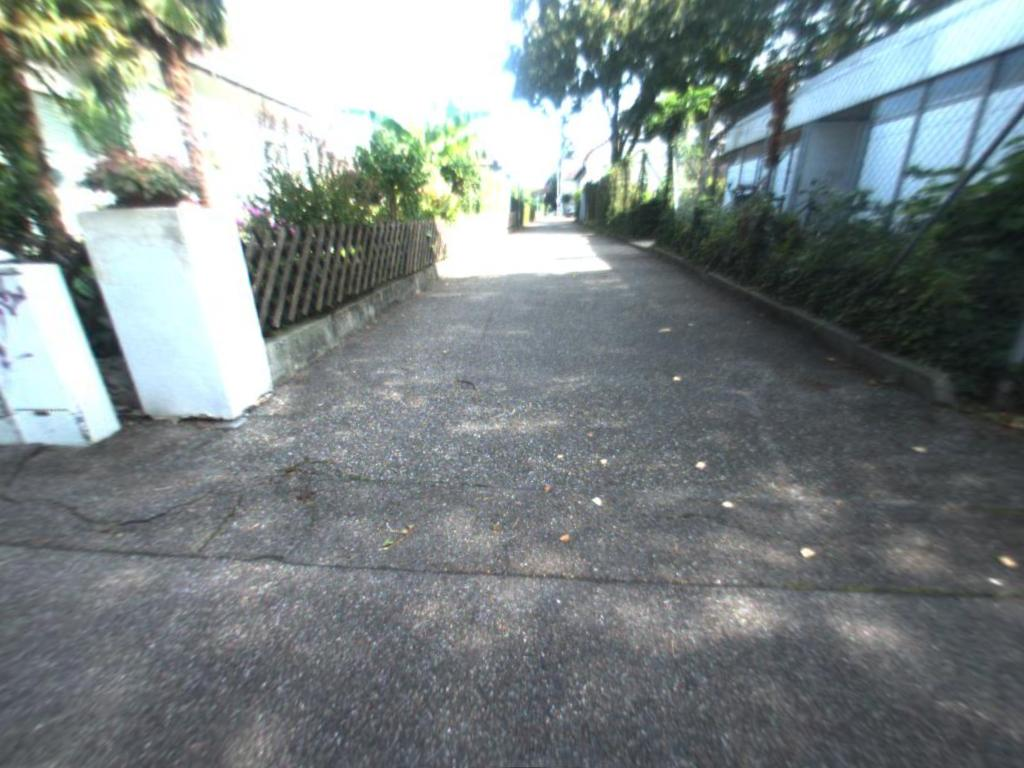} & 
          \includegraphics[width=0.95\linewidth]{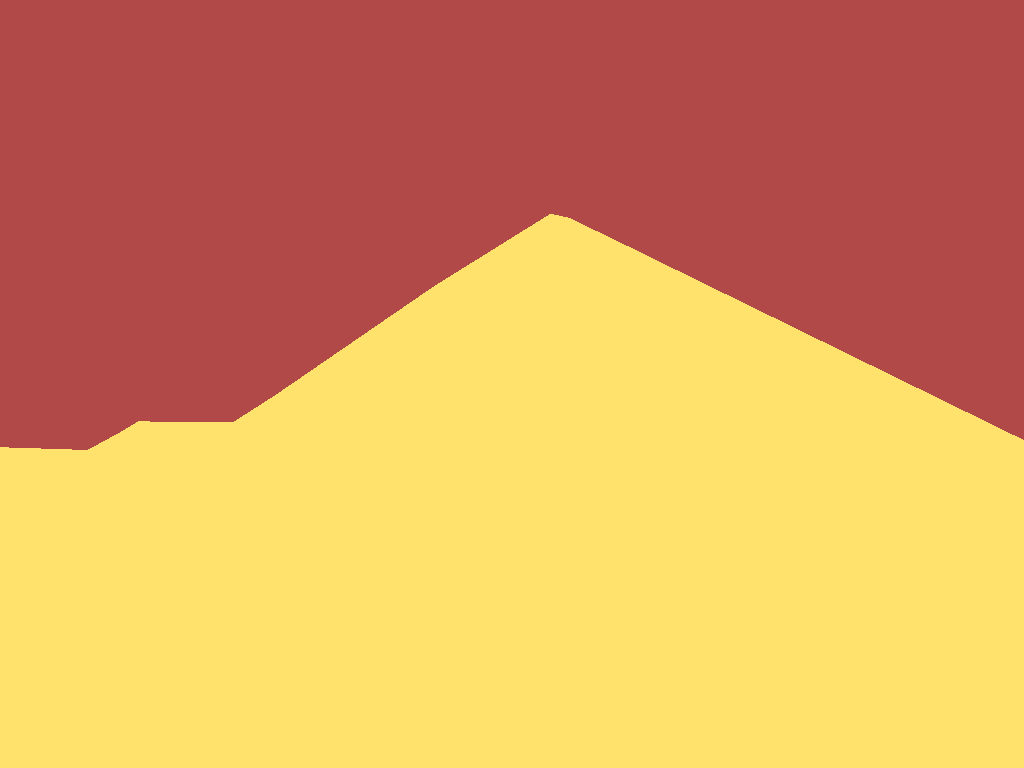} & \includegraphics[width=0.95\linewidth]{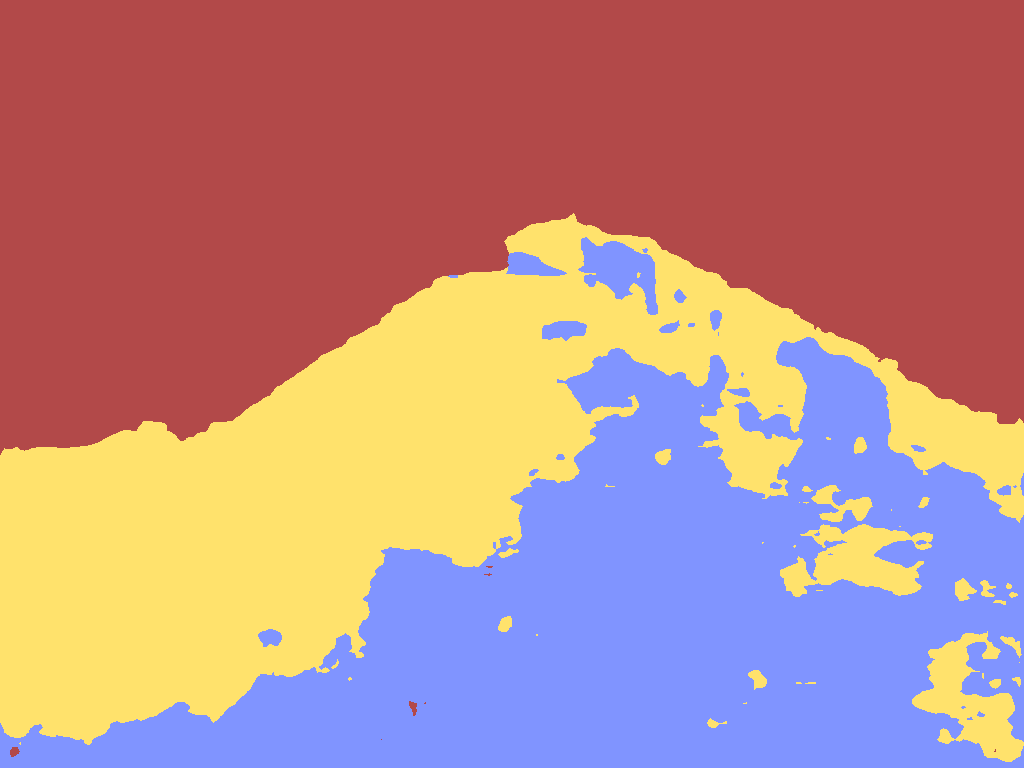} &  \includegraphics[width=0.95\linewidth]{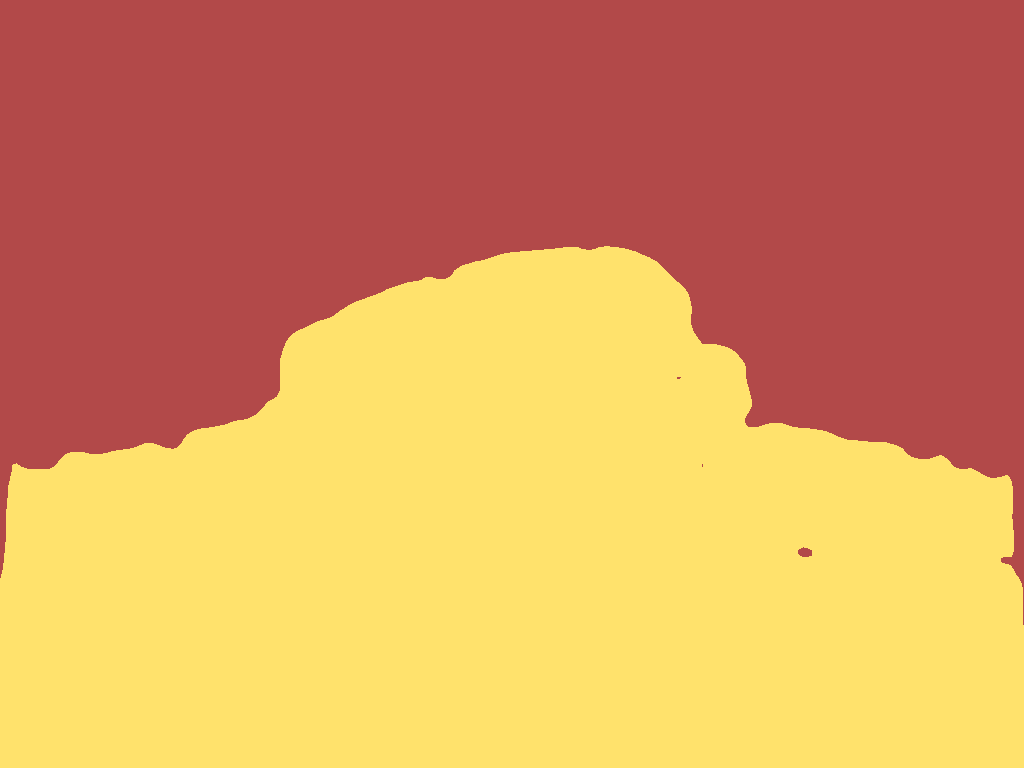} & \includegraphics[width=0.95\linewidth]{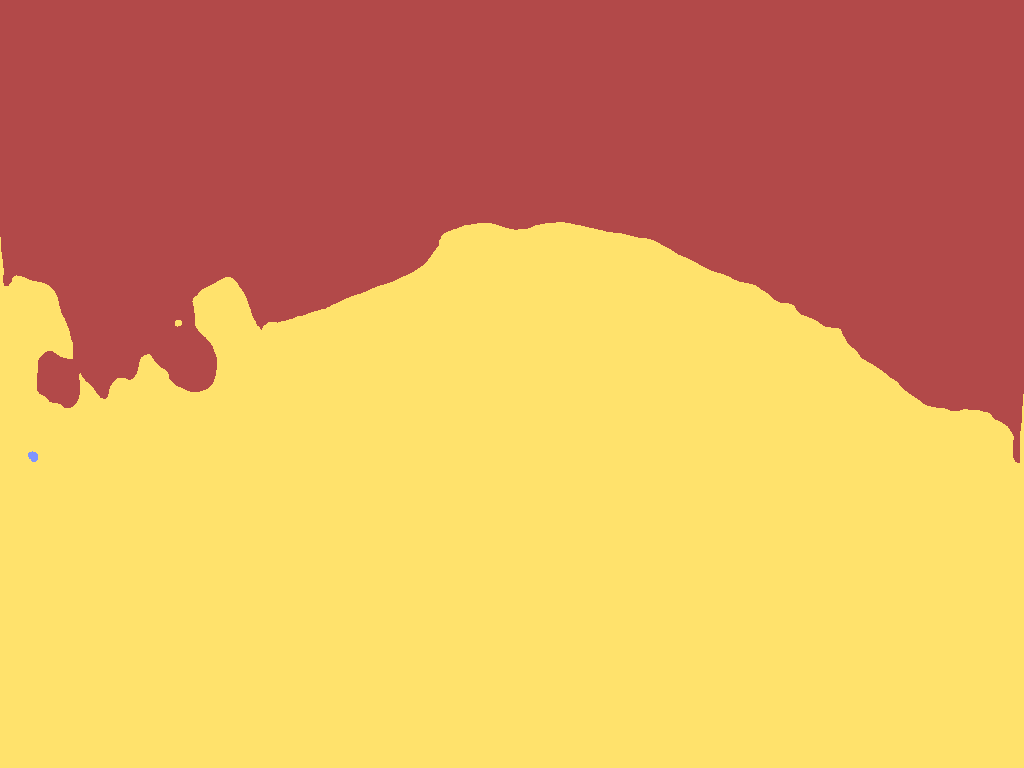} \\

          \includegraphics[width=0.95\linewidth]{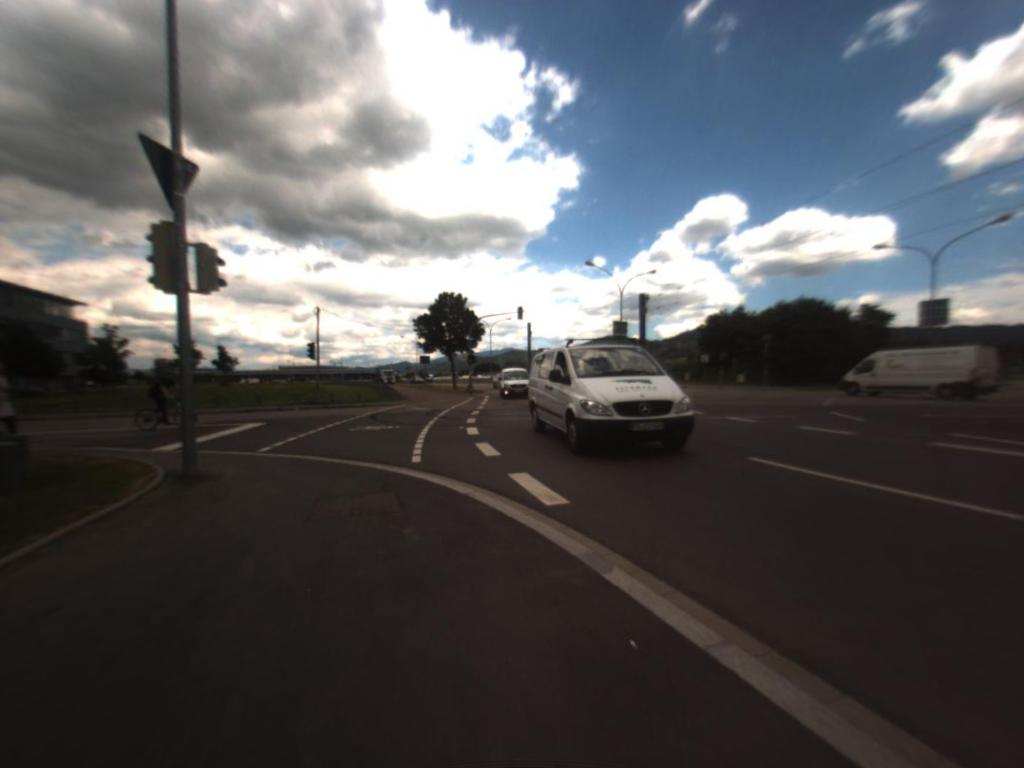} & 
          \includegraphics[width=0.95\linewidth]{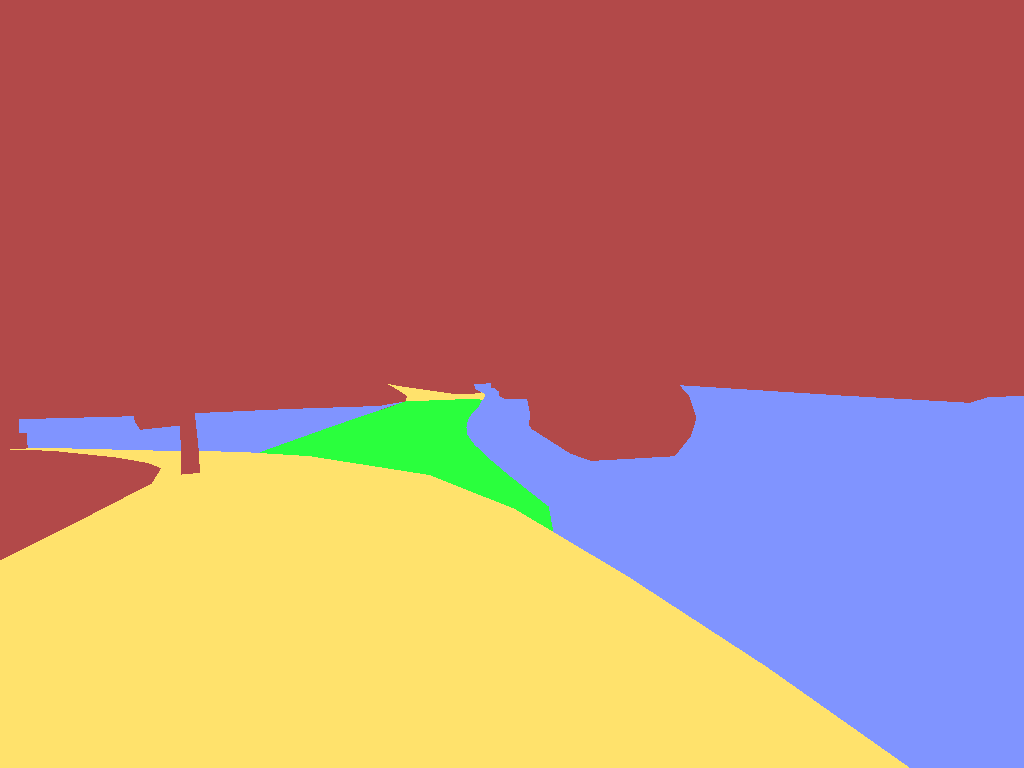} & \includegraphics[width=0.95\linewidth]{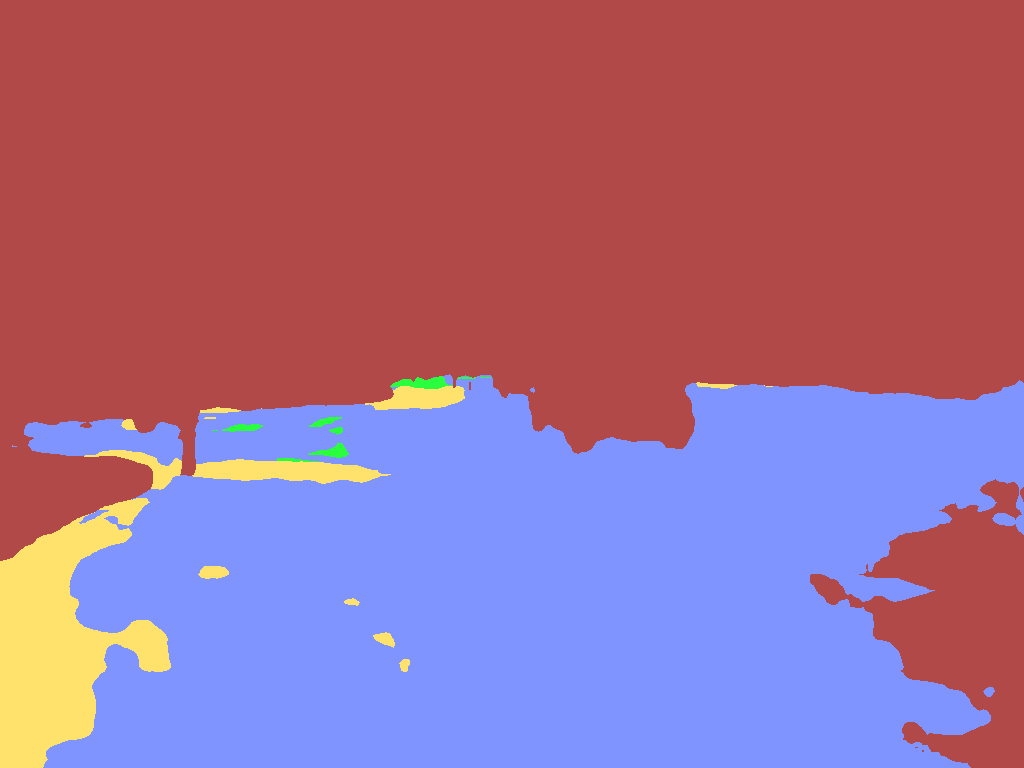} &  \includegraphics[width=0.95\linewidth]{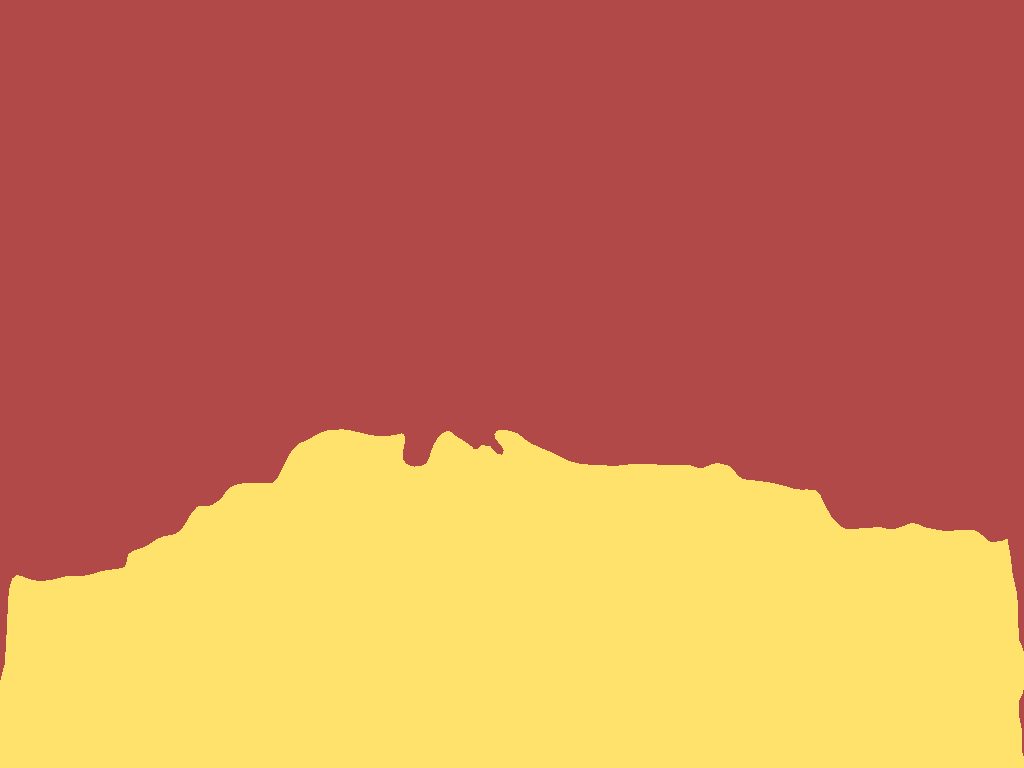} & \includegraphics[width=0.95\linewidth]{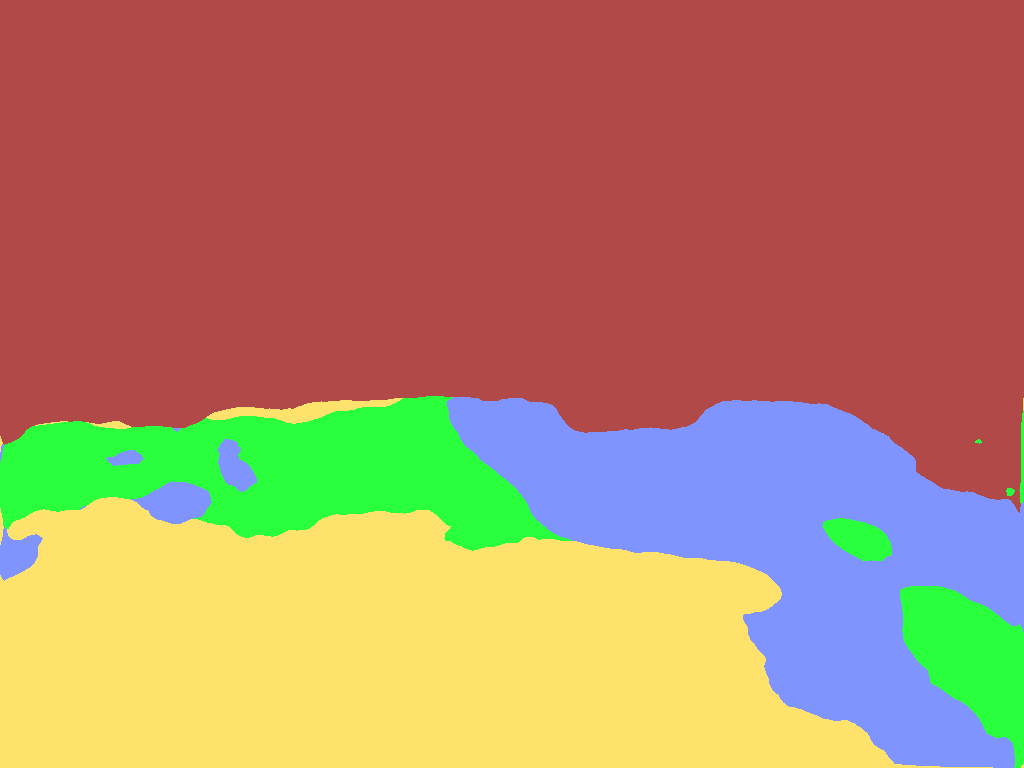} \\

          \includegraphics[width=0.95\linewidth]{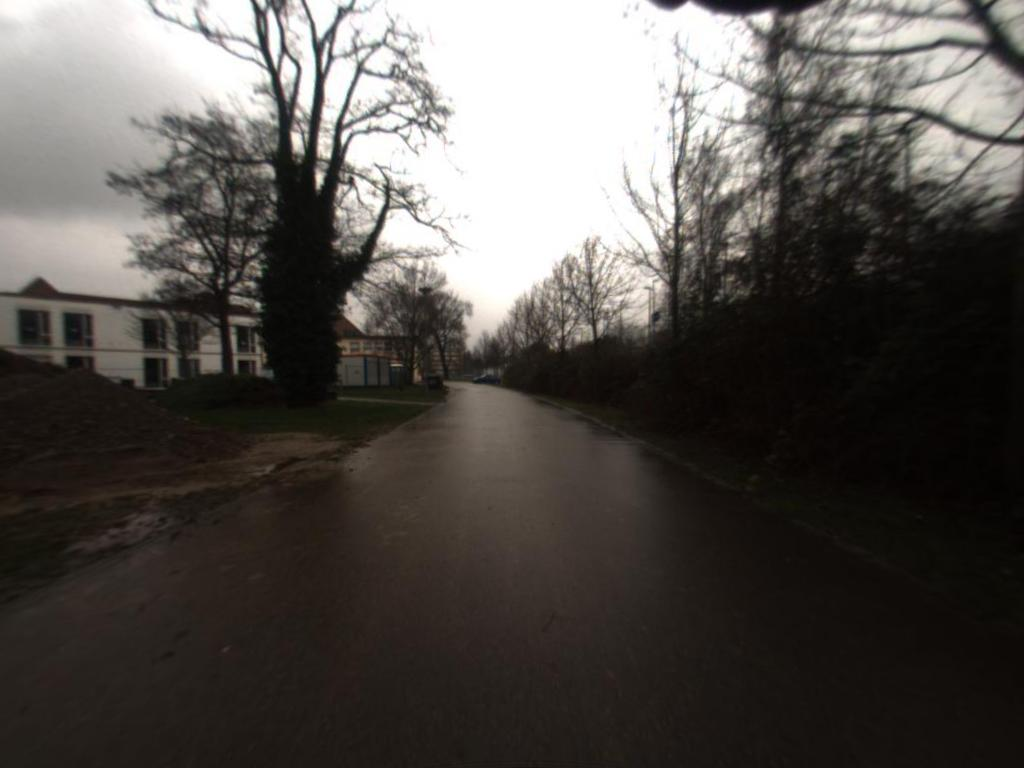} & 
          \includegraphics[width=0.95\linewidth]{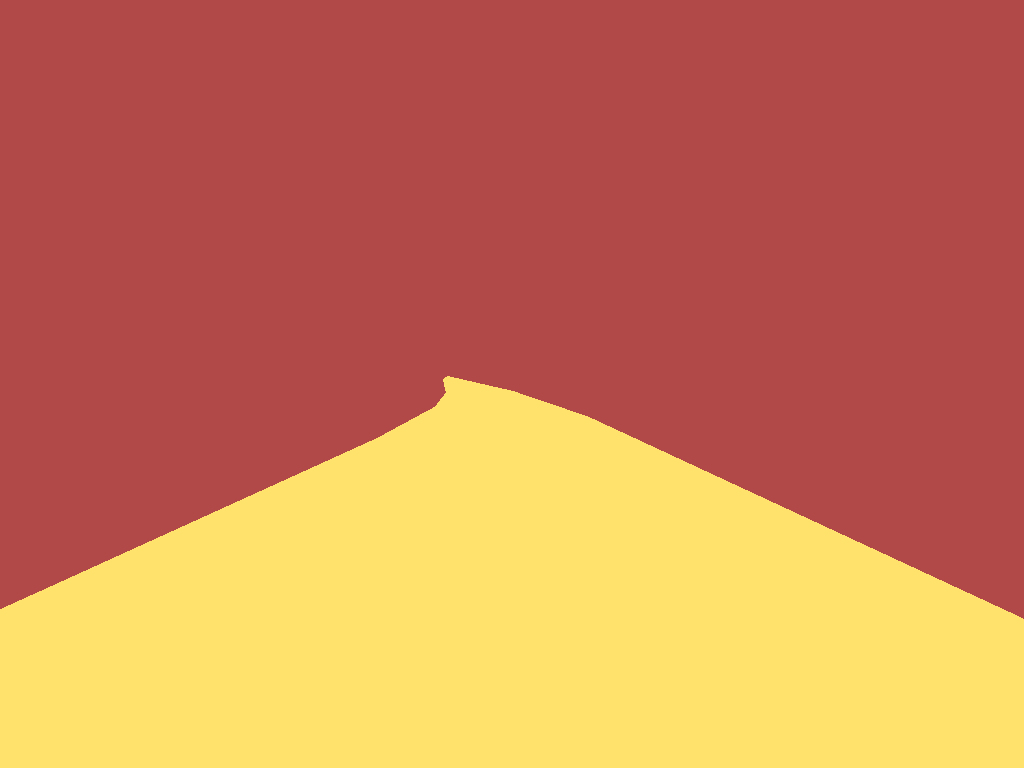} & \includegraphics[width=0.95\linewidth]{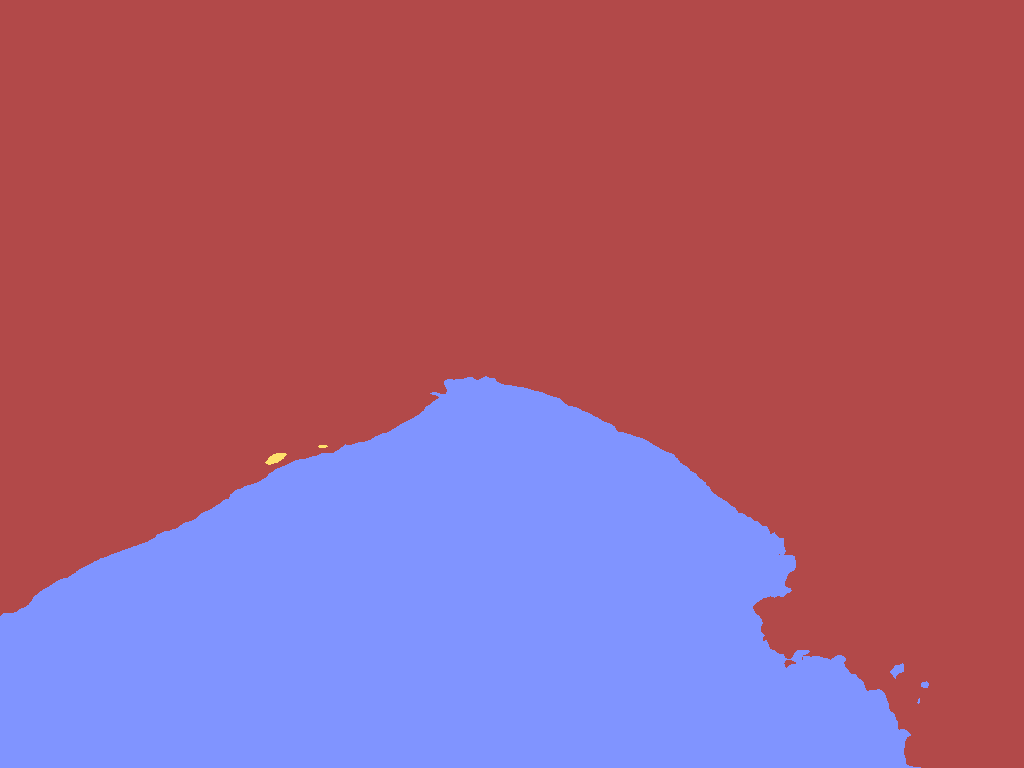} &  \includegraphics[width=0.95\linewidth]{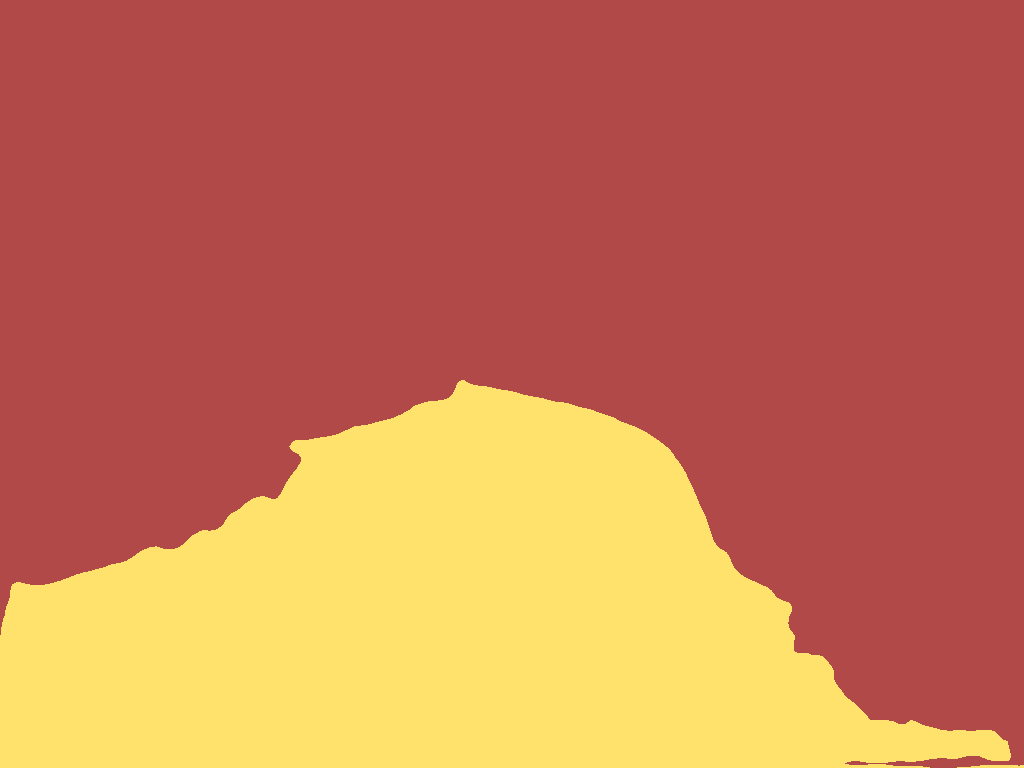} & \includegraphics[width=0.95\linewidth]{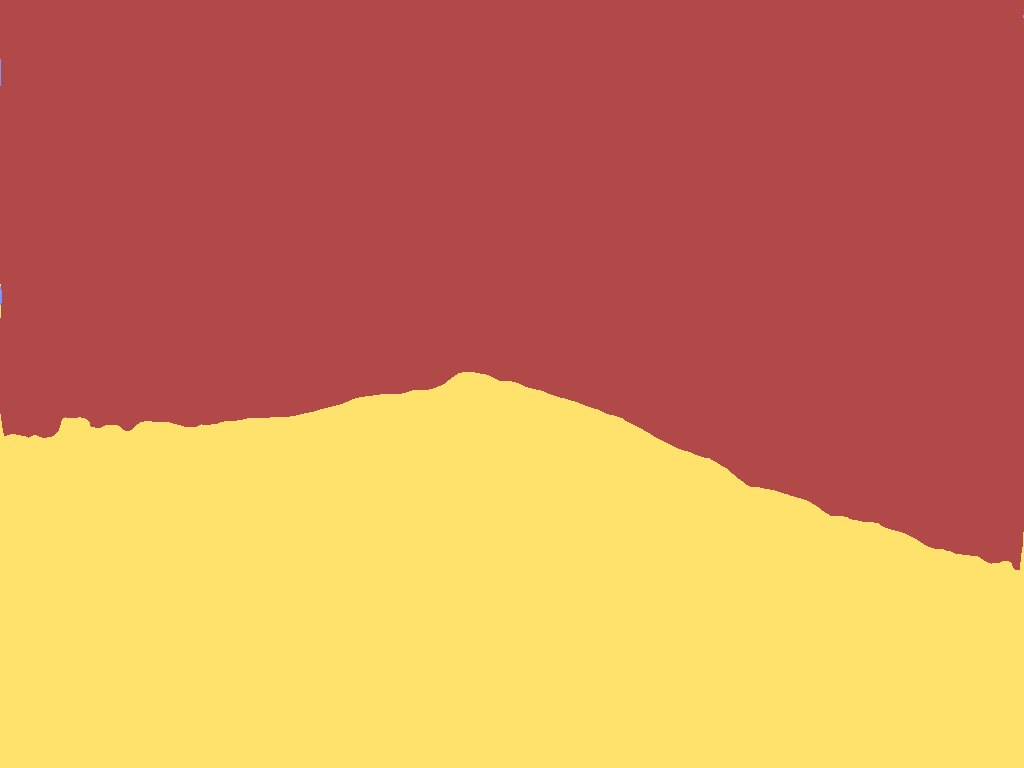} \\

          \includegraphics[width=0.95\linewidth]{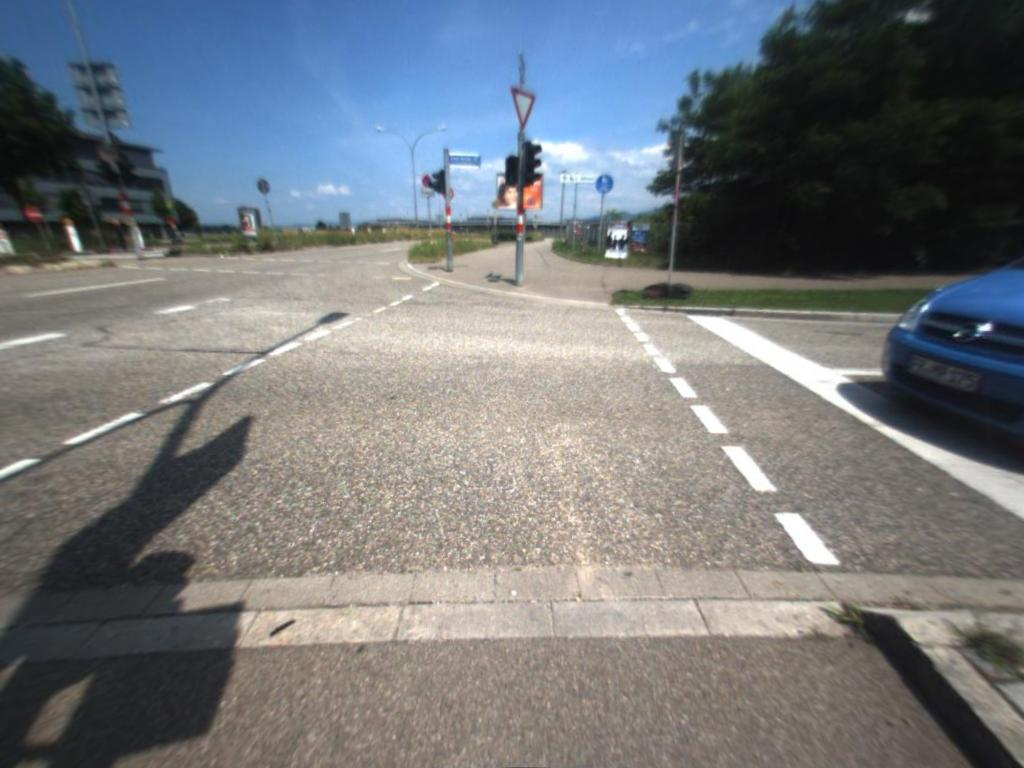} & 
          \includegraphics[width=0.95\linewidth]{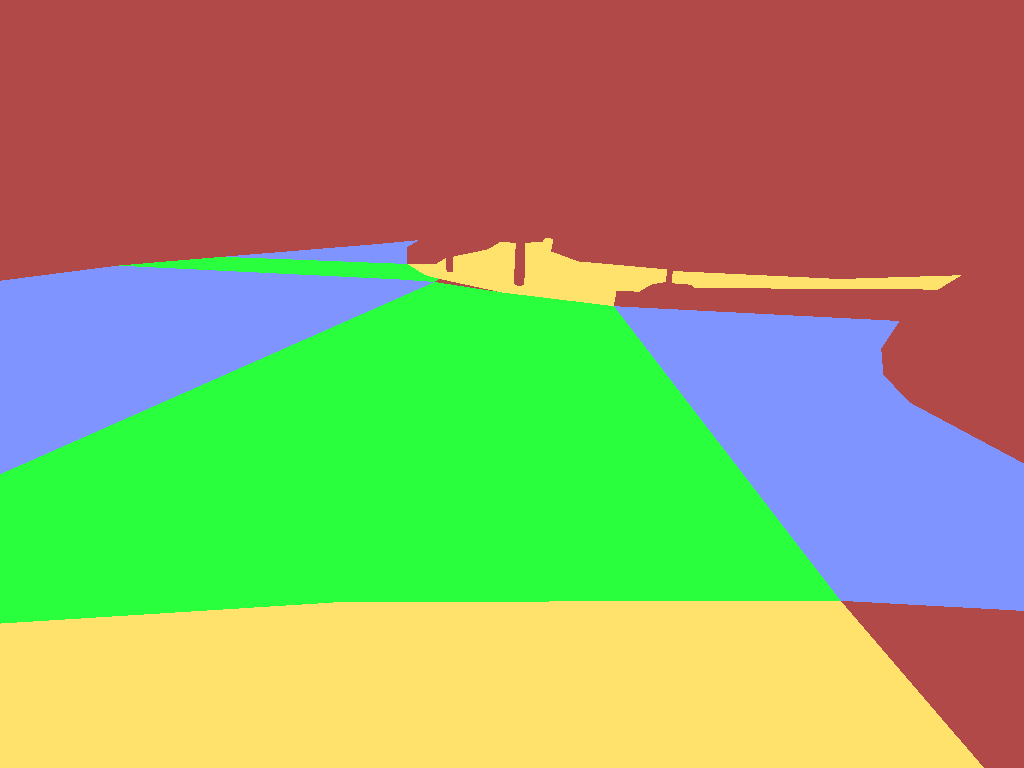} & \includegraphics[width=0.95\linewidth]{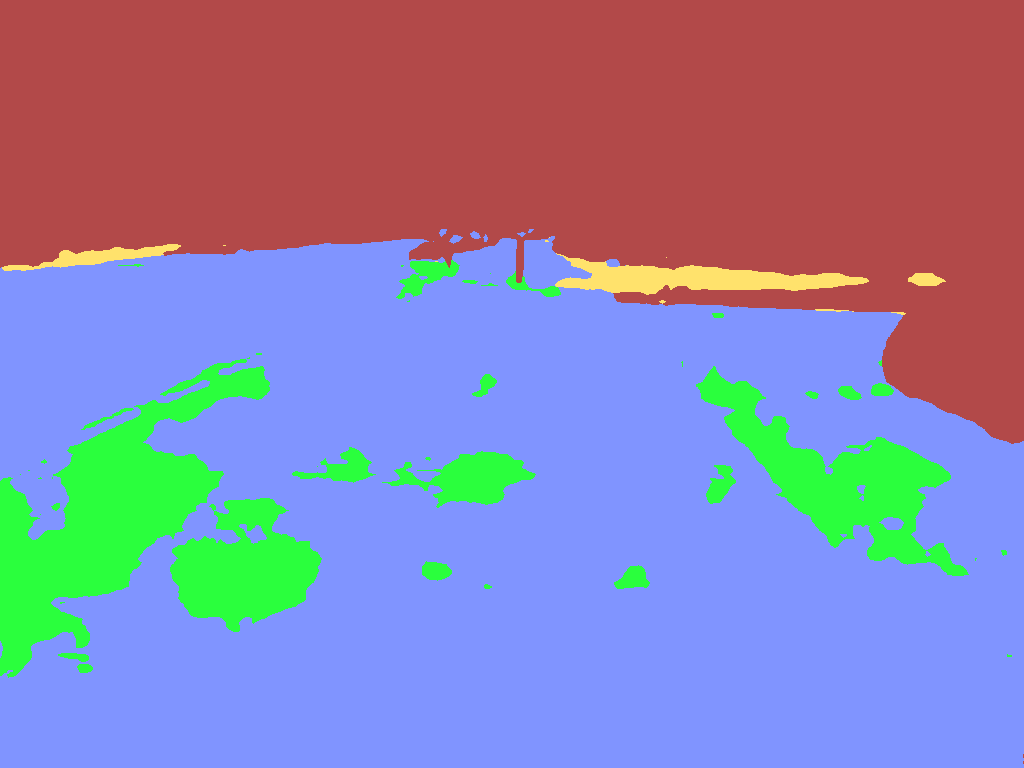} &  \includegraphics[width=0.95\linewidth]{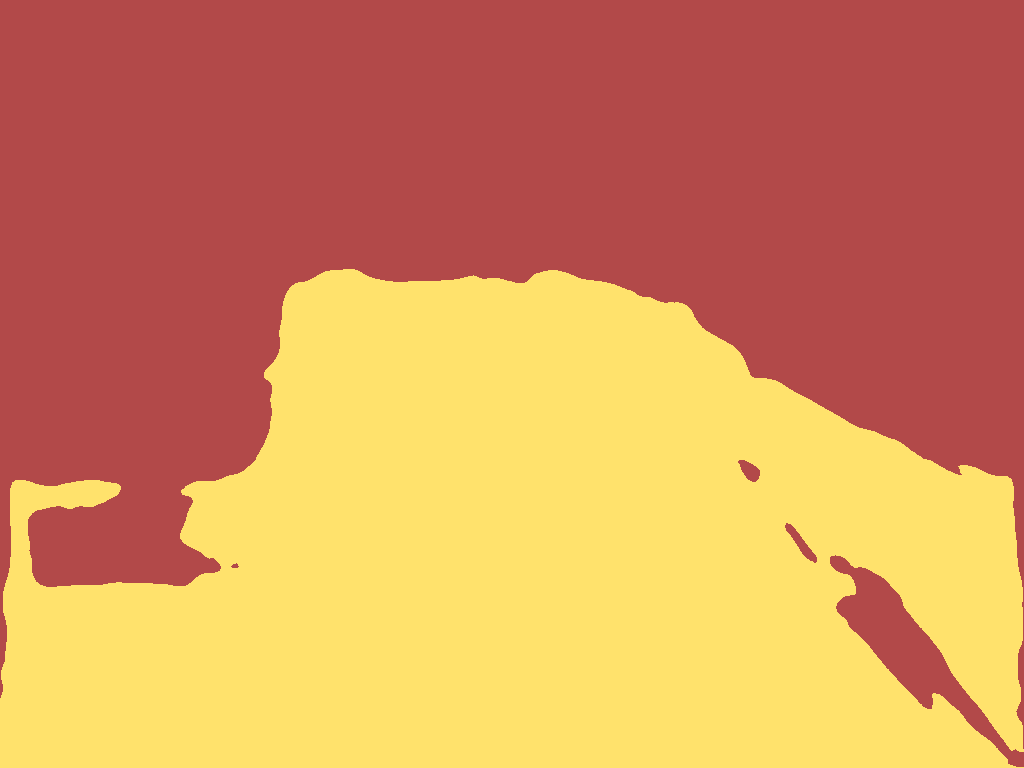} & \includegraphics[width=0.95\linewidth]{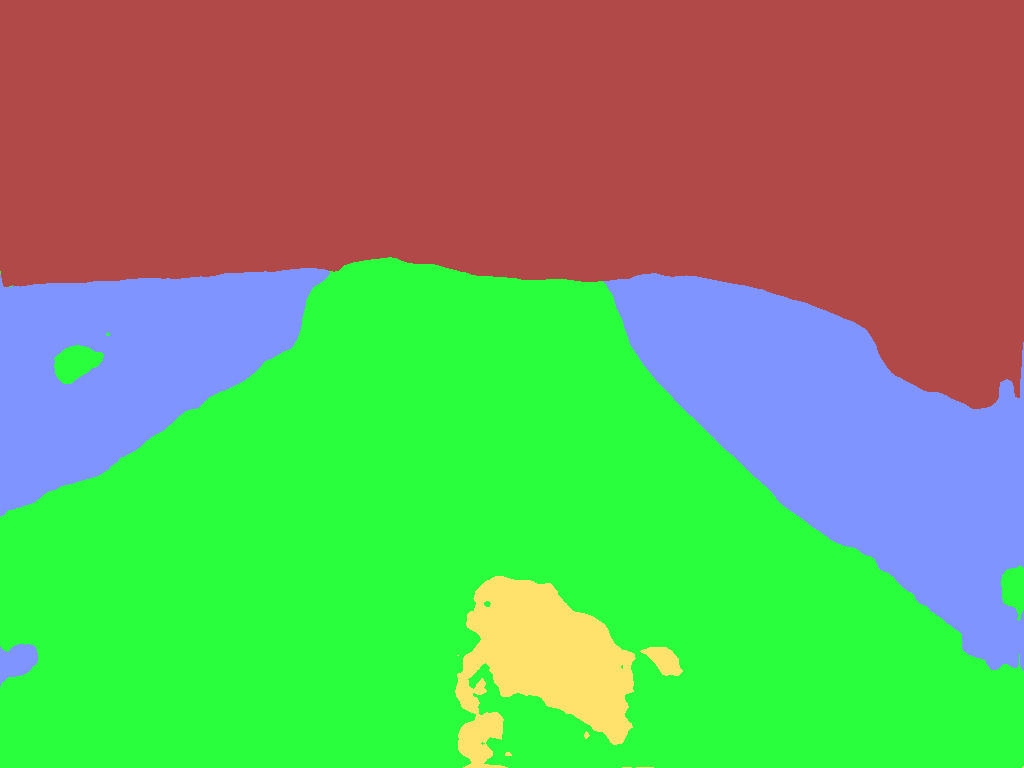} \\

          \includegraphics[width=0.95\linewidth]{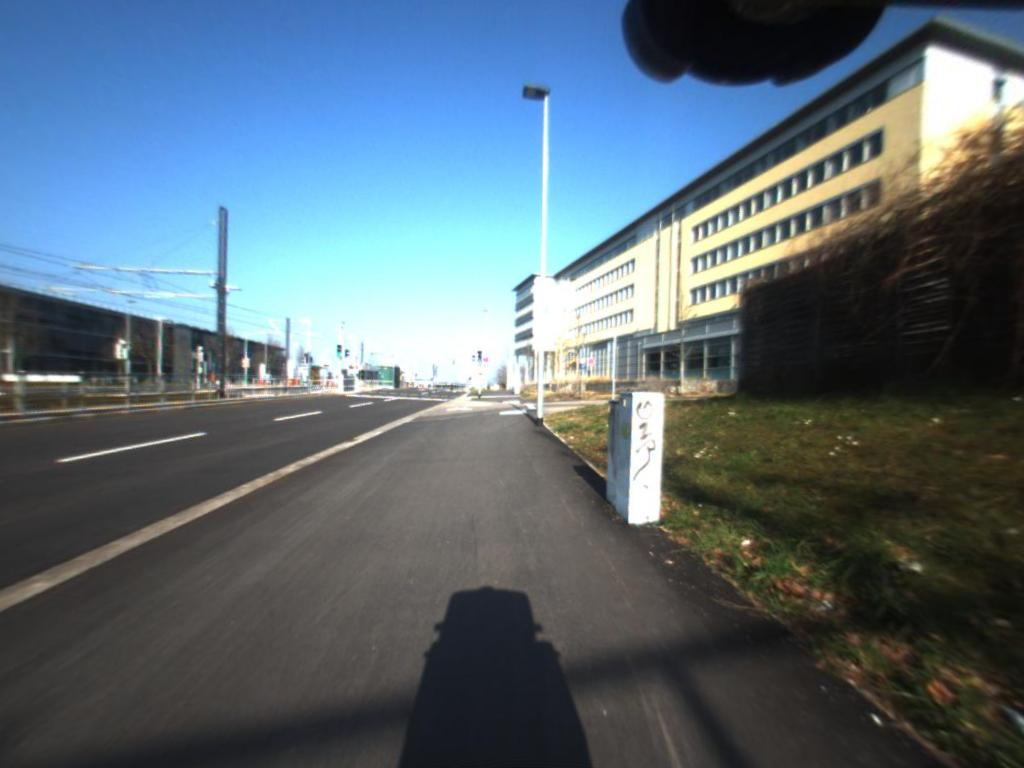} & 
          \includegraphics[width=0.95\linewidth]{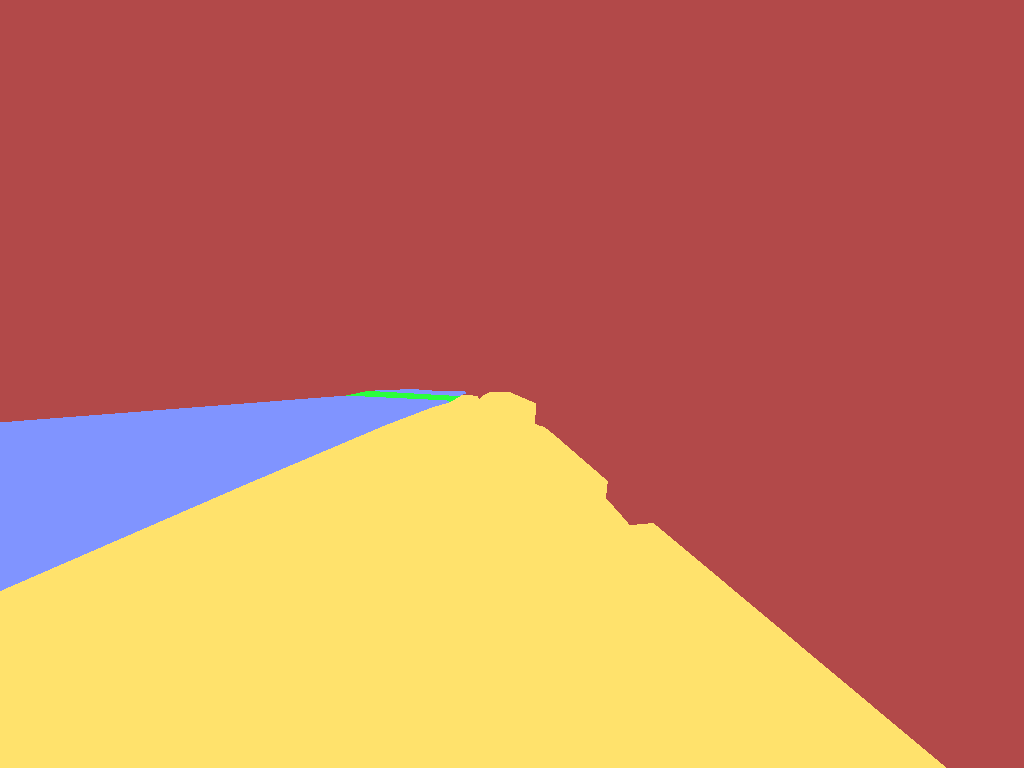} & \includegraphics[width=0.95\linewidth]{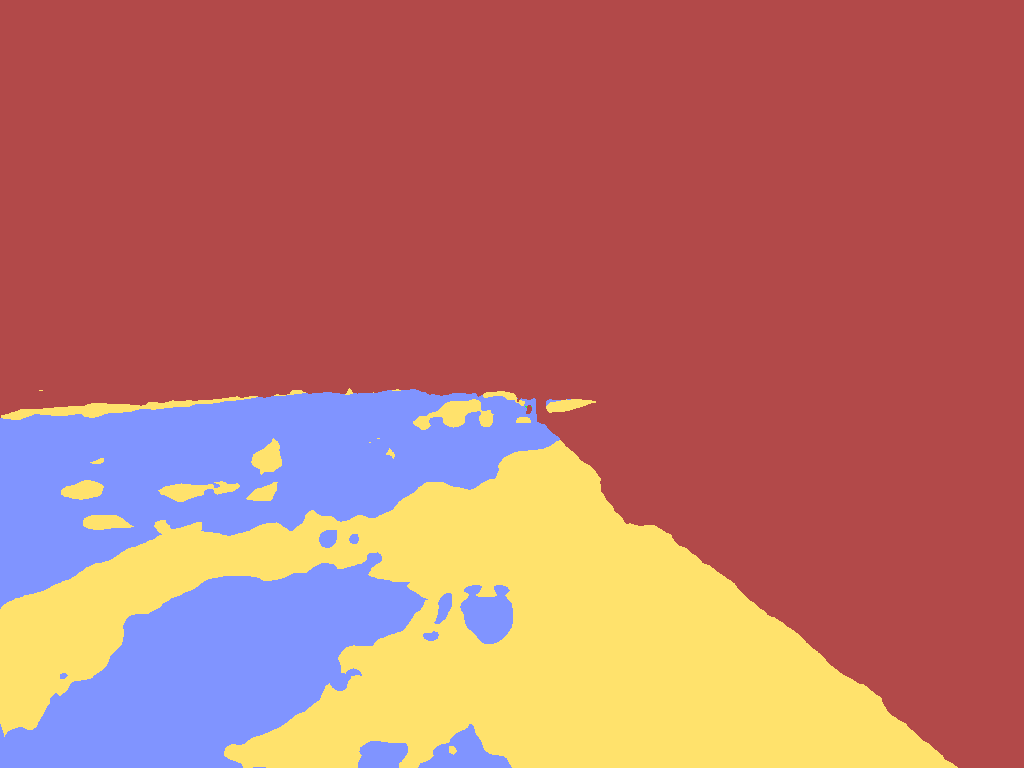} &  \includegraphics[width=0.95\linewidth]{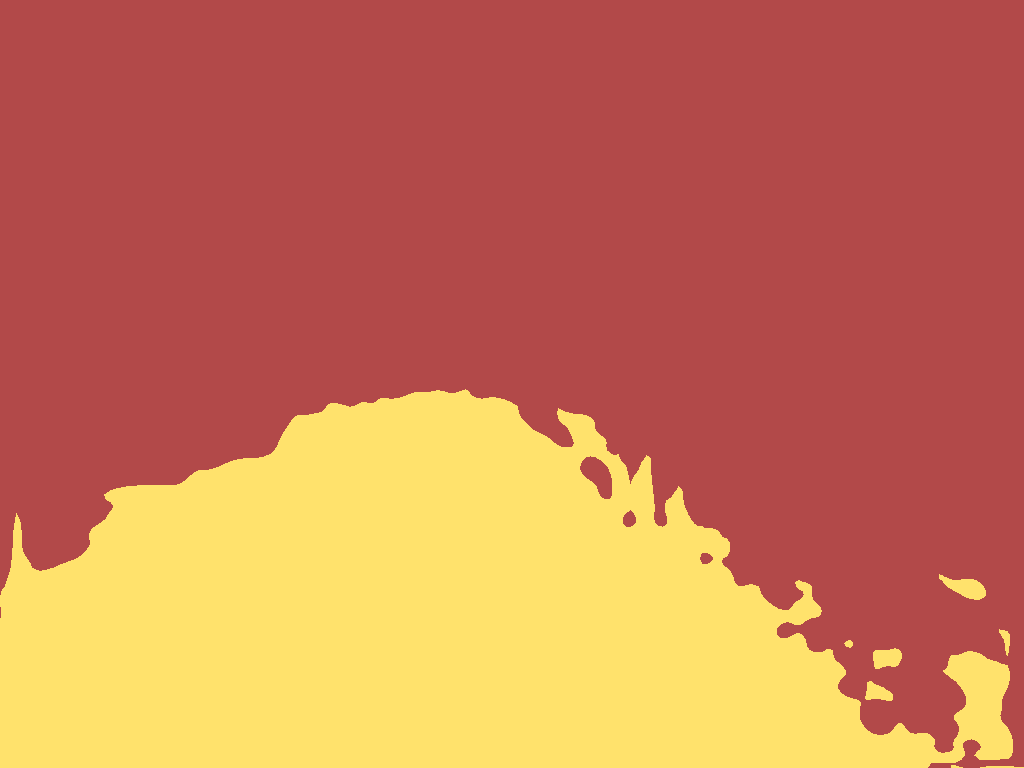} & \includegraphics[width=0.95\linewidth]{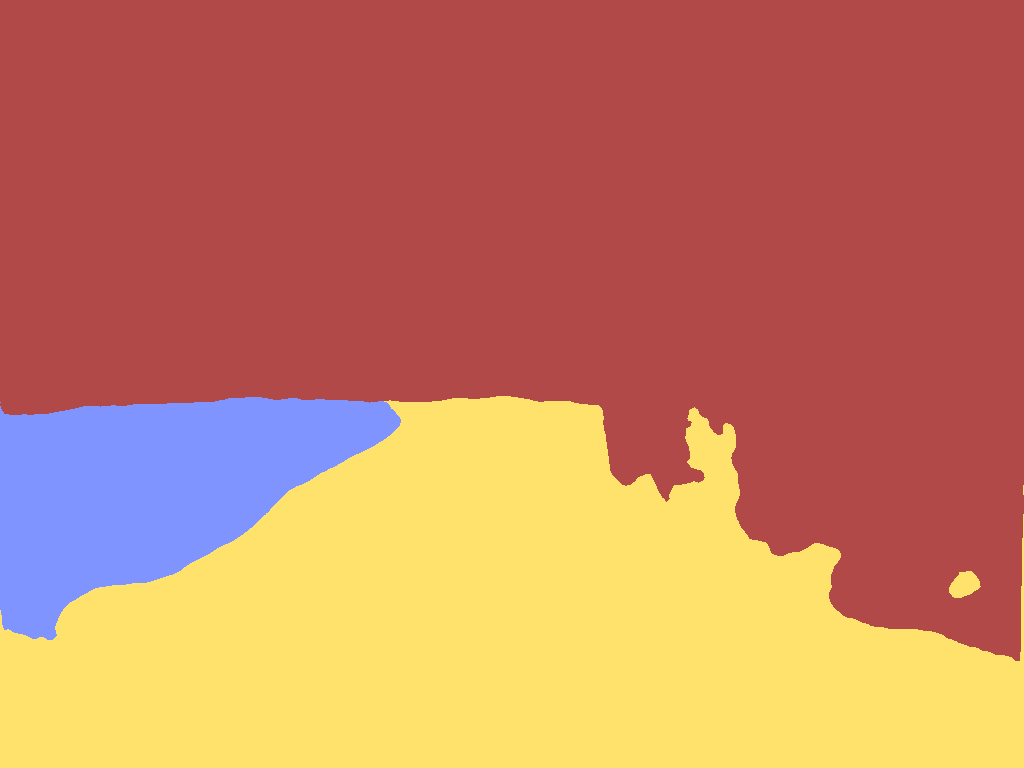} \\

          \arrayrulecolor{black!50}\specialrule{2pt}{2\jot}{1pc}

          \includegraphics[width=0.95\linewidth]{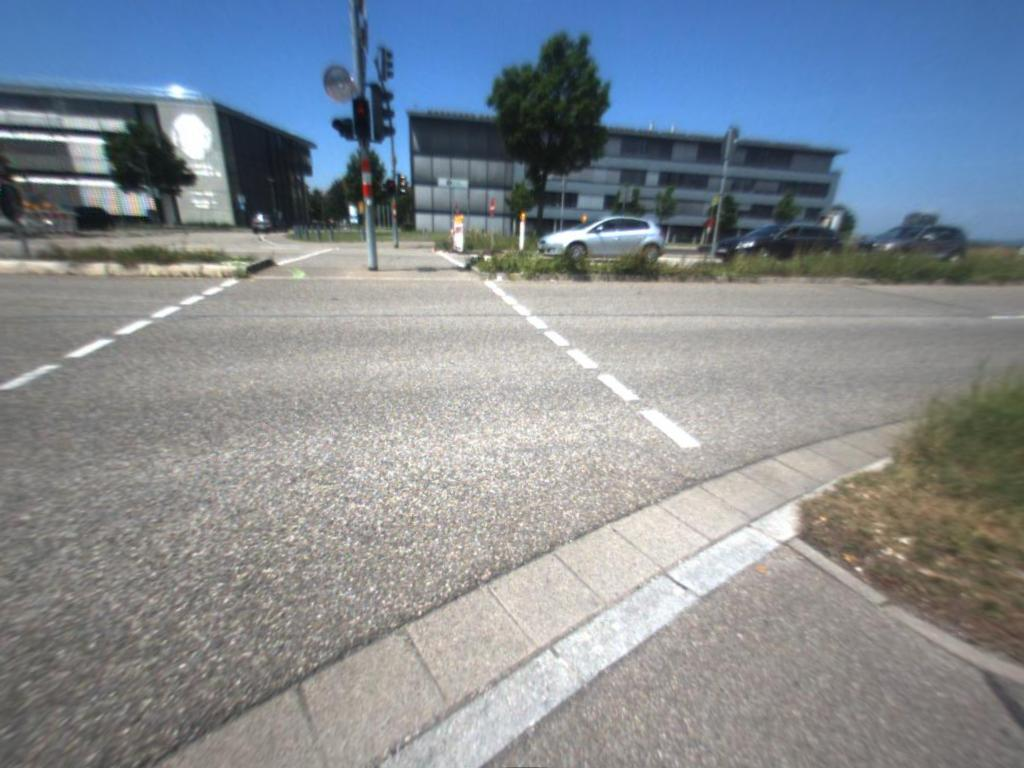} & 
          \includegraphics[width=0.95\linewidth]{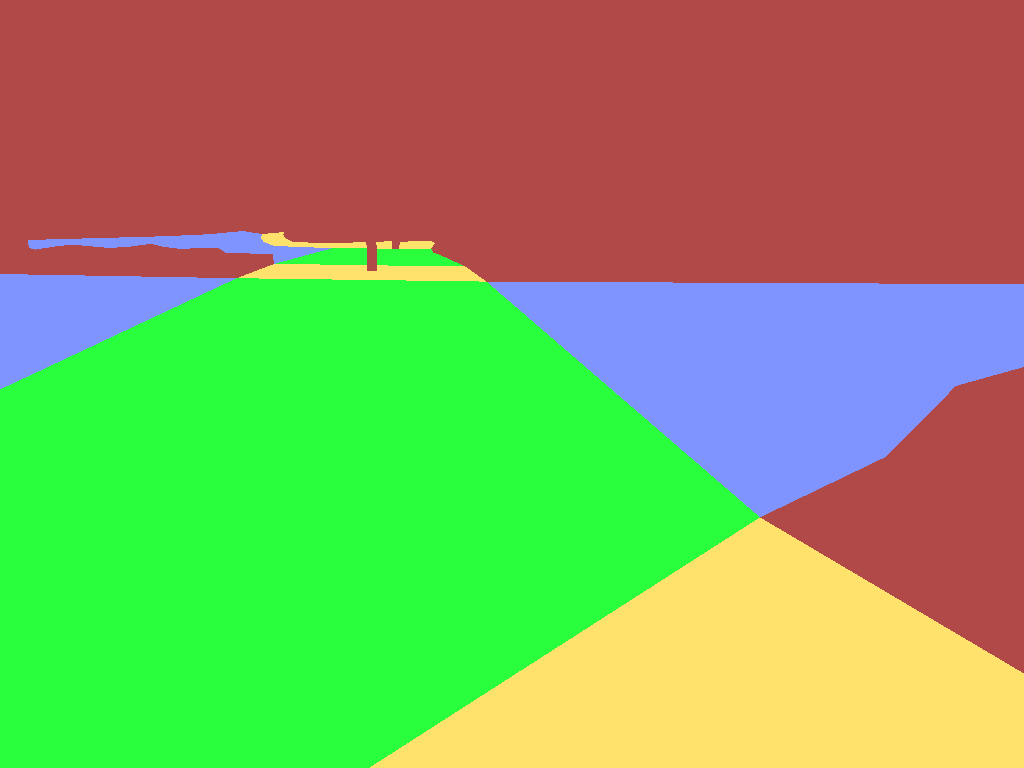} & \includegraphics[width=0.95\linewidth]{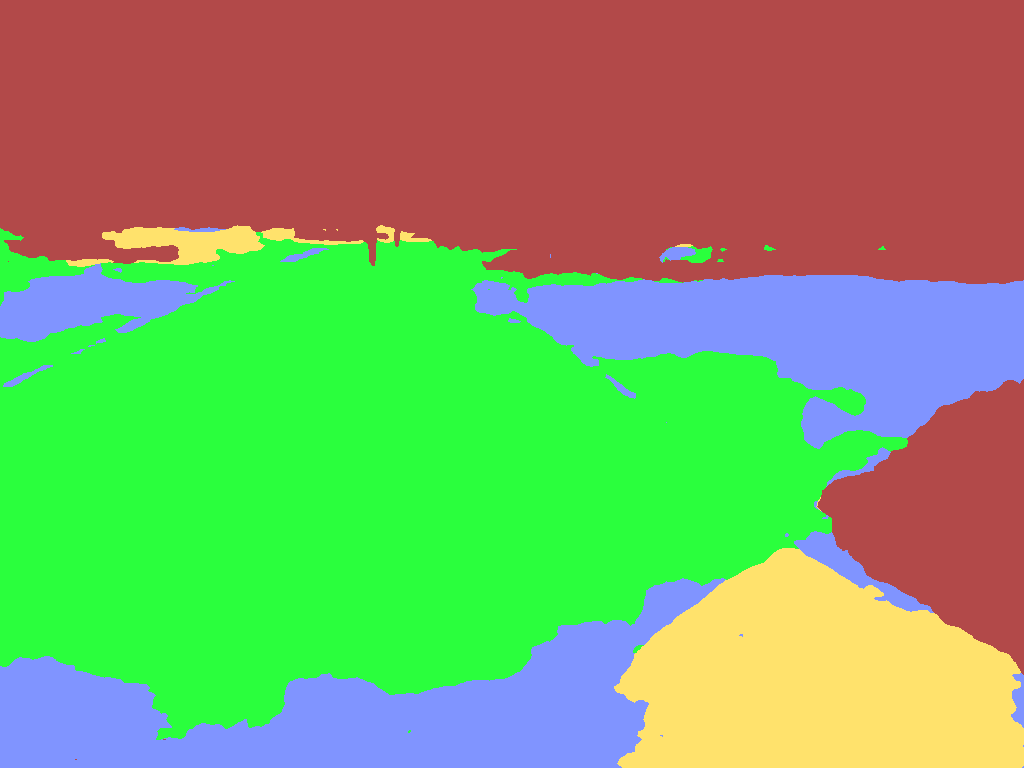} &  \includegraphics[width=0.95\linewidth]{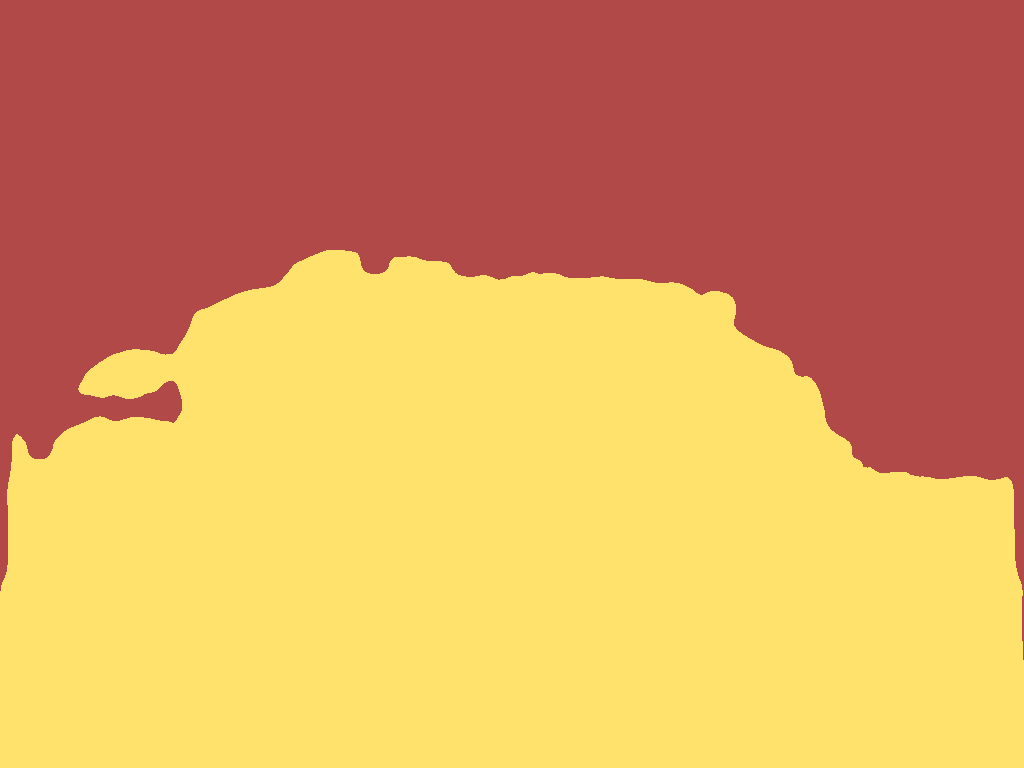} & \includegraphics[width=0.95\linewidth]{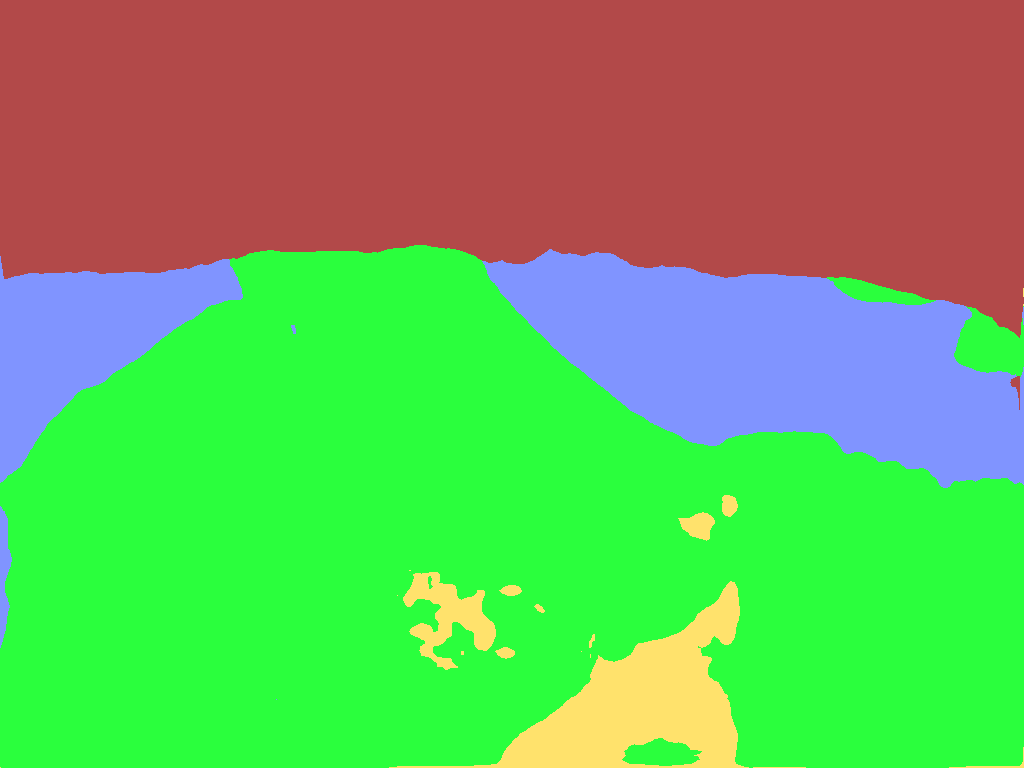} \\

          \includegraphics[width=0.95\linewidth]{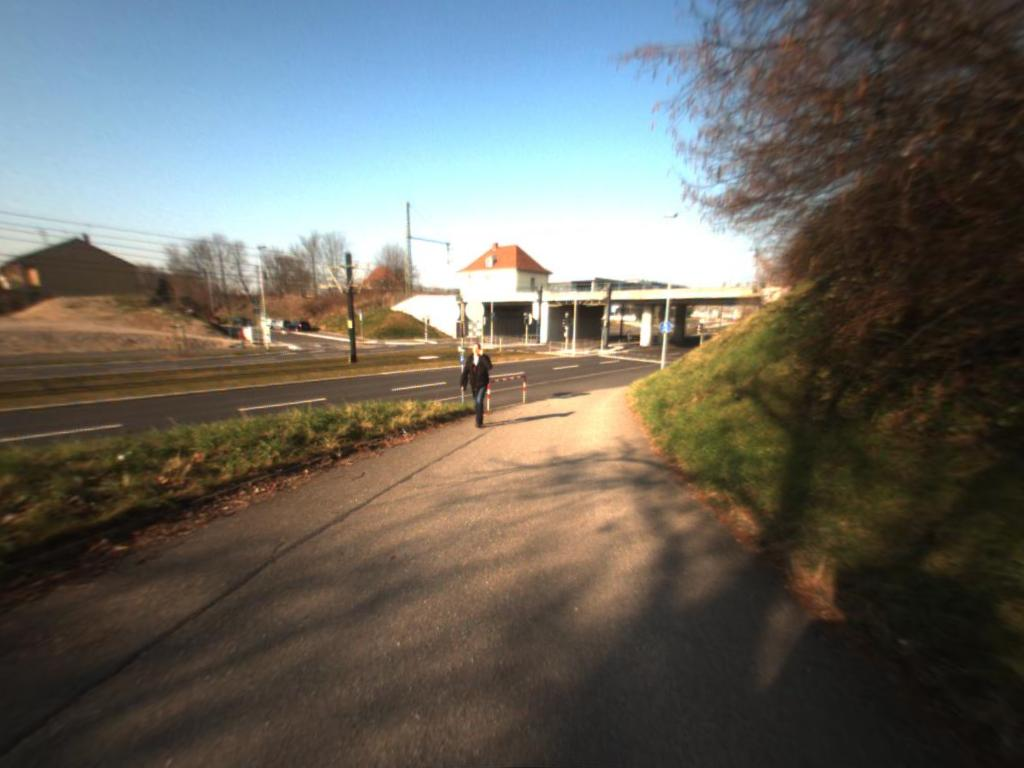} & 
          \includegraphics[width=0.95\linewidth]{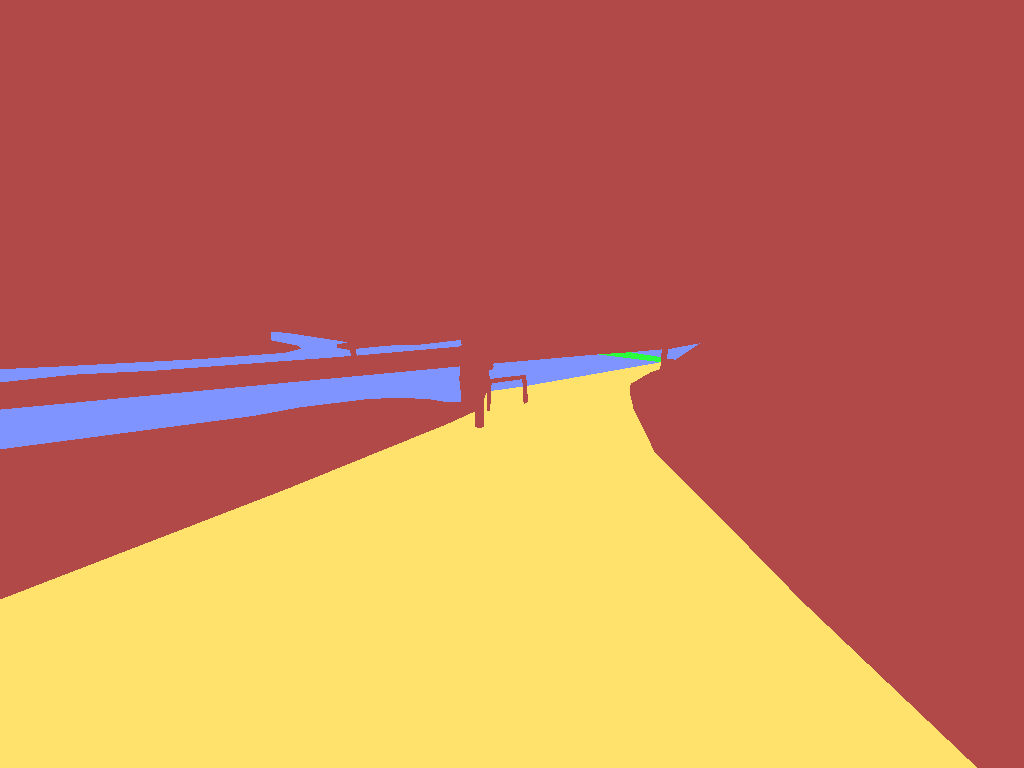} & \includegraphics[width=0.95\linewidth]{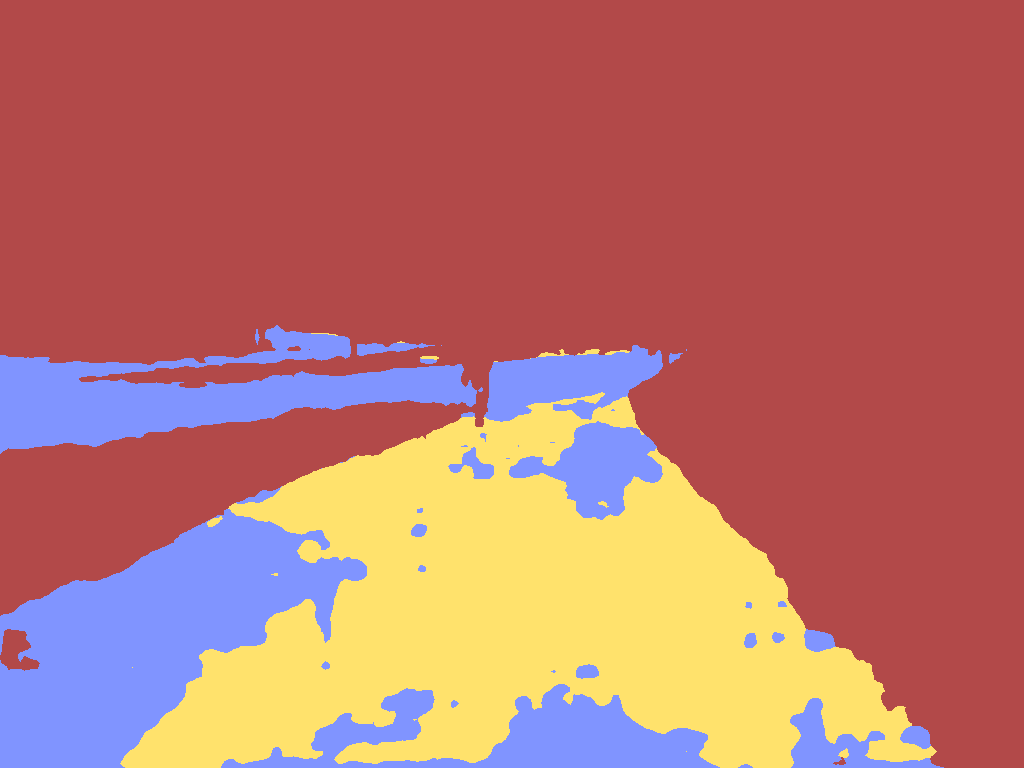} &  \includegraphics[width=0.95\linewidth]{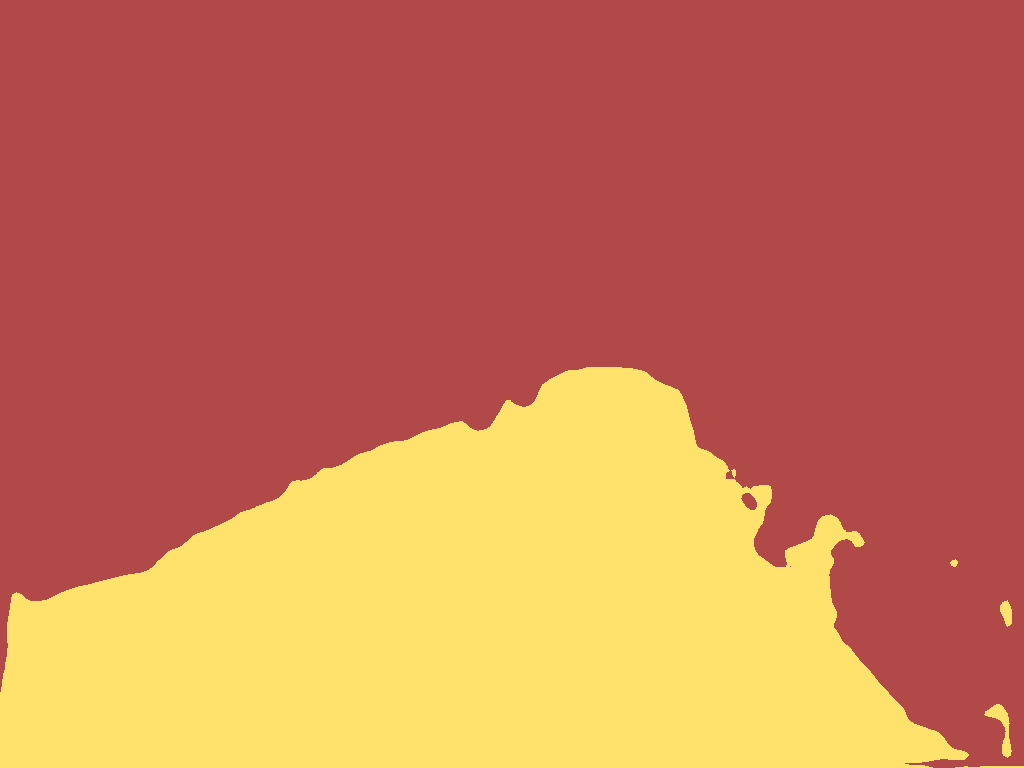} & \includegraphics[width=0.95\linewidth]{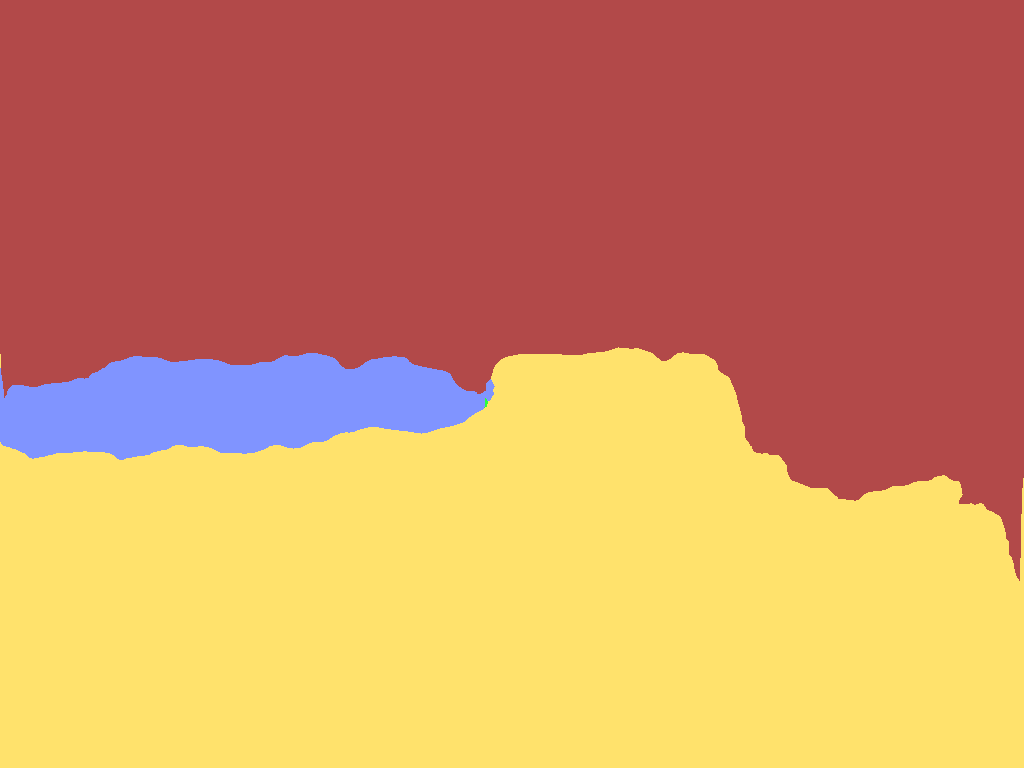} \\

    \end{tabular}
    \caption{Qualitative results of models trained on different datasets and evaluated on the test split of our \textit{Freiburg Pedestrian Scenes} dataset. We visualize RGB input images, ground truth annotations, and model predictions. Images below the horizontal gray line illustrate failure cases. Color-codes for the semantic classes are: \crule[road]{0.2cm}{0.2cm} \textit{Road}, \crule[pedestrian]{0.2cm}{0.2cm} \textit{Pedestrian}, \crule[crossing]{0.2cm}{0.2cm} \textit{Crossing}, \crule[obstacle]{0.2cm}{0.2cm} \textit{Obstacle}.}
    \label{fig:qualitative} 
\end{figure}

We compare our automatic annotation approach with several baseline approaches. All results are listed in Tab \ref{tab:results}. We first evaluate a model trained on the Vistas dataset~\cite{neuhold2017mapillary}. To perform a quantitative comparison, we re-map the Vistas classes to the \textit{Freiburg Pedestrian Scenes} class labels. While we obtain a high test mean Intersection-over-Union (mIoU) of $62.9\%$ on the Vistas test split, we obtain relatively low mIoU values of $30.4 \%$ on our dataset. This can be attributed to the domain gap between the two datasets due to the inconsistent camera viewpoints. Leveraging the robot ego-trajectory yields greatly improved results for the \textit{Pedestrian} class but cannot account for any other semantic class. Our tracklet-based method, in contrast, shows better performance than the baseline model in all classes but the \textit{Obstacle} class. What's more, our experiments indicate an improvement of IoU values when leveraging the aggregated semantic surface map (constituting dataset $\mathcal{D}_1$) for model training. This is most likely due to the larger number of annotated pixels and the increased annotation consistency due to prediction aggregation. We illustrate qualitative results in Fig.~\ref{fig:qualitative}. Generally speaking, the model trained on Vistas shows many false-positive \textit{Road}-classifications due to the camera viewpoint bias present in the Vistas dataset. We also observe that segmentation masks of our best-performing model are well aligned with the ground-truth annotations. However, due to the challenging visual similarity between ground classes, not all areas are predicted correctly. Most incorrect predictions are produced in crossing regions and in places where sidewalks and streets are not easily distinguishable (see Fig.~\ref{fig:qualitative}, failure cases).

\subsection{Evaluation of Semantic Surface Maps}

\begin{figure}

\centering
\includegraphics[width=\textwidth,height=4.8cm]{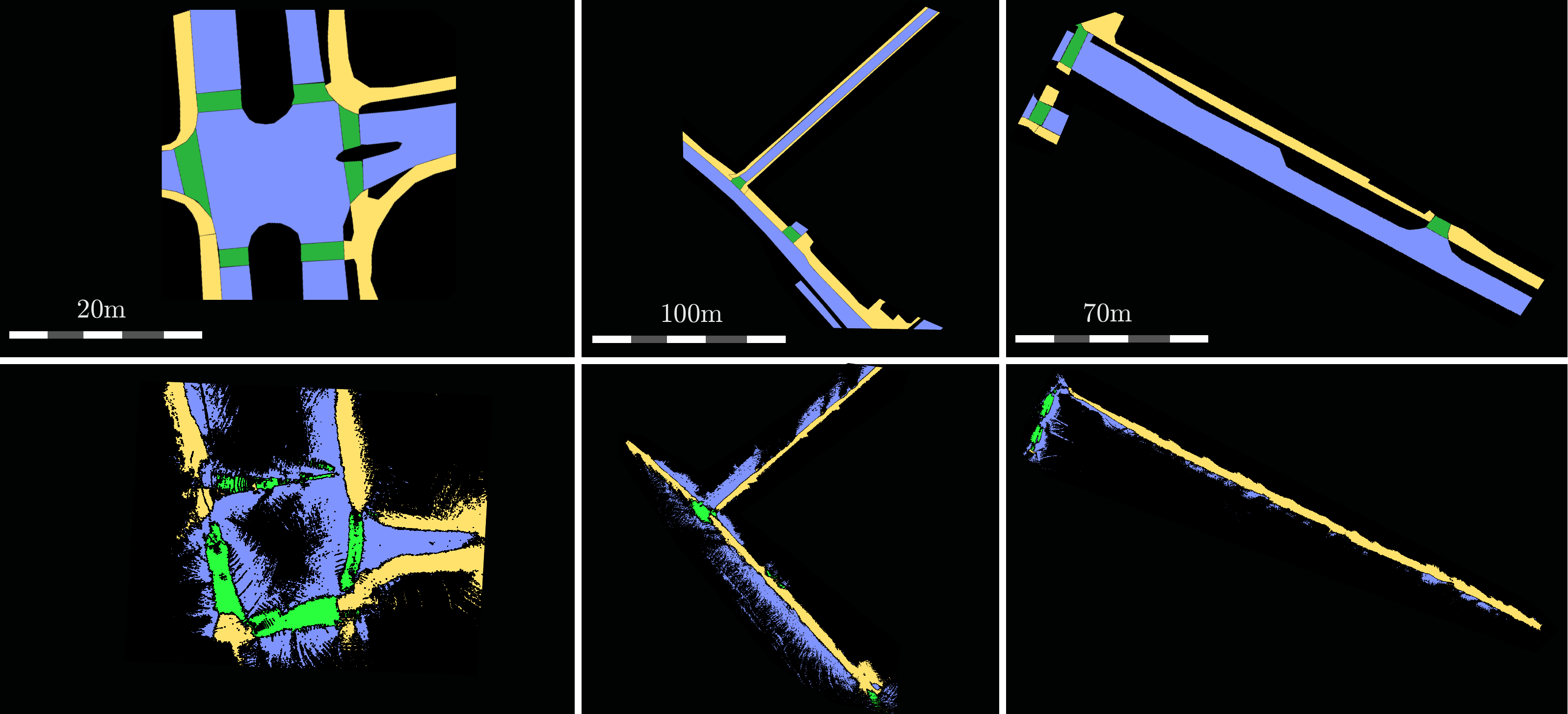}
\caption{Exemplary illustrations of maps produced with our approach (best viewed zoomed in). The top row shows the respective ground-truth map and the bottom row shows the aligned semantic surface maps obtained with our aggregation approach. Color-codes for the semantic classes are identical to Fig.~\ref{fig:qualitative}. Black color denotes not annotated / unobserved areas.}
\label{fig:bev}

\end{figure}

We qualitatively evaluate the semantic surface maps obtained with our approach. To generate the maps, we use our segmentation model and aggregate its predictions as described in Subsec. \ref{sec:mapping}. Figure \ref{fig:bev} illustrates maps produced with our approach and the respective ground truth maps. We observe that the generated maps exhibit more consistent class assignments compared to the ego-view image predictions due to the prediction aggregation procedure for map generation. We observe that in most areas, the predicted ground class equals the actual ground class. In particular, the classes \textit{Pedestrian} and \textit{Road} align well with the ground truth areas. Challenging street crossings are accounted for in all regions. However, we note that the spatial extent of some crossing regions due to \textit{Crossing/Road} and \textit{Crossing/Pedestrian} misclassifications leaves further room for improvement. \rebuttal{For more experimental evaluations, please refer to Suppl. Material Sec. H and I}. 

\subsection{Limitations}

Despite the fact that the aggregated maps are mostly well-aligned with the ground truth maps, not all annotations are correct, leading to partial bleeding of classes into each other. The class \textit{Crossing} is particularly challenging for two reasons: Firstly, the annotations produced by our approach are not always consistent since not all street crossings are covered by observed trajectories. Secondly, the \textit{Crossing} pixels have substantial overlap in terms of texture with pixels of classes \textit{Road} and \textit{Pedestrian}, requiring the model to rely on contextual information such as line markings, which is not present in all scenes. Furthermore, our approach requires highly accurate localization, sensor calibration, and object tracker performance in order to generate correct annotations. Finally, the annotation quality depends on the behavior of traffic in accordance with traffic rules. If pedestrians jaywalk to cross streets or vehicles drive in pedestrian areas, the annotations can be inconsistent, leading to reduced model performance.

\section{Conclusion}
\label{sec:conclusion}

In this work, we showed how a semantic segmentation model for urban surface segmentation can be trained from projections of the ego-trajectory and projections of tracklets of other traffic participants. We also showed that the segmentation model can be further improved via self-distillation by spatially aggregating the model predictions into a semantic map. Regarding possible future work, there is room for improvement in terms of overall segmentation quality. Furthermore, future work might include the extension of the approach to more types of traffic participants such as bicycles and railways to accommodate more urban environments and annotating higher-level map attributes compared to surface types such as road graphs and lane graphs.


\acknowledgments{We would like to thank our former colleagues at AIS for providing parts of the datasets. This includes in particular Noha Radwan, Wera Winterhalter, and Bastian Steder. We would also like to thank the reviewers for their helpful comments and suggestions. Finally, we would like to thank the DFG for providing funding for the project under grant DFG BU 865/10-2 - Autonomous Street Crossing with Pedestrian Assistant Robots.}


\bibliography{root}  

\begin{thebibliography}{37}
\providecommand{\natexlab}[1]{#1}
\providecommand{\url}[1]{\texttt{#1}}
\expandafter\ifx\csname urlstyle\endcsname\relax
  \providecommand{\doi}[1]{doi: #1}\else
  \providecommand{\doi}{doi: \begingroup \urlstyle{rm}\Url}\fi

\bibitem[Radwan et~al.(2017)Radwan, Winterhalter, Dornhege, and
  Burgard]{radwan2017did}
N.~Radwan, W.~Winterhalter, C.~Dornhege, and W.~Burgard.
\newblock Why did the robot cross the road?—learning from multi-modal sensor
  data for autonomous road crossing.
\newblock In \emph{2017 IEEE/RSJ International Conference on Intelligent Robots
  and Systems (IROS)}, pages 4737--4742. IEEE, 2017.

\bibitem[Radwan et~al.(2020)Radwan, Burgard, and Valada]{radwan2020multimodal}
N.~Radwan, W.~Burgard, and A.~Valada.
\newblock Multimodal interaction-aware motion prediction for autonomous street
  crossing.
\newblock \emph{The International Journal of Robotics Research}, 39\penalty0
  (13):\penalty0 1567--1598, 2020.

\bibitem[Z{\"u}rn and Burgard(2022)]{zurn2022self}
J.~Z{\"u}rn and W.~Burgard.
\newblock Self-supervised moving vehicle detection from audio-visual cues.
\newblock \emph{arXiv preprint arXiv:2201.12771}, 2022.

\bibitem[K{\"u}mmerle et~al.(2015)K{\"u}mmerle, Ruhnke, Steder, Stachniss, and
  Burgard]{kummerle2015autonomous}
R.~K{\"u}mmerle, M.~Ruhnke, B.~Steder, C.~Stachniss, and W.~Burgard.
\newblock Autonomous robot navigation in highly populated pedestrian zones.
\newblock \emph{Journal of Field Robotics}, 32\penalty0 (4):\penalty0 565--589,
  2015.

\bibitem[Radwan and Burgard(2018)]{radwan2018effective}
N.~Radwan and W.~Burgard.
\newblock Effective interactionaware trajectory prediction using temporal
  convolutional neural networks.
\newblock In \emph{Workshop on Crowd Navigation: Current Challenges and New
  Paradigms for Safe Robot Navigation in Dense Crowds at IEEE/RSJ International
  Conference on Intelligent Robots and Systems (IROS)}, 2018.

\bibitem[Petek et~al.(2021)Petek, Sirohi, B{\"u}scher, and
  Burgard]{petek2021robust}
K.~Petek, K.~Sirohi, D.~B{\"u}scher, and W.~Burgard.
\newblock Robust monocular localization in sparse hd maps leveraging multi-task
  uncertainty estimation.
\newblock \emph{arXiv preprint arXiv:2110.10563}, 2021.

\bibitem[Neuhold et~al.(2017)Neuhold, Ollmann, Rota~Bulo, and
  Kontschieder]{neuhold2017mapillary}
G.~Neuhold, T.~Ollmann, S.~Rota~Bulo, and P.~Kontschieder.
\newblock The mapillary vistas dataset for semantic understanding of street
  scenes.
\newblock In \emph{Proceedings of the IEEE international conference on computer
  vision}, pages 4990--4999, 2017.

\bibitem[Cordts et~al.(2016)Cordts, Omran, Ramos, Rehfeld, Enzweiler, Benenson,
  Franke, Roth, and Schiele]{cordts2016cityscapes}
M.~Cordts, M.~Omran, S.~Ramos, T.~Rehfeld, M.~Enzweiler, R.~Benenson,
  U.~Franke, S.~Roth, and B.~Schiele.
\newblock The cityscapes dataset for semantic urban scene understanding.
\newblock In \emph{Proceedings of the IEEE conference on computer vision and
  pattern recognition}, pages 3213--3223, 2016.

\bibitem[Maddern et~al.(2017)Maddern, Pascoe, Linegar, and
  Newman]{maddern20171}
W.~Maddern, G.~Pascoe, C.~Linegar, and P.~Newman.
\newblock 1 year, 1000 km: The oxford robotcar dataset.
\newblock \emph{The International Journal of Robotics Research}, 36\penalty0
  (1):\penalty0 3--15, 2017.

\bibitem[Barnes et~al.(2017)Barnes, Maddern, and Posner]{barnes2017find}
D.~Barnes, W.~Maddern, and I.~Posner.
\newblock Find your own way: Weakly-supervised segmentation of path proposals
  for urban autonomy.
\newblock In \emph{2017 IEEE International Conference on Robotics and
  Automation (ICRA)}, pages 203--210. IEEE, 2017.

\bibitem[Onozuka et~al.(2021)Onozuka, Matsumi, and Shino]{onozuka2021weakly}
Y.~Onozuka, R.~Matsumi, and M.~Shino.
\newblock Weakly-supervised recommended traversable area segmentation using
  automatically labeled images for autonomous driving in pedestrian environment
  with no edges.
\newblock \emph{Sensors}, 21\penalty0 (2):\penalty0 437, 2021.

\bibitem[Brooks and Iagnemma(2012)]{brooks2012self}
C.~A. Brooks and K.~Iagnemma.
\newblock Self-supervised terrain classification for planetary surface
  exploration rovers.
\newblock \emph{Journal of Field Robotics}, 29\penalty0 (3):\penalty0 445--468,
  2012.

\bibitem[Z{\"u}rn et~al.(2020)Z{\"u}rn, Burgard, and Valada]{zurn2020self}
J.~Z{\"u}rn, W.~Burgard, and A.~Valada.
\newblock Self-supervised visual terrain classification from unsupervised
  acoustic feature learning.
\newblock \emph{IEEE Transactions on Robotics}, 37\penalty0 (2):\penalty0
  466--481, 2020.

\bibitem[Wellhausen et~al.(2019)Wellhausen, Dosovitskiy, Ranftl, Walas, Cadena,
  and Hutter]{wellhausen2019should}
L.~Wellhausen, A.~Dosovitskiy, R.~Ranftl, K.~Walas, C.~Cadena, and M.~Hutter.
\newblock Where should i walk? predicting terrain properties from images via
  self-supervised learning.
\newblock \emph{IEEE Robotics and Automation Letters}, 4\penalty0 (2):\penalty0
  1509--1516, 2019.

\bibitem[Paz et~al.(2020)Paz, Zhang, Li, Xiang, and
  Christensen]{paz2020probabilistic}
D.~Paz, H.~Zhang, Q.~Li, H.~Xiang, and H.~I. Christensen.
\newblock Probabilistic semantic mapping for urban autonomous driving
  applications.
\newblock In \emph{2020 IEEE/RSJ International Conference on Intelligent Robots
  and Systems (IROS)}, pages 2059--2064. IEEE, 2020.

\bibitem[Otsu et~al.(2016)Otsu, Ono, Fuchs, Baldwin, and
  Kubota]{otsu2016autonomous}
K.~Otsu, M.~Ono, T.~J. Fuchs, I.~Baldwin, and T.~Kubota.
\newblock Autonomous terrain classification with co-and self-training approach.
\newblock \emph{IEEE Robotics and Automation Letters}, 1\penalty0 (2):\penalty0
  814--819, 2016.

\bibitem[Valada and Burgard(2017)]{valada2017deep}
A.~Valada and W.~Burgard.
\newblock Deep spatiotemporal models for robust proprioceptive terrain
  classification.
\newblock \emph{The International Journal of Robotics Research}, 36\penalty0
  (13-14):\penalty0 1521--1539, 2017.

\bibitem[Moosmann et~al.(2009)Moosmann, Pink, and
  Stiller]{moosmann2009segmentation}
F.~Moosmann, O.~Pink, and C.~Stiller.
\newblock Segmentation of 3d lidar data in non-flat urban environments using a
  local convexity criterion.
\newblock In \emph{2009 IEEE Intelligent Vehicles Symposium}, pages 215--220.
  IEEE, 2009.

\bibitem[Douillard et~al.(2010)Douillard, Underwood, Melkumyan, Singh,
  Vasudevan, Brunner, and Quadros]{douillard2010hybrid}
B.~Douillard, J.~Underwood, N.~Melkumyan, S.~Singh, S.~Vasudevan, C.~Brunner,
  and A.~Quadros.
\newblock Hybrid elevation maps: 3d surface models for segmentation.
\newblock In \emph{2010 IEEE/RSJ International Conference on Intelligent Robots
  and Systems}, pages 1532--1538. IEEE, 2010.

\bibitem[Aijazi et~al.(2013)Aijazi, Checchin, and
  Trassoudaine]{aijazi2013segmentation}
A.~K. Aijazi, P.~Checchin, and L.~Trassoudaine.
\newblock Segmentation based classification of 3d urban point clouds: A
  super-voxel based approach with evaluation.
\newblock \emph{Remote Sensing}, 5\penalty0 (4):\penalty0 1624--1650, 2013.

\bibitem[Mayr et~al.(2018)Mayr, Unger, and Tombari]{mayr2018self}
J.~Mayr, C.~Unger, and F.~Tombari.
\newblock Self-supervised learning of the drivable area for autonomous
  vehicles.
\newblock In \emph{2018 IEEE/RSJ International Conference on Intelligent Robots
  and Systems (IROS)}, pages 362--369. IEEE, 2018.

\bibitem[Cho et~al.(2018)Cho, Kim, Jung, Oh, Youn, and Sohn]{cho2018multi}
J.~Cho, Y.~Kim, H.~Jung, C.~Oh, J.~Youn, and K.~Sohn.
\newblock Multi-task self-supervised visual representation learning for
  monocular road segmentation.
\newblock In \emph{2018 IEEE International Conference on Multimedia and Expo
  (ICME)}, pages 1--6. IEEE, 2018.

\bibitem[Bruls et~al.(2018)Bruls, Maddern, Morye, and Newman]{bruls2018mark}
T.~Bruls, W.~Maddern, A.~A. Morye, and P.~Newman.
\newblock Mark yourself: Road marking segmentation via weakly-supervised
  annotations from multimodal data.
\newblock In \emph{2018 IEEE International Conference on Robotics and
  Automation (ICRA)}, pages 1863--1870. IEEE, 2018.

\bibitem[Wang et~al.(2019)Wang, Sun, and Liu]{wang2019self}
H.~Wang, Y.~Sun, and M.~Liu.
\newblock Self-supervised drivable area and road anomaly segmentation using
  rgb-d data for robotic wheelchairs.
\newblock \emph{IEEE Robotics and Automation Letters}, 4\penalty0 (4):\penalty0
  4386--4393, 2019.

\bibitem[K{\"u}mmerle et~al.(2011)K{\"u}mmerle, Grisetti, Strasdat, Konolige,
  and Burgard]{kummerle2011g}
R.~K{\"u}mmerle, G.~Grisetti, H.~Strasdat, K.~Konolige, and W.~Burgard.
\newblock g 2 o: A general framework for graph optimization.
\newblock In \emph{2011 IEEE International Conference on Robotics and
  Automation}, pages 3607--3613. IEEE, 2011.

\bibitem[Zhang et~al.(2021)Zhang, Sun, Jiang, Yu, Yuan, Luo, Liu, and
  Wang]{zhang2021bytetrack}
Y.~Zhang, P.~Sun, Y.~Jiang, D.~Yu, Z.~Yuan, P.~Luo, W.~Liu, and X.~Wang.
\newblock Bytetrack: Multi-object tracking by associating every detection box.
\newblock \emph{arXiv preprint arXiv:2110.06864}, 2021.

\bibitem[Tan et~al.(2020)Tan, Pang, and Le]{tan2020efficientdet}
M.~Tan, R.~Pang, and Q.~V. Le.
\newblock Efficientdet: Scalable and efficient object detection.
\newblock In \emph{Proceedings of the IEEE/CVF conference on computer vision
  and pattern recognition}, pages 10781--10790, 2020.

\bibitem[Lin et~al.(2014)Lin, Maire, Belongie, Hays, Perona, Ramanan,
  Doll{\'a}r, and Zitnick]{lin2014microsoft}
T.-Y. Lin, M.~Maire, S.~Belongie, J.~Hays, P.~Perona, D.~Ramanan,
  P.~Doll{\'a}r, and C.~L. Zitnick.
\newblock Microsoft coco: Common objects in context.
\newblock In \emph{European conference on computer vision}, pages 740--755.
  Springer, 2014.

\bibitem[Paigwar et~al.(2020)Paigwar, Erkent, Sierra-Gonzalez, and
  Laugier]{paigwar2020gndnet}
A.~Paigwar, {\"O}.~Erkent, D.~Sierra-Gonzalez, and C.~Laugier.
\newblock Gndnet: Fast ground plane estimation and point cloud segmentation for
  autonomous vehicles.
\newblock In \emph{2020 IEEE/RSJ International Conference on Intelligent Robots
  and Systems (IROS)}, pages 2150--2156. IEEE, 2020.

\bibitem[Ramos et~al.(2017)Ramos, Gehrig, Pinggera, Franke, and
  Rother]{ramos2017detecting}
S.~Ramos, S.~Gehrig, P.~Pinggera, U.~Franke, and C.~Rother.
\newblock Detecting unexpected obstacles for self-driving cars: Fusing deep
  learning and geometric modeling.
\newblock In \emph{2017 IEEE Intelligent Vehicles Symposium (IV)}, pages
  1025--1032. IEEE, 2017.

\bibitem[Zhang et~al.(2019)Zhang, Song, Gao, Chen, Bao, and Ma]{zhang2019your}
L.~Zhang, J.~Song, A.~Gao, J.~Chen, C.~Bao, and K.~Ma.
\newblock Be your own teacher: Improve the performance of convolutional neural
  networks via self distillation.
\newblock In \emph{Proceedings of the IEEE/CVF International Conference on
  Computer Vision}, pages 3713--3722, 2019.

\bibitem[Gou et~al.(2021)Gou, Yu, Maybank, and Tao]{gou2021knowledge}
J.~Gou, B.~Yu, S.~J. Maybank, and D.~Tao.
\newblock Knowledge distillation: A survey.
\newblock \emph{International Journal of Computer Vision}, 129\penalty0
  (6):\penalty0 1789--1819, 2021.

\bibitem[Asgharivaskasi and Atanasov(2021)]{asgharivaskasi2021active}
A.~Asgharivaskasi and N.~Atanasov.
\newblock Active bayesian multi-class mapping from range and semantic
  segmentation observations.
\newblock In \emph{2021 IEEE International Conference on Robotics and
  Automation (ICRA)}, pages 1--7. IEEE, 2021.

\bibitem[Dewan and Burgard(2020)]{dewan2020deeptemporalseg}
A.~Dewan and W.~Burgard.
\newblock Deeptemporalseg: Temporally consistent semantic segmentation of 3d
  lidar scans.
\newblock In \emph{2020 IEEE International Conference on Robotics and
  Automation (ICRA)}, pages 2624--2630. IEEE, 2020.

\bibitem[Taubin(1995)]{taubin1995curve}
G.~Taubin.
\newblock Curve and surface smoothing without shrinkage.
\newblock In \emph{Proceedings of IEEE international conference on computer
  vision}, pages 852--857. IEEE, 1995.

\bibitem[Chen et~al.(2018)Chen, Zhu, Papandreou, Schroff, and
  Adam]{chen2018encoder}
L.-C. Chen, Y.~Zhu, G.~Papandreou, F.~Schroff, and H.~Adam.
\newblock Encoder-decoder with atrous separable convolution for semantic image
  segmentation.
\newblock In \emph{Proceedings of the European conference on computer vision
  (ECCV)}, pages 801--818, 2018.

\bibitem[Grisetti et~al.(2010)Grisetti, K{\"u}mmerle, Stachniss, and
  Burgard]{grisetti2010tutorial}
G.~Grisetti, R.~K{\"u}mmerle, C.~Stachniss, and W.~Burgard.
\newblock A tutorial on graph-based slam.
\newblock \emph{IEEE Intelligent Transportation Systems Magazine}, 2\penalty0
  (4):\penalty0 31--43, 2010.

\end{thebibliography}

\clearpage

\pagestyle{headings}
\appendix   

\begin{center}
{\LARGE Supplementary Material}
\end{center}
\vspace{10mm}

\thispagestyle{plain}

\section{Data Recording Platform}

\begin{figure}[h]
\centering
\includegraphics[width=0.5\textwidth]{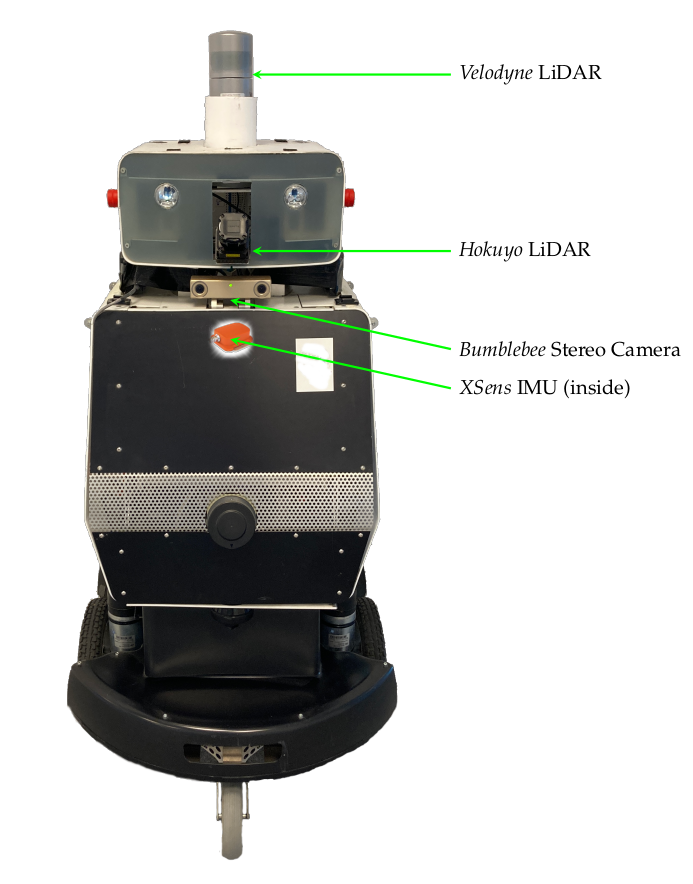}
\caption{Our robotic data recording platform is equipped with LiDAR, RGB vision, and an IMU unit.}
\label{fig:platform}
\end{figure}

Below (Fig.~\ref{fig:platform}), we schematically illustrate the data recording platform used for recording our dataset. It features a 32-beam Velodyne LiDAR mounted on top of the robot, a tilting LiDAR in the front, a Stereo RGB camera facing forward, and an IMU unit. The total height of the robot is approx. 1.80 m, offering camera viewpoints similar to those of a pedestrian. A photograph of the platform is redacted to reduce the risk of submission anonymity violation.

\section{Trajectory Projection}

\begin{figure}[h]
\centering
\includegraphics[width=0.7\textwidth]{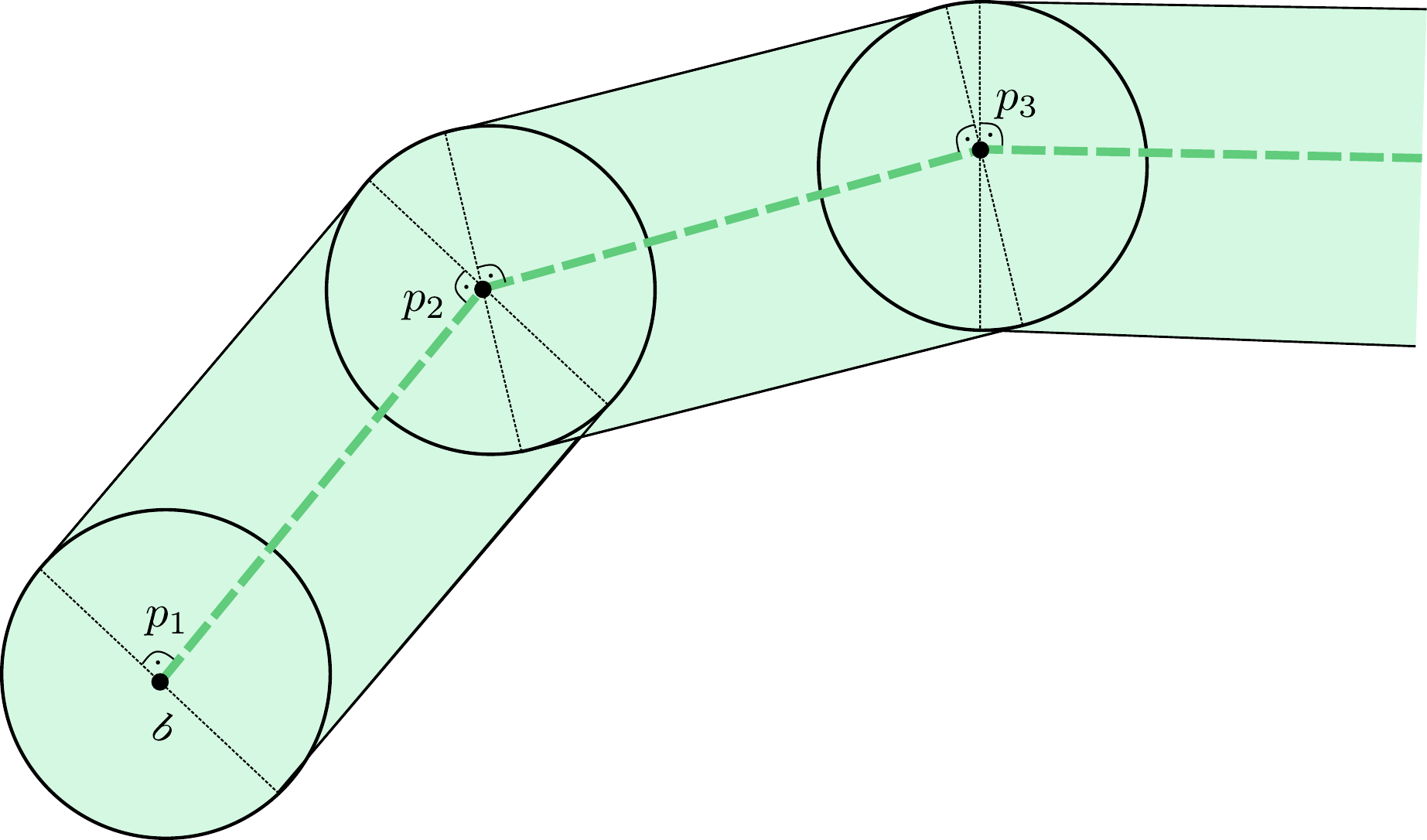}
\caption{Geometric construction of the tracklet annotations from a list of poses. The scalar $b$ denotes the object base width.}
\label{fig:geometric}
\end{figure}

In the following, we illustrate the precise scheme of pixel-annotations based on a list of poses. These poses may be the robot ego-poses or the observed tracklets of other traffic participants. Fig. \ref{fig:geometric} illustrates the geometric construction of the tracklet annotations from a list of poses. Adjacent poses $(p_i, p_{i+1})$ are connected and form a 3D surface.

\begin{figure}[h]
\centering
\includegraphics[width=0.5\textwidth]{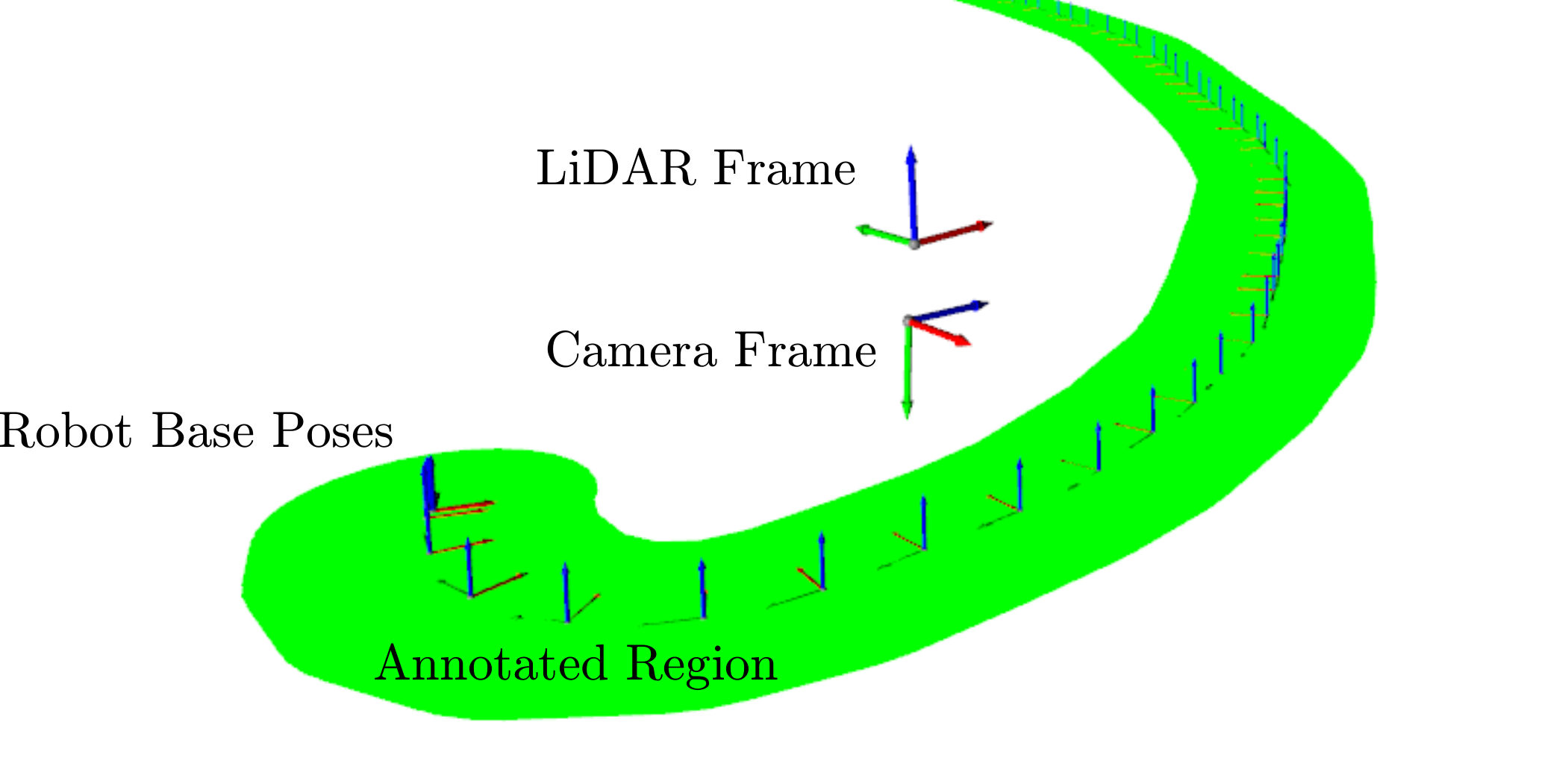}
\caption{Visualization of the one specific LiDAR coordinate frame and camera coordinate frame, and multiple object base poses translated to the ground plane. We visualize the robot trajectory as a green surface.}
\label{fig:trajectory_coordinates}
\end{figure}

We illustrate a rendered ego-trajectory surface in Fig. \ref{fig:trajectory_coordinates}. The trajectories of observed traffic participants are rendered using the same method. The object base width (i.e. footprint) changes depending on the type of traffic participant.

\section{Obstacles}

We found that the stixel-based approach described in Sec. 3.1 leads to some false-negative Obstacle annotations. We, therefore, additionally leverage the bounding boxes obtained from our object tracker and mark all pixels within detected bounding boxes as obstacles.

\section{Mesh Rendering}

To render the semantic mesh generated with our prediction aggregation scheme, we use the pyOpenGL framework. Below, we list an exemplary code snippet for initializing the shaders required to render a mesh into an image and setting parameters for the virtual camera capturing the mesh. We use a flat shader which does not consider any lighting effects, thus leading to monochrome surfaces for each surface types. When training a model on this data, the mesh colors can be mapped to a one-hot class encoding for each pixel.

\begin{lstlisting}[language=Python, caption=OpenGL Mesh Shader initialization and setting of virtual camera parameters]
window = glfw.create_window(self.w, self.h, "Projection", None, None)
glfw.make_context_current(window)

VERTEX_SHADER = """
                  #version 330
                  in vec3 position;
                  in vec3 color;
                  out vec3 newColor;

                  uniform mat4 projection;
                  uniform mat4 world_2_cam;

                  void main() {
                    //gl_Position = projection * vec4(position, 1.0f);
                    gl_Position = projection * world_2_cam * vec4(position, 1.0f);
                    newColor = color;
                    }
              """

FRAGMENT_SHADER = """
               #version 330
                in vec3 newColor;
                out vec3 outColor;
               void main() {
                  outColor = floor(newColor * 1.99);
               }
           """
shader = OpenGL.GL.shaders.compileProgram(OpenGL.GL.shaders.compileShader(VERTEX_SHADER, GL_VERTEX_SHADER),
                                          OpenGL.GL.shaders.compileShader(FRAGMENT_SHADER, GL_FRAGMENT_SHADER))
VBO = glGenBuffers(1)
glBindBuffer(GL_ARRAY_BUFFER, VBO)
glBufferData(GL_ARRAY_BUFFER, vertices.itemsize * len(vertices), vertices, GL_STATIC_DRAW)

# Create EBO
EBO = glGenBuffers(1)
glBindBuffer(GL_ELEMENT_ARRAY_BUFFER, EBO)
glBufferData(GL_ELEMENT_ARRAY_BUFFER, indices.itemsize * len(indices), indices, GL_STATIC_DRAW)

# get the position from  shader
position = glGetAttribLocation(shader, 'position')
glVertexAttribPointer(position, 3, GL_FLOAT, GL_FALSE, vertices.itemsize * 6, ctypes.c_void_p(0))
glEnableVertexAttribArray(position)

# get the color from shader
color = 1
glBindAttribLocation(shader, color, 'color')
glVertexAttribPointer(color, 3, GL_FLOAT, GL_FALSE, vertices.itemsize * 6, ctypes.c_void_p(12))
glEnableVertexAttribArray(color)

glUseProgram(shader)
glClearColor(0.0, 0.0, 0.0, 1.0)
glEnable(GL_DEPTH_TEST)  # avoids rendering triangles behind other triangle

# specify virtual camera intrinsic parameters "camera_intrinsics"
proj_loc = glGetUniformLocation(shader, "projection")
glUniformMatrix4fv(proj_loc, 1, GL_FALSE, camera_intrinsics)

# set virtual camera pose according to actual robot pose "world_2_cam"
world_2_cam_loc = glGetUniformLocation(self.shader, "world_2_cam")
glUniformMatrix4fv(world_2_cam_loc, 1, GL_FALSE, world_2_cam.T)
\end{lstlisting}

\section{Dataset Details}

In Tab.~\ref{tab:dataset_details}, we list the duration of each of the data collection runs in our \textit{Freiburg Pedestrian Scenes} dataset.

\begin{table}
\centering
\normalsize
\caption{\textit{Freiburg Pedestrian Scenes} dataset collection runs}
\begin{tabular}{|p{4cm}|>{\raggedleft\arraybackslash}p{3cm}|}
Collection Run Name & Duration [min] \\
 \midrule
\texttt{Run01}    & 36.3  \\
\texttt{Run02}    & 27.1  \\
\texttt{Run03}    & 6.1  \\
\texttt{Run04}    & 192.8  \\
\texttt{Run05}    & 70.3  \\
\texttt{Run06}    & 23.5  \\
\texttt{Run07}    & 5.3  \\
\texttt{Run08}    & 3.1  \\
\texttt{Run09}    & 2.3  \\
\texttt{Run10}    & 1.4  \\
\texttt{Run11}    & 6.8  \\
\texttt{Run12}    & 6.6  \\
\texttt{Run13}    & 17.5  \\
\texttt{Run14}    & 57.9  \\
\texttt{Run15}    & 20.7  \\
\texttt{Run16}    & 16.5  \\
\texttt{Run17}    & 114.9  \\
\texttt{Run18}    & 119.0  \\
\texttt{Run19}    & 37.1  \\
\texttt{Run20}    & 14.9  \\
\texttt{Run21}    & 41.7  \\
\texttt{Run22}    & 6.6  \\
\texttt{Run23}    & 17.5
\label{tab:dataset_details}
\end{tabular}
\end{table}

\subsection{Ground Truth BEV Semantic map}

\begin{figure}
\centering
\includegraphics[width=\textwidth]{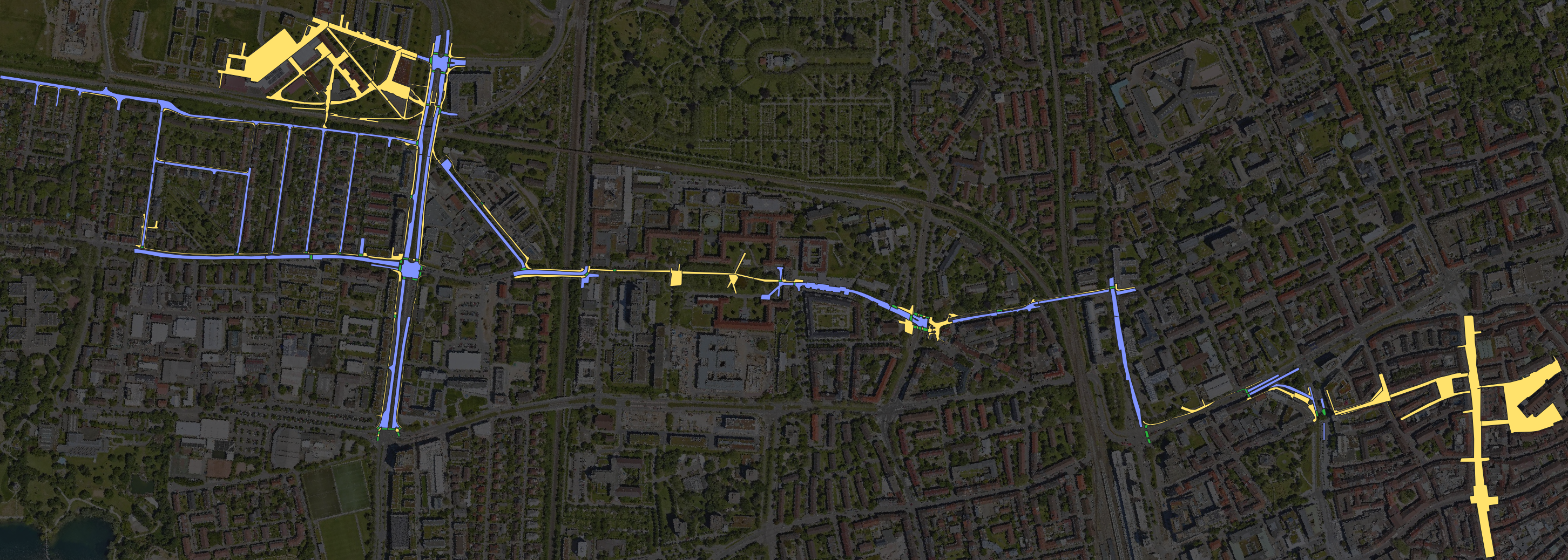}
\caption{Visualization of our BEV map annotations superimposed on an aligned RGB satellite image layer. Best viewed zoomed in. Color-codes for the semantic classes are: \crule[road]{0.2cm}{0.2cm} \textit{Road}, \crule[pedestrian]{0.2cm}{0.2cm} \textit{Pedestrian}, \crule[crossing]{0.2cm}{0.2cm} \textit{Crossing}.}
\label{fig:map_rgb}
\end{figure}

As part of our \textit{Freiburg Pedestrian Scenes} dataset, we also annotated major sections of traversed regions from a BEV perspective. The rendered annotations are shown in Fig. \ref{fig:map_rgb}. This map can serve as a reference to aggregated maps for qualitative and quantitative evaluations.

\subsection{On the Fraction of Annotated Pixels}

Using only the ego-trajectory, we were able to label 53\% of all image pixels with the classes \textit{Pedestrian} or \textit{Obstacle}. Using additional tracklets of other traffic participants, we were able to label 70\% of all pixels with the classes \textit{Road}, \textit{Crossing}, \textit{Pedestrian}, and \textit{Obstacle}. Finally, using our mesh aggregation scheme we were able to further increase the number of labeled pixels. Concretely,  With our aggregated map, we were able to label 87\% of all pixels.

\subsection{Visualization of 3D surface map}

\begin{figure}
\centering
\includegraphics[width=\textwidth]{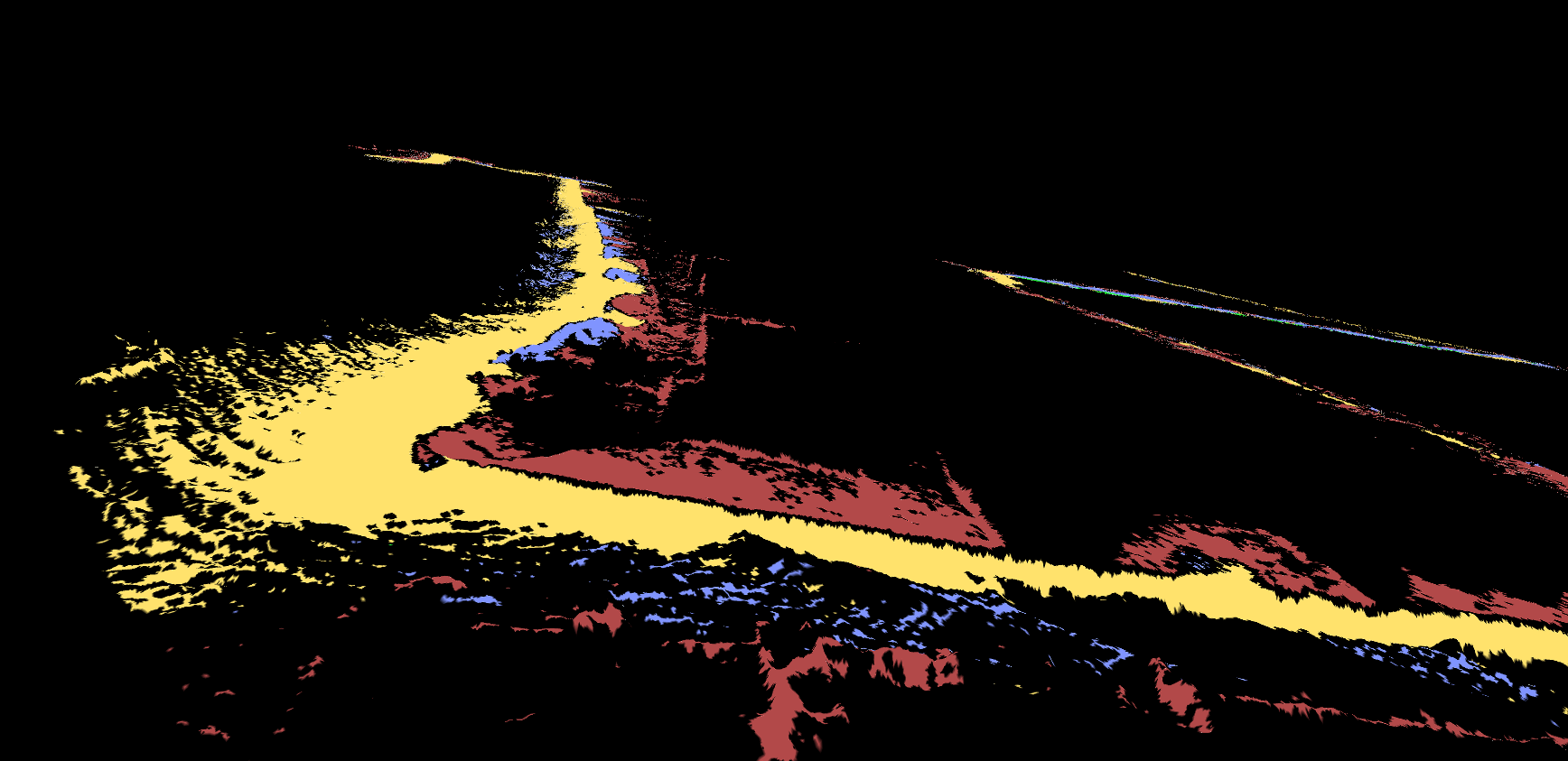}
\caption{Aggregated 3D ground map, illustrating a non-planar surface structure in the pedestrian area to the left of the image and in the obstacles present on both sides of the pedestrian pathway. Color code: \crule[road]{0.2cm}{0.2cm} \textit{Road}, \crule[pedestrian]{0.2cm}{0.2cm} \textit{Pedestrian}, \crule[crossing]{0.2cm}{0.2cm} \textit{Crossing}, \crule[obstacle]{0.2cm}{0.2cm} \textit{Obstacle}.}
\label{fig:3d_map}
\end{figure}

For illustrative purposes, we show an exemplary aggregated map in Fig. \ref{fig:3d_map}, including static obstacles close to the ground surface (red color). It features multiple regions of non-planar ground surfaces, showing our ability to model non-flat terrains with our approach.

\section{Training Details}

We train our models using the standard per-pixel weighted cross-entropy loss formulation:

\begin{equation}
    \mathcal{L} = - \sum_k \alpha_k y \log \hat{y}_k,
\end{equation}

where $\alpha_i$ denotes the loss weight for class $k$, $\hat{y}_i$ denotes the model class prediction, and $y_i$ denotes the ground-truth class. For our experiments we select the following loss class weights: $\alpha_{\text{Obstacle}} = 0.2, \alpha_{\text{Road}} = 1, \alpha_{\text{Pedestrian}} = 1, \alpha_{\text{Crossing}} = 5$, and $\alpha_{\text{Unknown}} = 0$.

We use the Adam optimizer with an initial learning rate of $\alpha = 0.001$, and parameters $\beta_0 = 0.9$ and $\beta_1 = 0.999$. The learning rate is adjusted according to an exponential decay with a decay rate of $0.9$.

\section{Evaluation of Aggregated BEV Maps}

\begin{figure}
\centering
\footnotesize
\setlength{\tabcolsep}{0.0cm}
    \begin{tabular}{P{1cm}P{5.4cm}P{5.4cm}}
    
          \texttt{map 0} & \includegraphics[width=0.95\linewidth]{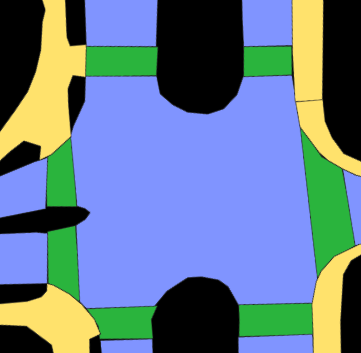} & \includegraphics[width=0.95\linewidth]{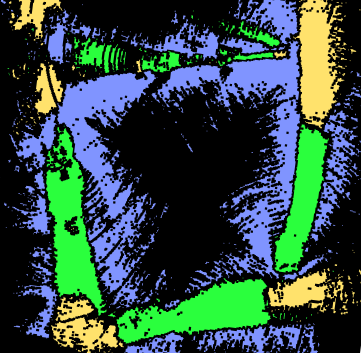} \\
          
          \texttt{map 1} & \includegraphics[width=0.95\linewidth]{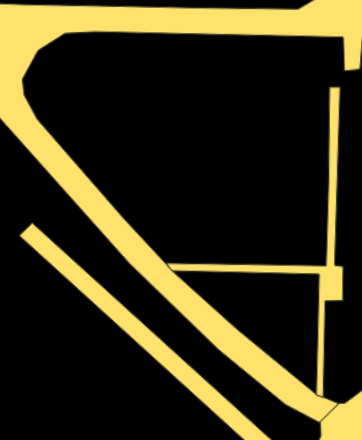} & \includegraphics[width=0.95\linewidth]{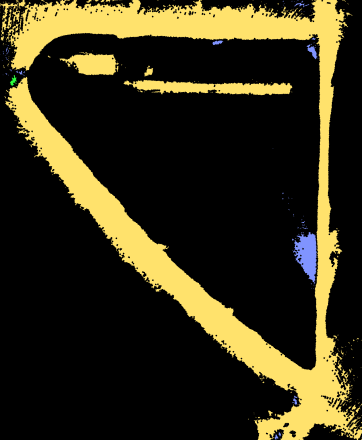} \\
          
          \texttt{map 2} & \includegraphics[width=0.95\linewidth]{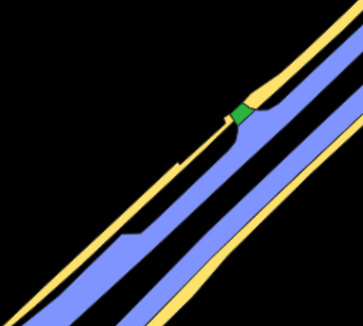} & \includegraphics[width=0.95\linewidth]{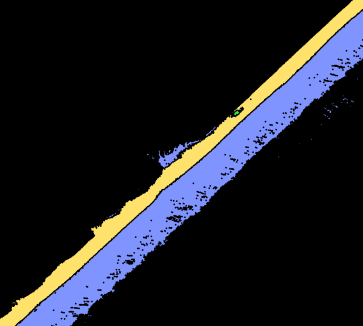} \\
        
          \texttt{map 3} & \includegraphics[width=0.95\linewidth]{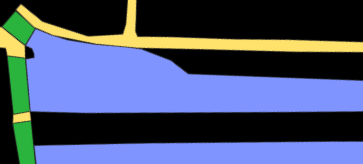} & \includegraphics[width=0.95\linewidth]{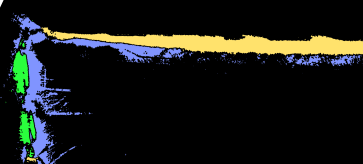} \\
      
          \texttt{map 4} & \includegraphics[width=0.95\linewidth]{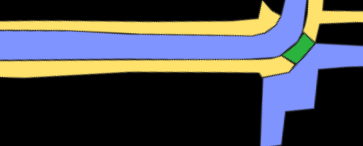} & \includegraphics[width=0.95\linewidth]{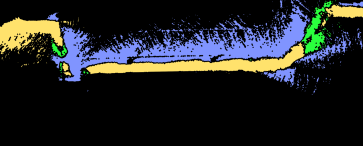} 
    \end{tabular}
    \caption{Visualization of ground-truth map annotations obtained from manual labeling efforts (left column) and corresponding crop of the aligned aggregated semantic map obtained with our approach (right column).}
    \label{fig:bev_eval} 
\end{figure}

In order to quantify the validity of the aggregated maps, we evaluate the IoU score, precision, and recall on a per-map basis. Tab. \ref{tab:bev_eval} lists the metrics for five regions while Fig. \ref{fig:bev_eval} visualizes the aggregated maps and their corresponding aligned ground-truth annotations.

We observe that for most regions, the predicted semantic ground class overlaps with the actual ground class. This hold true even for very complex environments such as the intersection depicted in map 0 and map 3. Furthermore, in most scenarios, the clear border between class \textit{Predestrian} and \textit{Road} is prominent, indicating a clear distinction between these two classes. It is crucially important for an autonomously operating robot to have a robust distinction between sidewalks and roads in order to navigate safely. We also observe misclassifications of surfaces, prominent in map 4. Note that the incompleteness of our aggregated maps stems from the fact that not all ground surfaces visible in the annotated map were visible in the onboard robot camera during the data collection runs. 

The quantitative evaluation underlines these findings. Please note that the IoU and recall values are of limited significance for evaluating the aggregated maps due to the incompleteness of these maps. The precision metric, in contrast, is more meaningful in this context. We find that for most maps, decent precision values (values generally above 50 \%) are obtained, indicating that when a surface patch is observed in the robot camera, the prediction quality for this patch is high.

\begin{table}
\centering
\caption{BEV map performance evaluation for five maps. We denote all metrics in \%.}
\begin{tabular}{|p{2cm}|p{1.5cm}|>{\raggedleft\arraybackslash}p{1.3cm}|>{\raggedleft\arraybackslash}p{2.0cm}|>{\raggedleft\arraybackslash}p{1.8cm}|>{\raggedleft\arraybackslash}p{1.8cm}}
Map Name & Metric & \crule[road]{0.2cm}{0.2cm} Road	& \crule[pedestrian]{0.2cm}{0.2cm} Pedestrian & \crule[crossing]{0.2cm}{0.2cm} Crossing  \\
\hline
Map 0  & IoU &   17.3 &  22.4 &  25.4  \\
  & Precision & 58.0 & 57.8 & 34.6   \\
  & Recall & 18.2 & 23.7 &  33.4  \\
  \hline
Map 1  & IoU & 0.0  & 55.0 &  0.0  \\
  & Precision & 0.0 & 57.6 &   0.0 \\
  & Recall & 0.0 & 55.0 &  0.0  \\
    \hline

Map 2  & IoU &  50.2 & 53.9 &  2.6  \\
  & Precision & 57.9 & 41.3 & 100.0   \\
  & Recall &  50.2 &  54.9 &  2.63  \\
    \hline

Map 3  & IoU &  4.0 &   62.8 &  21.2 \\
  & Precision & 44.0 & 69.2 &   70.0 \\
  & Recall & 40.2 &  62.9 &   22.0 \\
    \hline

Map 4  & IoU & 24.6 & 28.2 & 5.9   \\
  & Precision & 74.0 & 64.3 &  8.4  \\
  & Recall &  25.1  &  30.5 &  9.2  \\

\end{tabular}
\label{tab:bev_eval}
\end{table}

\section{Path Planning Experiments}

\begin{figure}[h]
\centering
\includegraphics[width=\textwidth]{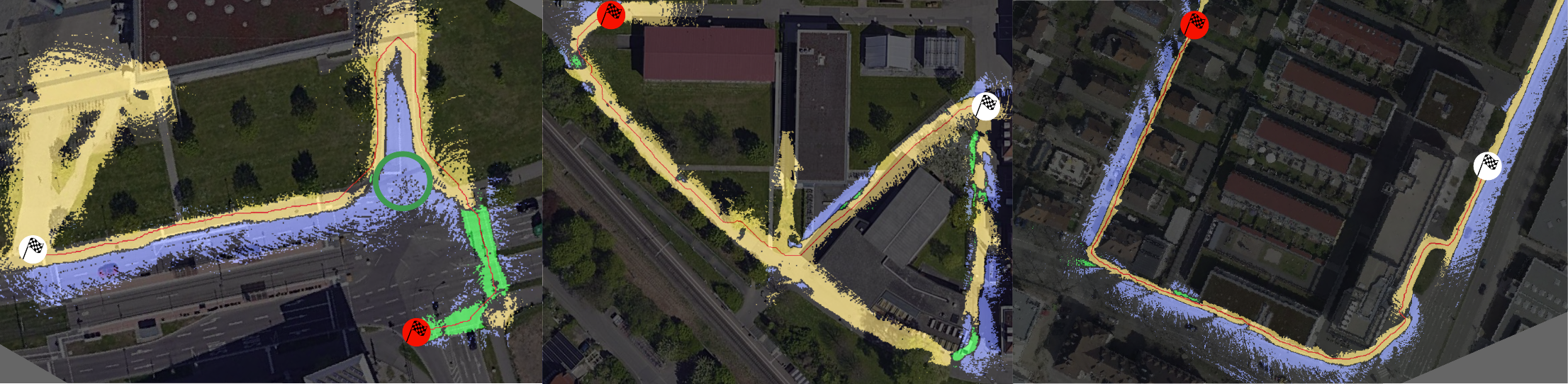}
\caption{Path planning experiments on three exemplary complex urban areas. We superimpose the color-coded semantic map onto an aligned satellite image. The start and goal positions are indicated with red and white flags, respectively. The planned route according to the semantic map is indicated as a red line. Best viewed zoomed in. Color code: \crule[road]{0.2cm}{0.2cm} \textit{Road}, \crule[pedestrian]{0.2cm}{0.2cm} \textit{Pedestrian}, \crule[crossing]{0.2cm}{0.2cm} \textit{Crossing}.}
\label{fig:planning}
\end{figure}

In addition to the quantitative IoU evaluation of the aggregated BEV maps, we conduct additional path planning experiments. One intended use-case of our map aggregation scheme is the ability for an autonomous robots to perform high-level planning on the aggregated semantic maps. We, therefore, convert the semantic class map into a costmap where each class is associated with a traversability cost. Since our robot is supposed to operate and navigate alongside pedestrians, we associate high cost with the classes \textit{Road} and \textit{Unknown}, while we associate low cost with the classes \textit{Pedestrian} and \textit{Crossing}. Finally, we smooth the produced costmap with a Gaussian filter to encourage the search algorithm to follow pathways that are centered within a given corridor of low-cost traversability such as sidewalks. We subsequently perform an A* search on the costmap to find optimal routes between a start position and a goal position. Fig. \ref{fig:planning} illustrates three exemplary planning tasks in complex urban areas. 

The results show that it is possible to use the semantic map as a data source for a planning algorithm. The planned path follows legal pathways through complex surroundings such as street crossings and sidewalks. As long as the SLAM solution to a given data collection run is accurate, large-scale maps such as shown in \ref{fig:planning}, rightmost map, are possible to generate. Fig. \ref{fig:planning}, leftmost map, shows an interesting failure case where the map does not contain a street crossing that would shorten the overall route length from start position to goal position (indicated with a green circle). In this case, the map contains a longer but also safe route across the street closer to the building where the street surface is correctly classified as a pedestrian area (it turns into a pedestrian area after the crossing).

\section{Exemplary visualization of Tracklet Annotations}

\begin{figure}
\centering
\footnotesize
\setlength{\tabcolsep}{0.0cm}
    \begin{tabular}{P{3.5cm}P{3.5cm}P{3.5cm}P{3.5cm}}

          \includegraphics[trim={0 0 0 4cm},clip,width=\linewidth]{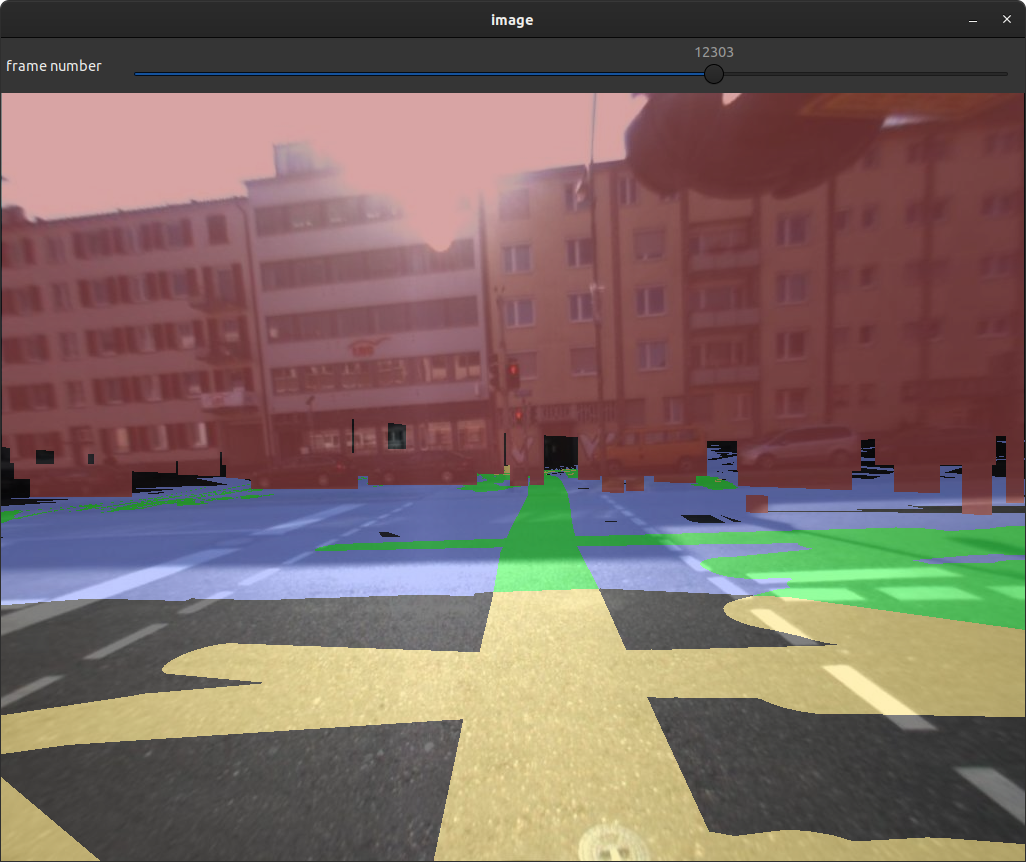} & 
          \includegraphics[trim={0 0 0 4cm},clip,width=\linewidth]{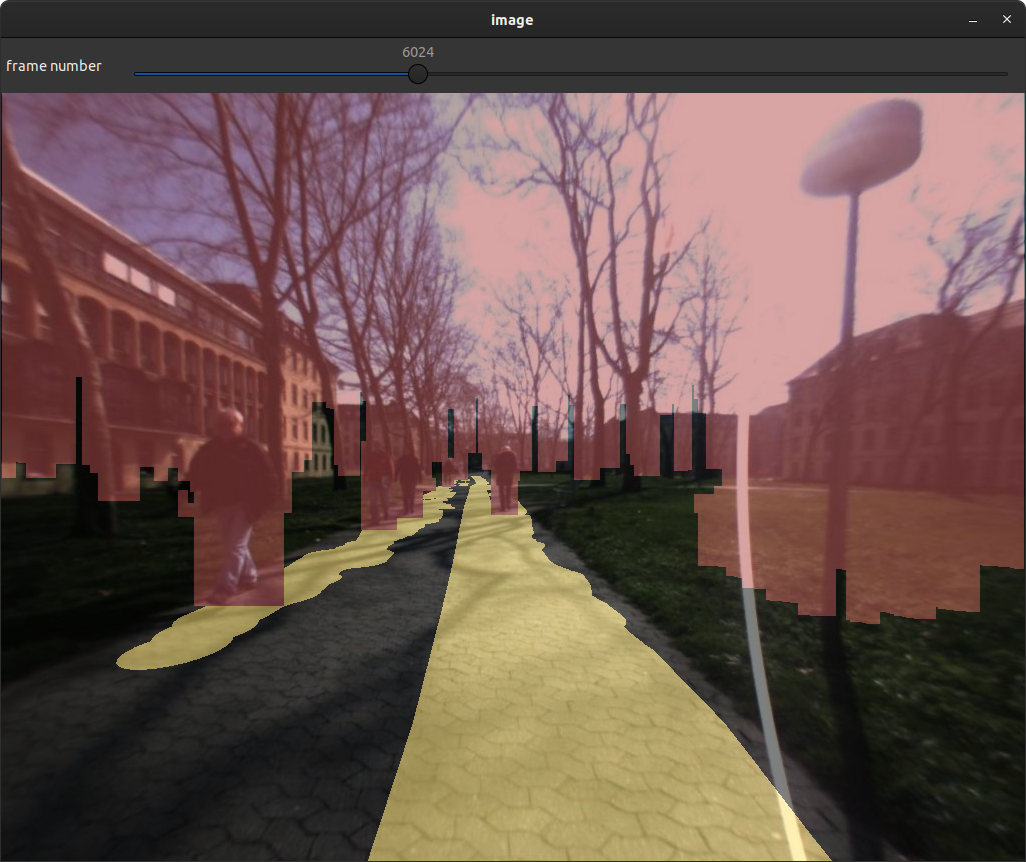} & \includegraphics[trim={0 0 0 4cm},clip,width=\linewidth]{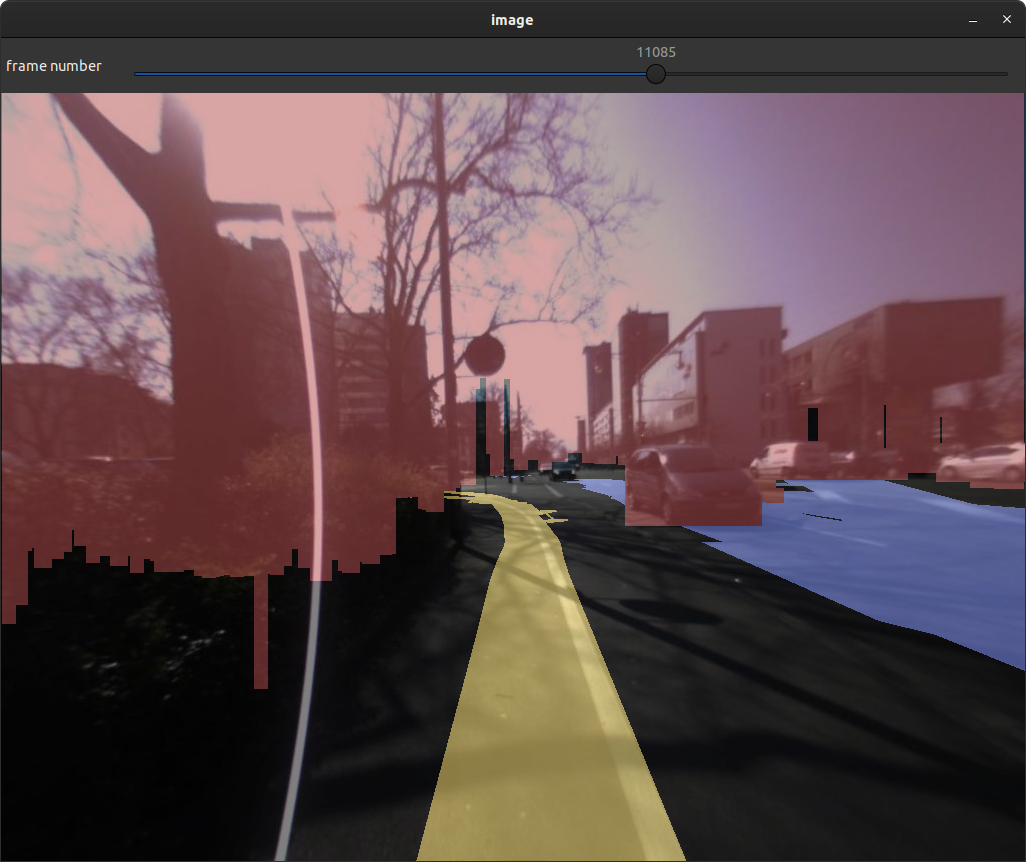} & \includegraphics[trim={0 0 0 4cm},clip,width=\linewidth]{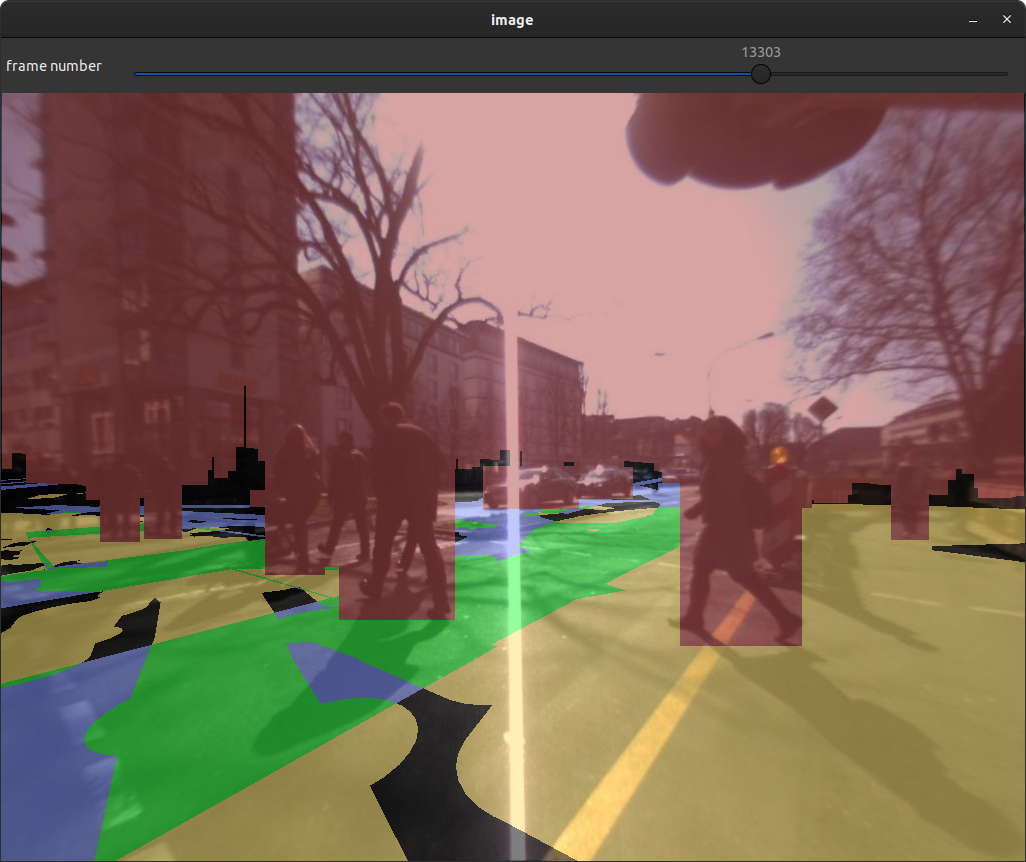} \\
          
          \includegraphics[trim={0 0 0 4cm},clip,width=\linewidth]{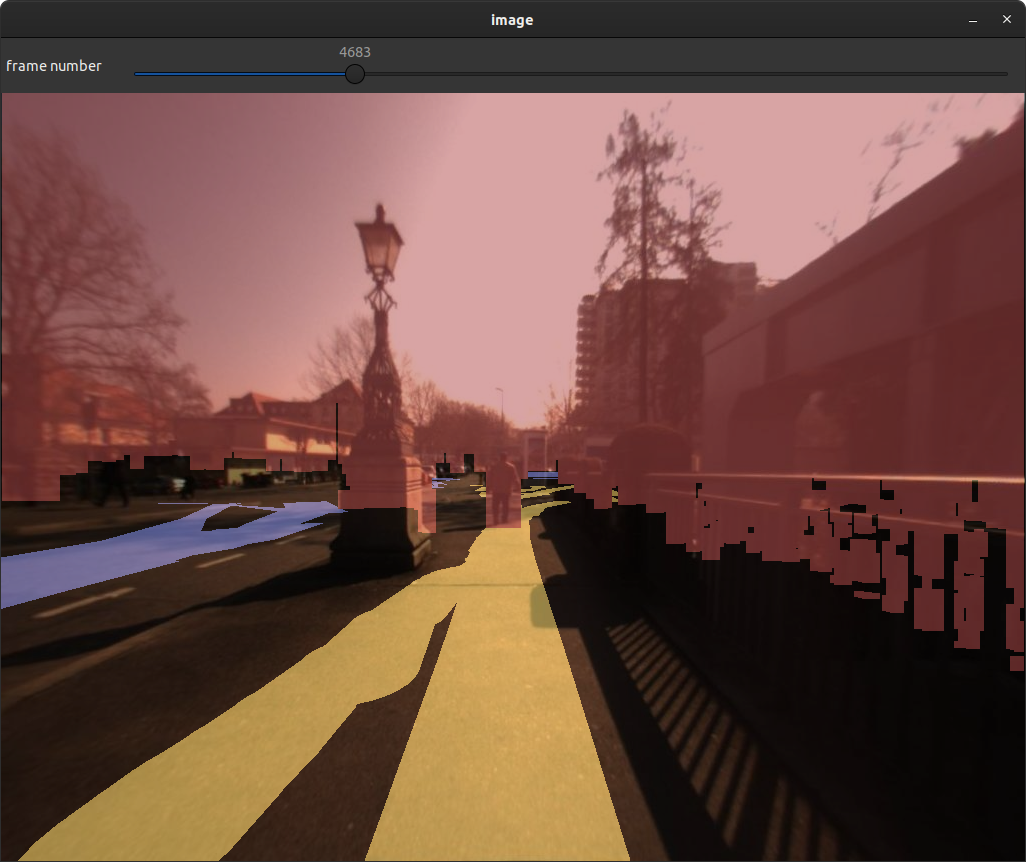} & 
          \includegraphics[trim={0 0 0 4cm},clip,width=\linewidth]{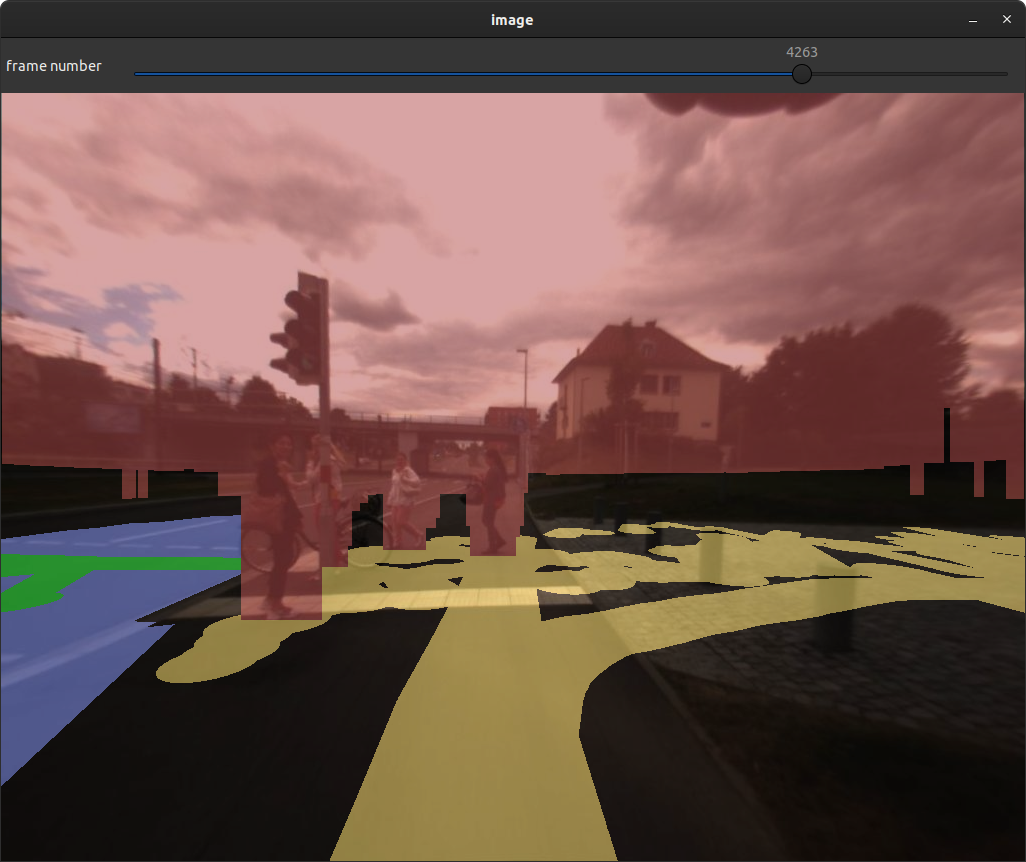} & \includegraphics[trim={0 0 0 4cm},clip,width=\linewidth]{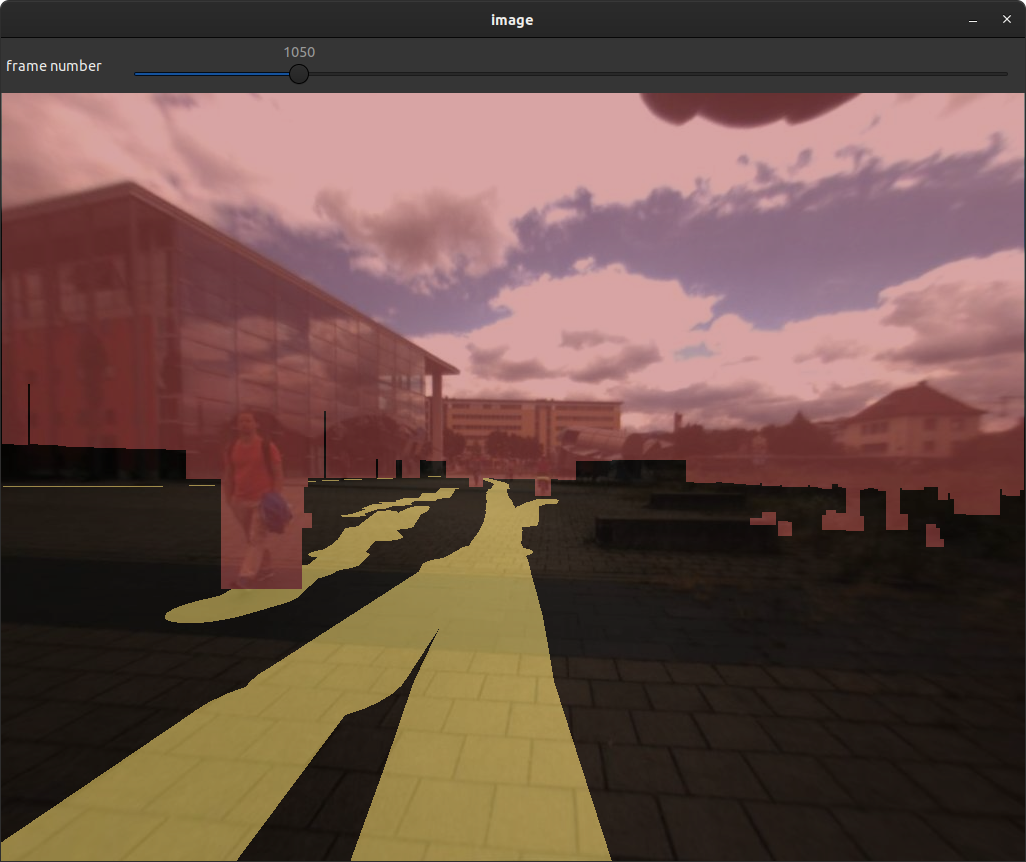} & \includegraphics[trim={0 0 0 4cm},clip,width=\linewidth]{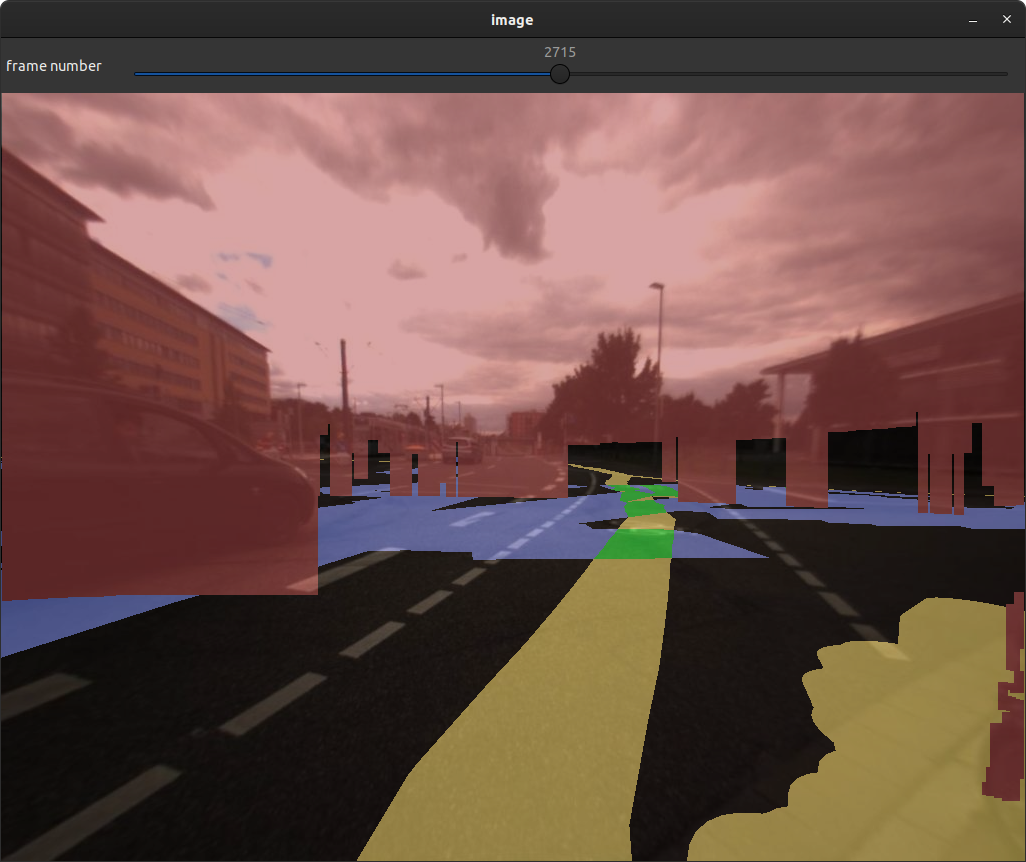} 

    \end{tabular}
    \caption{Exemplary visualizations of annotation masks obtained with our tracklet-based annotation scheme. Color code: \crule[road]{0.2cm}{0.2cm} \textit{Road}, \crule[pedestrian]{0.2cm}{0.2cm} \textit{Pedestrian}, \crule[crossing]{0.2cm}{0.2cm} \textit{Crossing}, \crule[obstacle]{0.2cm}{0.2cm} \textit{Obstacle}.}
    \label{fig:tracklets} 
\end{figure}

In Fig. \ref{fig:tracklets}, we illustrate exemplary semantic annotations obtained from the projected tracklets in each scene (dataset $\mathcal{D}_0$).

\section{Exemplary visualization of Semantic Map Projections}

\begin{figure}
\centering
\footnotesize
\setlength{\tabcolsep}{0.0cm}
    \begin{tabular}{P{3.5cm}P{3.5cm}P{3.5cm}P{3.5cm}}

          \includegraphics[trim={0 0 0 4cm},clip,width=\linewidth]{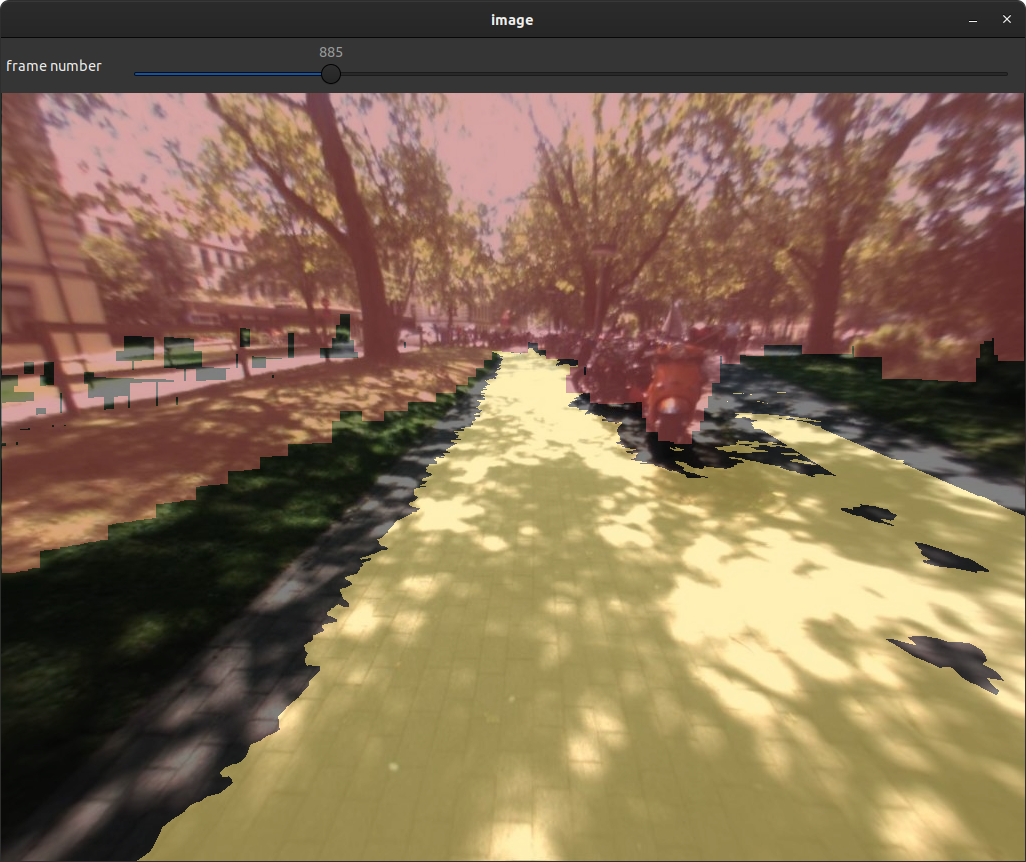} & 
          \includegraphics[trim={0 0 0 4cm},clip,width=\linewidth]{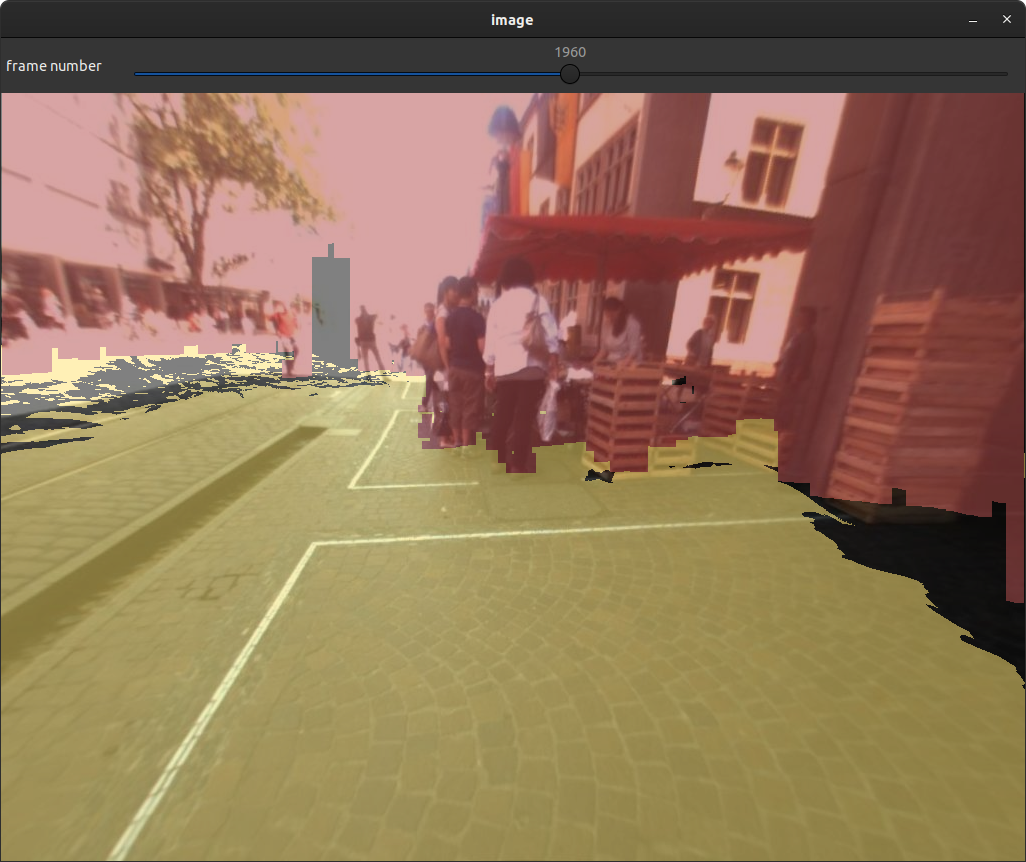} & \includegraphics[trim={0 0 0 4cm},clip,width=\linewidth]{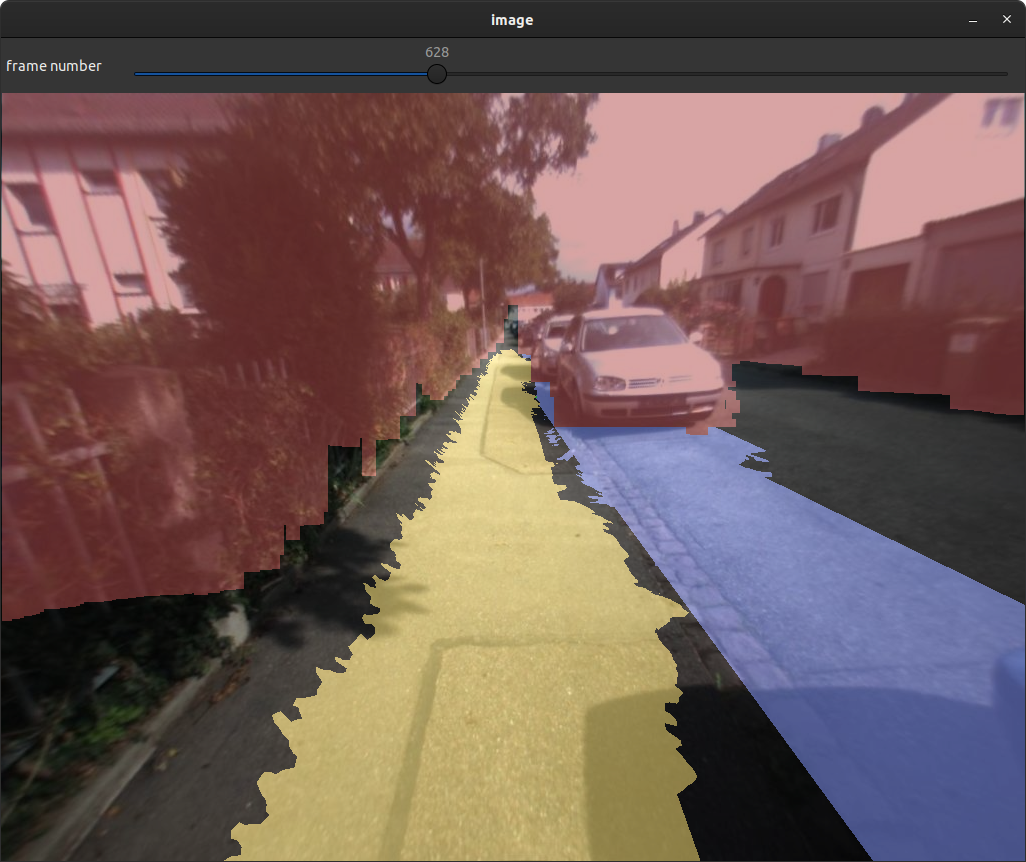} & \includegraphics[trim={0 0 0 4cm},clip,width=\linewidth]{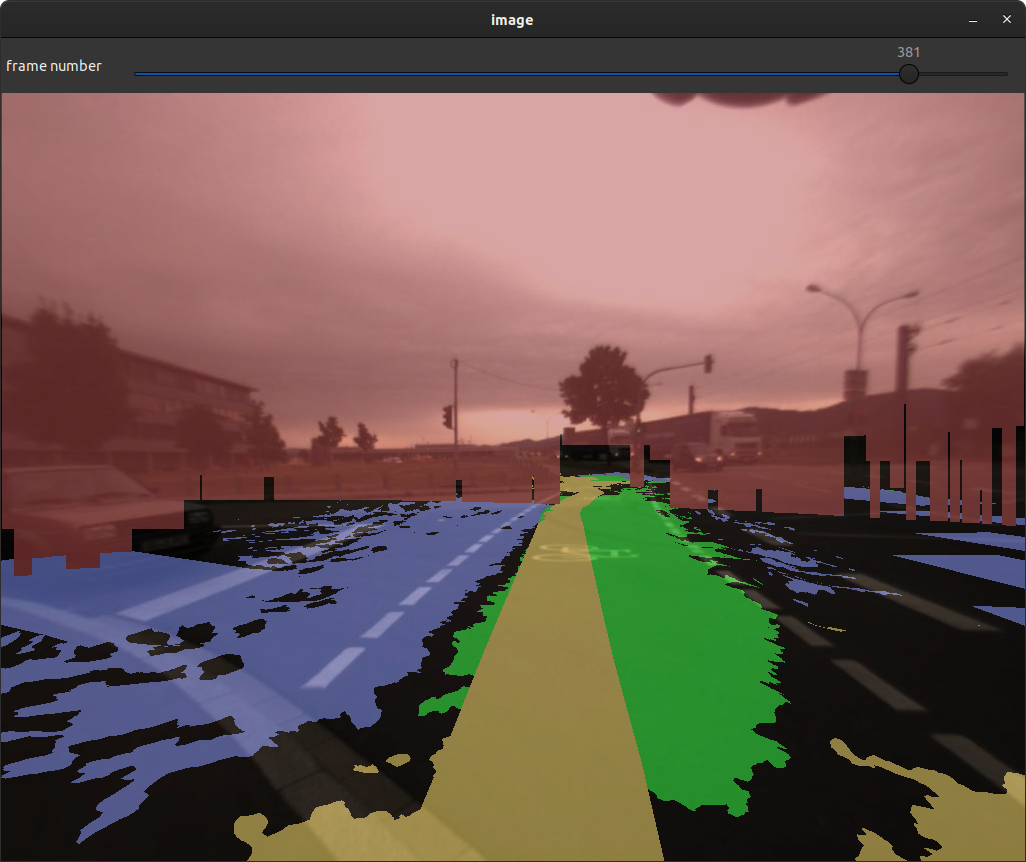} \\
          
          \includegraphics[trim={0 0 0 4cm},clip,width=\linewidth]{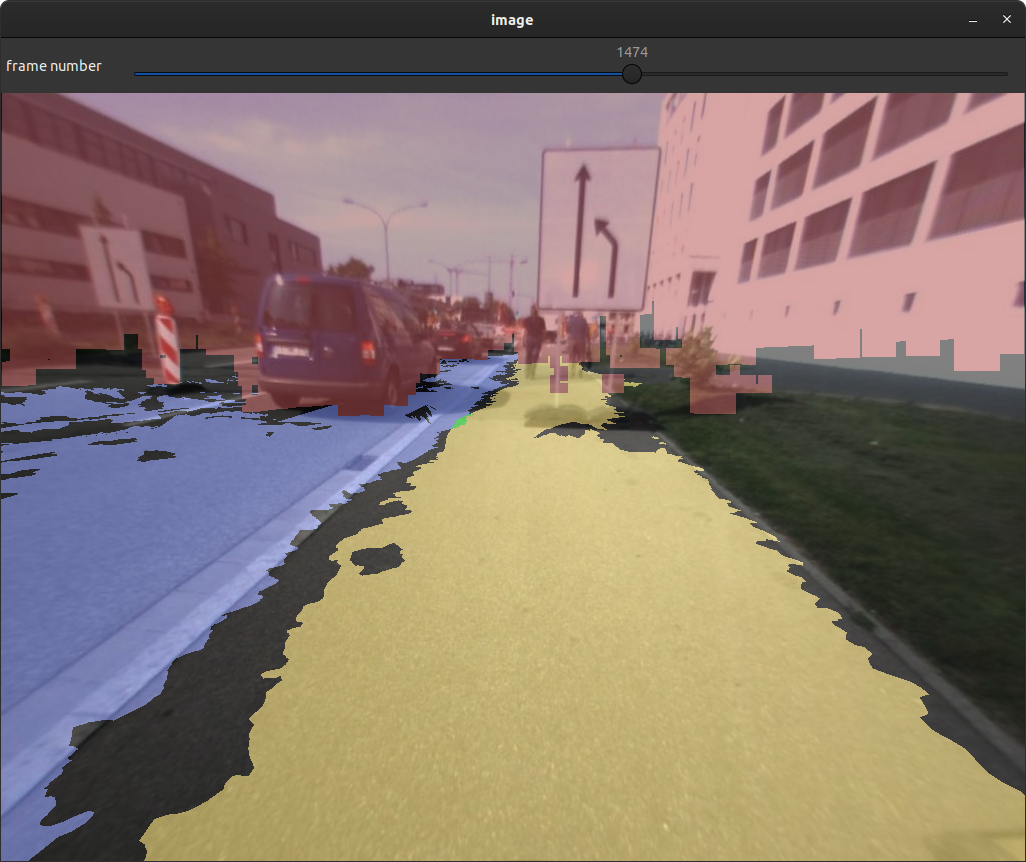} & 
          \includegraphics[trim={0 0 0 4cm},clip,width=\linewidth]{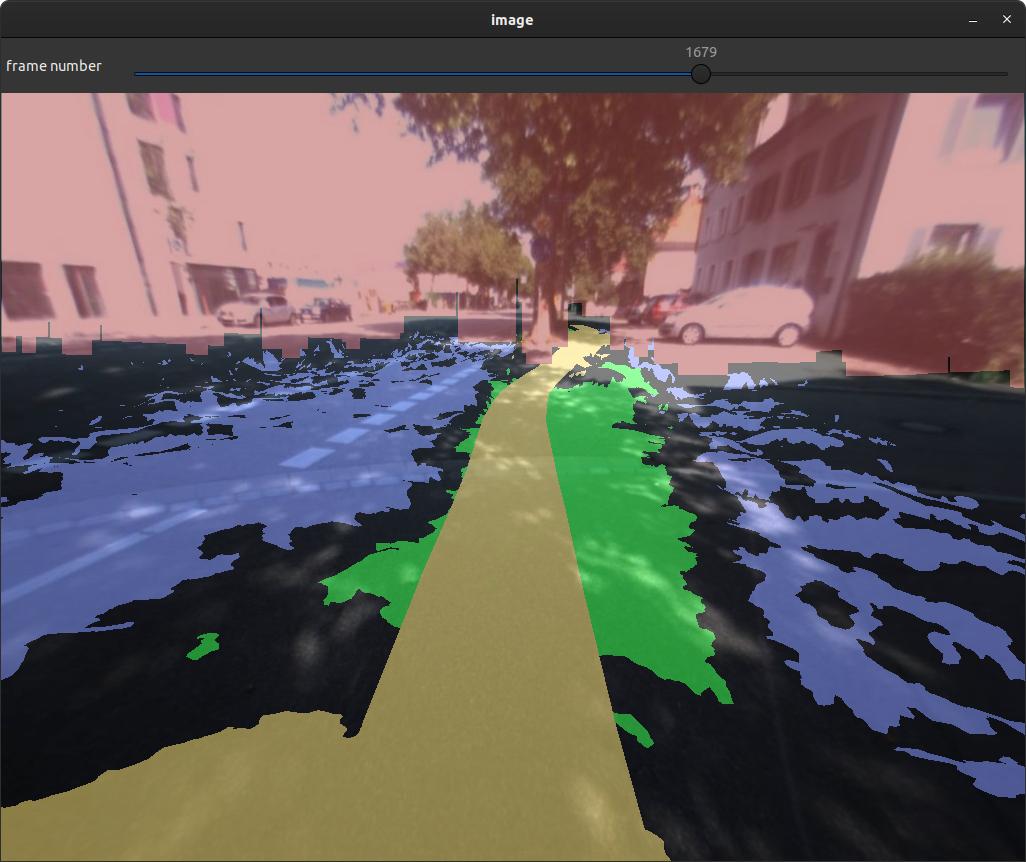} & \includegraphics[trim={0 0 0 4cm},clip,width=\linewidth]{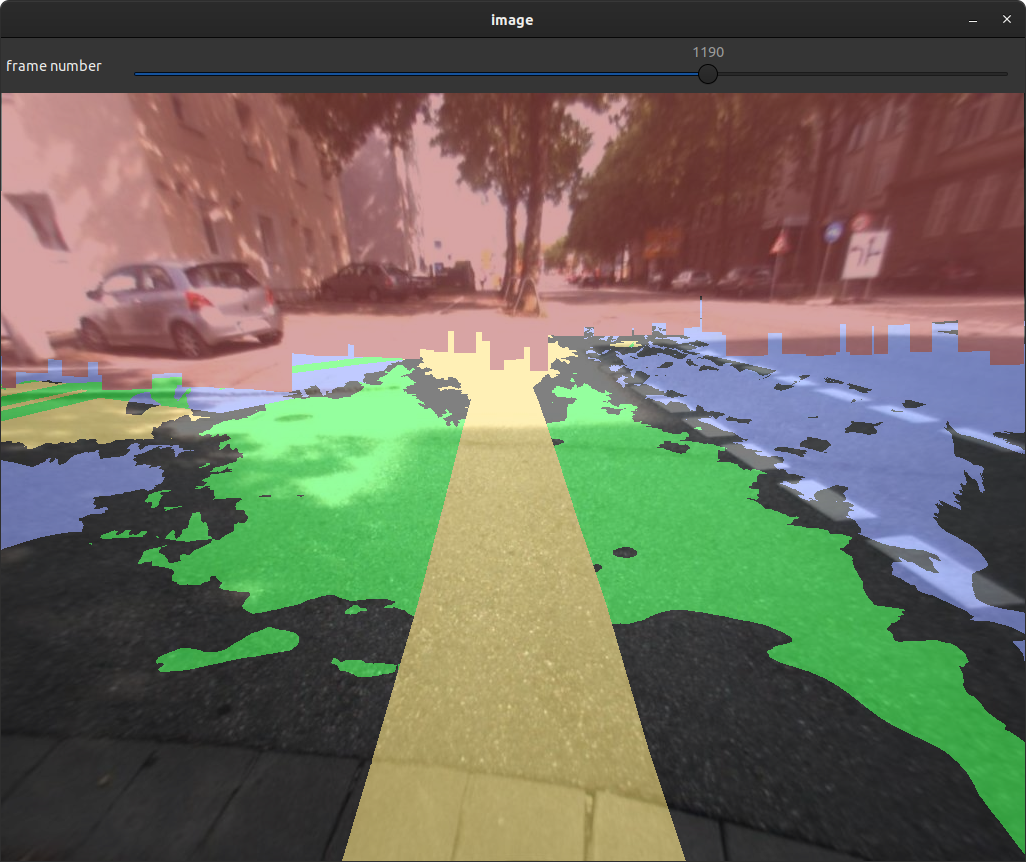} & \includegraphics[trim={0 0 0 4cm},clip,width=\linewidth]{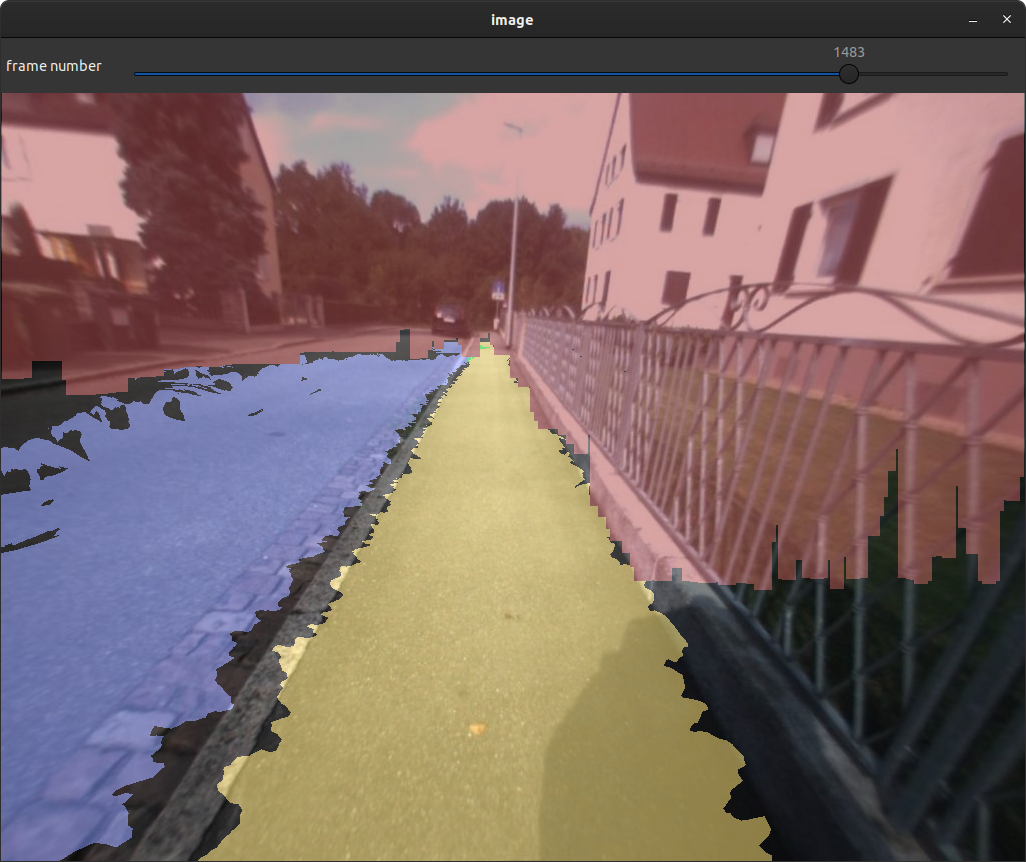} 

    \end{tabular}
    \caption{Exemplary visualizations of annotation masks obtained with our map reprojection annotation scheme. Note that the yellow-colored ego-trajectory is superimposed on the projected map for visualization purposes and is not used to provide the annotations for the model. Color code: \crule[road]{0.2cm}{0.2cm} \textit{Road}, \crule[pedestrian]{0.2cm}{0.2cm} \textit{Pedestrian}, \crule[crossing]{0.2cm}{0.2cm} \textit{Crossing}, \crule[obstacle]{0.2cm}{0.2cm} \textit{Obstacle}.}
    \label{fig:reproj} 
\end{figure}

In Fig. \ref{fig:reproj}, we illustrate exemplary map projections obtained from the aggregated surface maps in each scene  (dataset $\mathcal{D}_1$). Note how the number of labeled pixels is increased compared to the annotations in Fig. \ref{fig:tracklets}.

\end{document}